\documentclass{article}
\usepackage{bm}
\usepackage{url}
\usepackage[utf8]{inputenc}
\usepackage{amsmath}
\usepackage{amssymb}
\usepackage{fullpage}
\usepackage{bbm}
\usepackage[normalem]{ulem}
\usepackage{graphicx}
\usepackage[usenames]{xcolor}
\usepackage[font=small,labelfont=bf]{caption}
\usepackage{mathtools}
\usepackage{cite}
\usepackage{titlesec}
\usepackage{etoolbox}
\newcommand{\kl}[1]{\left( #1 \right)}
\newcommand{\ekl}[1]{\left[ #1 \right]}

\newcommand{\xx}[1]{$\times$}

\usepackage[singlespacing]{setspace}

\title{Unmasking Clever Hans Predictors and\\ Assessing What Machines Really Learn}
\author
{Sebastian Lapuschkin$^{1}$, Stephan W\"aldchen$^{2}$, Alexander Binder$^{3}$, \\
Gr\'egoire Montavon$^{2}$, Wojciech Samek$^{1\dag}$ and Klaus-Robert M\"uller$^{2,4,5\dag}$\thanks{This preprint has been accepted for publication and will appear as Lapuschkin et al.\ ``Unmasking Clever Hans Predictors and Assessing What Machines Really Learn'', {\it Nature Communications}, 2019. {http://dx.doi.org/10.1038/s41467-019-08987-4}}\\
\\
\normalsize{$^{1}$Department of Video Coding \& Analytics, Fraunhofer Heinrich Hertz Institute}\\
\normalsize{Einsteinufer 37, 10587 Berlin, Germany}\\
\normalsize{$^{2}$Department of Electrical Engineering and Computer Science, Technische Universit\"at Berlin}\\
\normalsize{Marchstr. 23, 10587 Berlin, Germany}\\
\normalsize{$^{3}$ISTD Pillar, Singapore University of Technology and Design}\\
\normalsize{8 Somapah Rd, Singapore 487372, Singapore}\\
\normalsize{$^{4}$Department of Brain and Cognitive Engineering, Korea University}\\
\normalsize{Anam-dong, Seongbuk-ku, Seoul 136-713, Republic of Korea}\\
\normalsize{$^{5}$Max Planck Institut f\"ur Informatik}\\
\normalsize{Campus E1 4, Stuhlsatzenhausweg, 66123 Saarbr\"ucken, Germany}\\
\\
\normalsize{$^\dag$Corresponding authors. E-mail:  wojciech.samek@hhi.fraunhofer.de, } \\
\normalsize{klaus-robert.mueller@tu-berlin.de.}
}
\date{}

\newcommand{\x}{\textrm{\bf x}}
\newcommand{\R}{\textrm{\bf{R}}}

\begin{document} 
\maketitle 

\begin{abstract}
Current learning machines have successfully solved hard application problems, reaching high accuracy and displaying seemingly “intelligent” behavior. Here we apply recent techniques for explaining decisions of state-of-the-art learning machines and analyze various tasks from computer vision and arcade games. This showcases a spectrum of problem-solving behaviors ranging from naive and short-sighted, to well-informed and strategic. We observe that standard performance evaluation metrics can be oblivious to distinguishing these diverse problem solving behaviors. Furthermore, we propose our semi-automated Spectral Relevance Analysis that provides a practically effective way of characterizing and validating the behavior of nonlinear learning machines. This helps to assess whether a learned model indeed delivers reliably for the problem that it was conceived for. Furthermore, our work intends to add a voice of caution to the ongoing excitement about machine intelligence and pledges to evaluate and judge some of these recent successes in a more nuanced manner.
\end{abstract}

\section{Introduction}
Artificial intelligence systems, based on machine learning, are increasingly assisting our daily life. They enable industry and the sciences to convert a never ending stream of data -- which per se is not informative -- into information that may be helpful and actionable. Machine learning has become a basis of many services and products that we use.

\begin{figure}[t]
\centering
\includegraphics[width=0.99\textwidth]{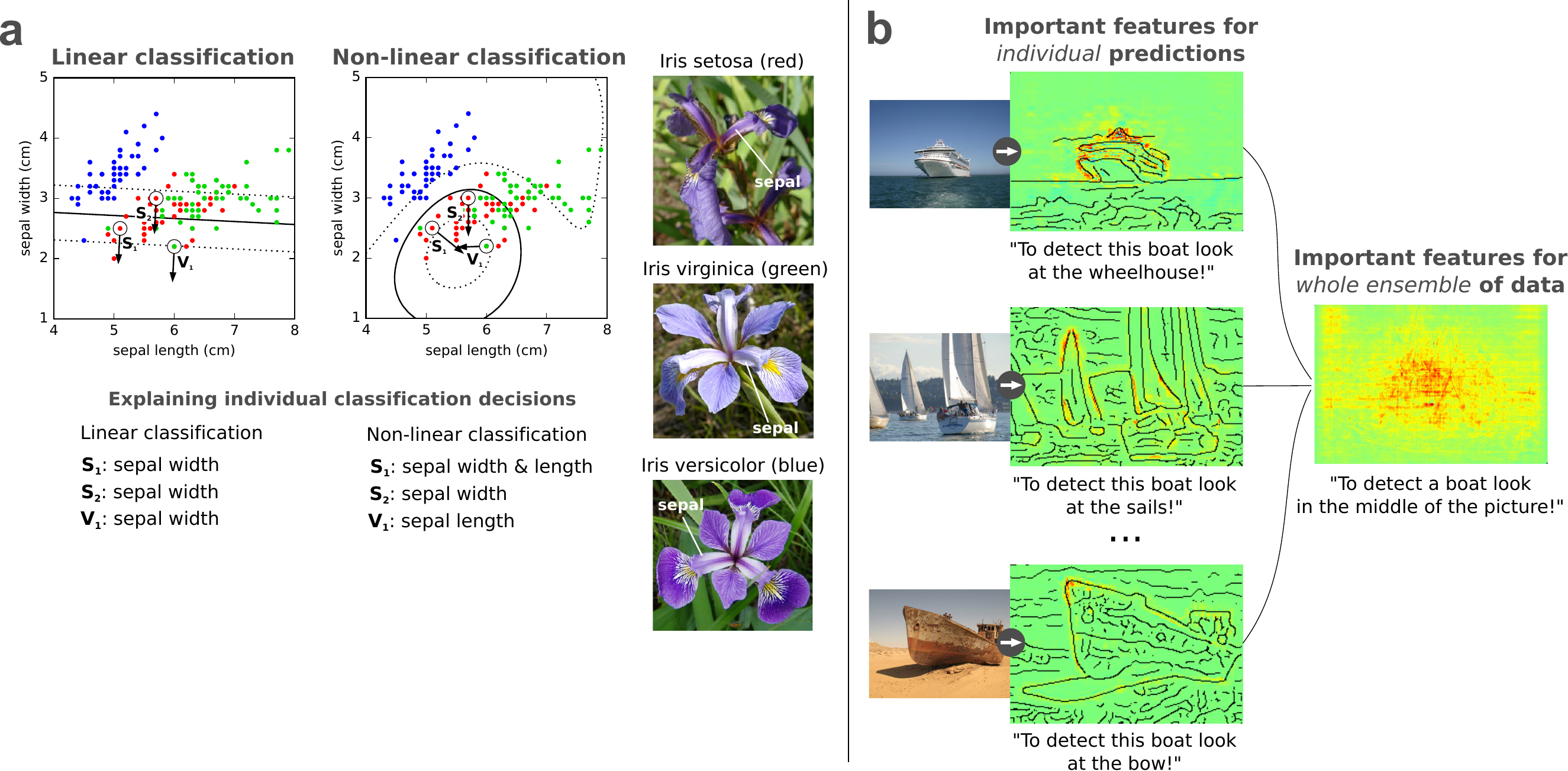}
\caption{Explanation of a linear and non-linear classifier. (\textbf{a}) In linear models the importance of each feature is the same for every data point. It can be expressed in the weight vector perpendicular to the decision
surface where more important features have larger weights. In nonlinear models the important features can be different for every data point. In this example the classifiers are trained to separate ``Iris setosa'' (red dots) from ``Iris virginica'' (green dots) and ``Iris versicolor'' (blue dots). The linear model for all examples uses the sepal width as discriminative feature, whereas the non-linear classifier uses different combinations of sepal width and  sepal length for every data point. (\textbf{b}) Different features can be important (here for a deep neural network) to detect a ship in an image. For some ships, the wheelhouse is a good indicator for class ``ship'', for others the sails or the bow is more important. Therefore individual predictions exhibit very different heatmaps (showing the most relevant locations for the predictor). In feature selection, one identifies salient features for the whole ensemble of training data. For ships (in contrast to e.g.\ airplanes) the most salient region (average of individual heatmaps) is the center of the image.}
\label{fig:main1}
\end{figure}

While it is broadly accepted that the nonlinear machine learning (ML) methods being used as predictors to maximize some prediction accuracy, are effectively (with few exceptions such as shallow decision trees) black boxes; this intransparency regarding explanation and reasoning is preventing a wider usage of nonlinear prediction methods in the sciences (see Figure \ref{fig:main1}\textbf{a} why understanding nonlinear learning machines is difficult). Due to this black-box character, a scientist may not be able to extract deep insights about what the nonlinear system has learned, despite the urge to unveil the underlying natural structures. In particular, the conclusion in many scientific fields has so far been to prefer linear models \cite{DBLP:journals/bmcbi/MaSH07,DBLP:journals/ploscb/Devarajan08,10.1371/journal.pone.0029348,DBLP:journals/neuroimage/HaufeMGDHBB14} in order to rather gain insight (e.g.\ regression coefficients and correlations) even if this comes at the expense of predictivity.

Recently, impressive applications of machine learning in the context of complex games (Atari games \cite{mnih2015human,Mnih2013NIPS}, Go \cite{silver2016mastering,silver2017mastering, Silver1140}, Texas hold'em poker \cite{MorScience17}) have led to speculations about the advent of machine learning systems embodying true ``intelligence''. In this note we would like to argue that for validating and assessing machine behavior, independent of the application domain (sciences, games etc.), we need to go beyond predictive performance measures such as the
test set error towards understanding the AI system.

\begin{figure}[!t]
\centering
\includegraphics[width=0.9\textwidth]{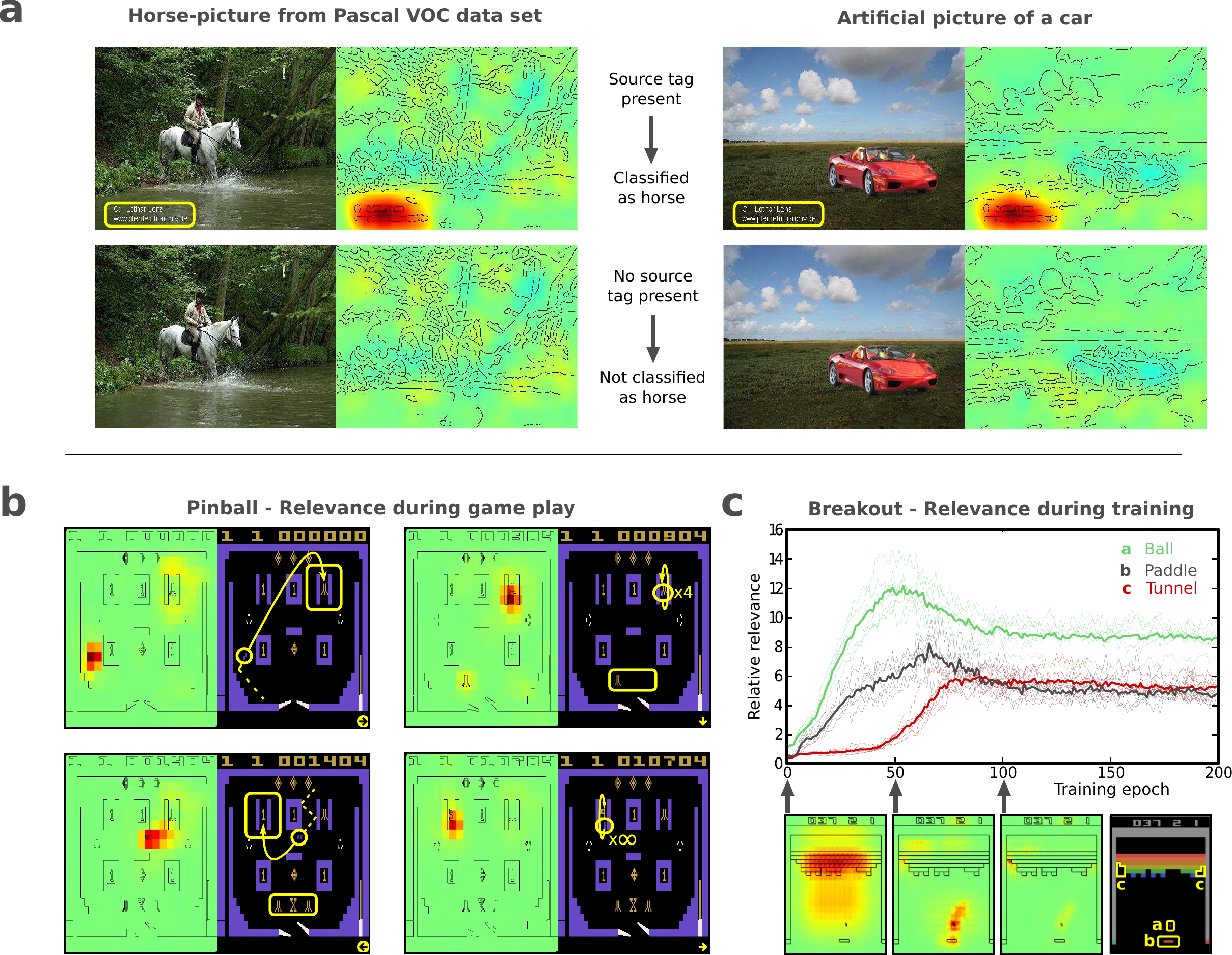}
\caption{Assessing problem-solving capabilities of learning machines using explanation methods. (\textbf{a}) The Fisher vector classifier trained on the PASCAL VOC 2007 data set focuses on a source tag present in about one fifth of the horse figures. Removing the tag also removes the ability to classify the picture as a horse. Furthermore, inserting the tag on a car image changes the classification from car to horse. (\textbf{b}) A neural network learned to play the Atari Pinball game. The model moves the pinball into a scoring switch four times to activate a multiplier (indicated as symbols marked in yellow box) and then maneuvers the ball to score infinitely. This is done purely by ``nudging the table'' and not by using the flippers. In fact, heatmaps show that the flippers are completely ignored by the model throughout the entire game, as they are not needed to control the movement of the ball. (\textbf{c}) Development of the relative relevance of different game objects in Atari Breakout over the training time. Relative relevance is the mean relevance of pixels belonging to the object (ball, paddle, tunnel) divided by the mean relevance of all pixels in the frame. Thin lines: 6 training runs. Thick line: average over the 6 runs.}
\label{fig:main2}
\end{figure}

When assessing machine behavior, the general task solving ability must be evaluated (e.g.\ by measuring the classification accuracy, or the total reward of a reinforcement learning system). At the same time it is important to comprehend the decision making process itself. In other words, transparency of the what and why in a decision of a nonlinear machine becomes very effective for the essential task of judging whether the learned strategy is valid and generalizable or whether the model has based its decision on a spurious correlation in the training data (see Figure \ref{fig:main2}\textbf{a}). In psychology the reliance on such spurious correlations is typically referred to as the Clever Hans phenomenon \cite{pfungst1911clever}. A model implementing a `Clever Hans' type decision strategy will likely fail to provide correct classification and thereby usefulness once it is deployed in the real world, where spurious or artifactual correlations may not be present.

While feature selection has traditionally explained the model by identifying features relevant for the whole ensemble of training data \cite{DBLP:journals/jmlr/GuyonE03} or some class prototype \cite{DBLP:journals/corr/SimonyanVZ13, yosinski2015understanding, nguyen2016multifaceted, nguyen2016synthesizing}, it is often necessary, especially for nonlinear models, to focus the explanation on the predictions of individual examples (see Figure \ref{fig:main1}\textbf{b}). A recent series of work \cite{DBLP:conf/eccv/ZeilerF14,DBLP:journals/corr/SimonyanVZ13,10.1371/journal.pone.0130140,DBLP:journals/jmlr/BaehrensSHKHM10,DBLP:conf/kdd/Ribeiro0G16, DBLP:conf/cvpr/ZhouKLOT16, MonPR17} has now begun to explain the predictions of nonlinear machine learning methods in a wide set of complex real-world problems (e.g.\ \cite{StuJNM16, DBLP:journals/corr/abs-1711-00138, DBLP:conf/icml/ZahavyBM16, ArrPLOS17}).
Individual explanations can take a variety of forms: An ideal (and so far not available) comprehensive explanation would extract the whole causal chain from input to output. In most works, reduced forms of explanation are considered, typically, collection of scores indicating the importance of each input pixel/feature for the prediction (note that computing an explanation does not require to understand neurons individually). These scores can be rendered as visual heatmaps (relevance maps) that can be interpreted by the user.

In the following we make use of this recent body of work, in particular, the layer-wise relevance propagation (LRP) method \cite{10.1371/journal.pone.0130140} (cf.~Section \ref{section:lrp}), and discuss qualitatively and quantitatively, for showcase scenarios, the effectiveness of explaining decisions for judging whether learning machines exhibit valid and useful problem solving abilities. Explaining decisions provides an easily interpretable and computationally efficient way of assessing the classification behavior from few examples (cf.\ Figure \ref{fig:main2}\textbf{a}). It can be used as a complement or practical alternative to a more comprehensive Turing test \cite{turing1950mind} or other theoretical measures of machine intelligence \cite{turing2009computing, legg2007universal, hernandez2017evaluation}. In addition, the present work contributes by further embedding these explanation methods into our framework SpRAy (spectral relevance analysis) that we present in Section \ref{section:methods}. SpRAy, on the basis of heatmaps, identifies in a semi-automated manner a wide spectrum of learned decision behaviors and thus helps to detect the unexpected or undesirable ones. This allows one to systematically investigate the classifier behavior on whole large-scale datasets --- an analysis which would otherwise be practically unfeasible with the human tightly in the loop. Our semi-automated decision anomaly detector thus addresses the last mile of explanation by providing an end-to-end method to evaluate ML models beyond test set accuracy or reward metrics.

\begin{figure}[!t]
\centering
\includegraphics[width=0.9\textwidth]{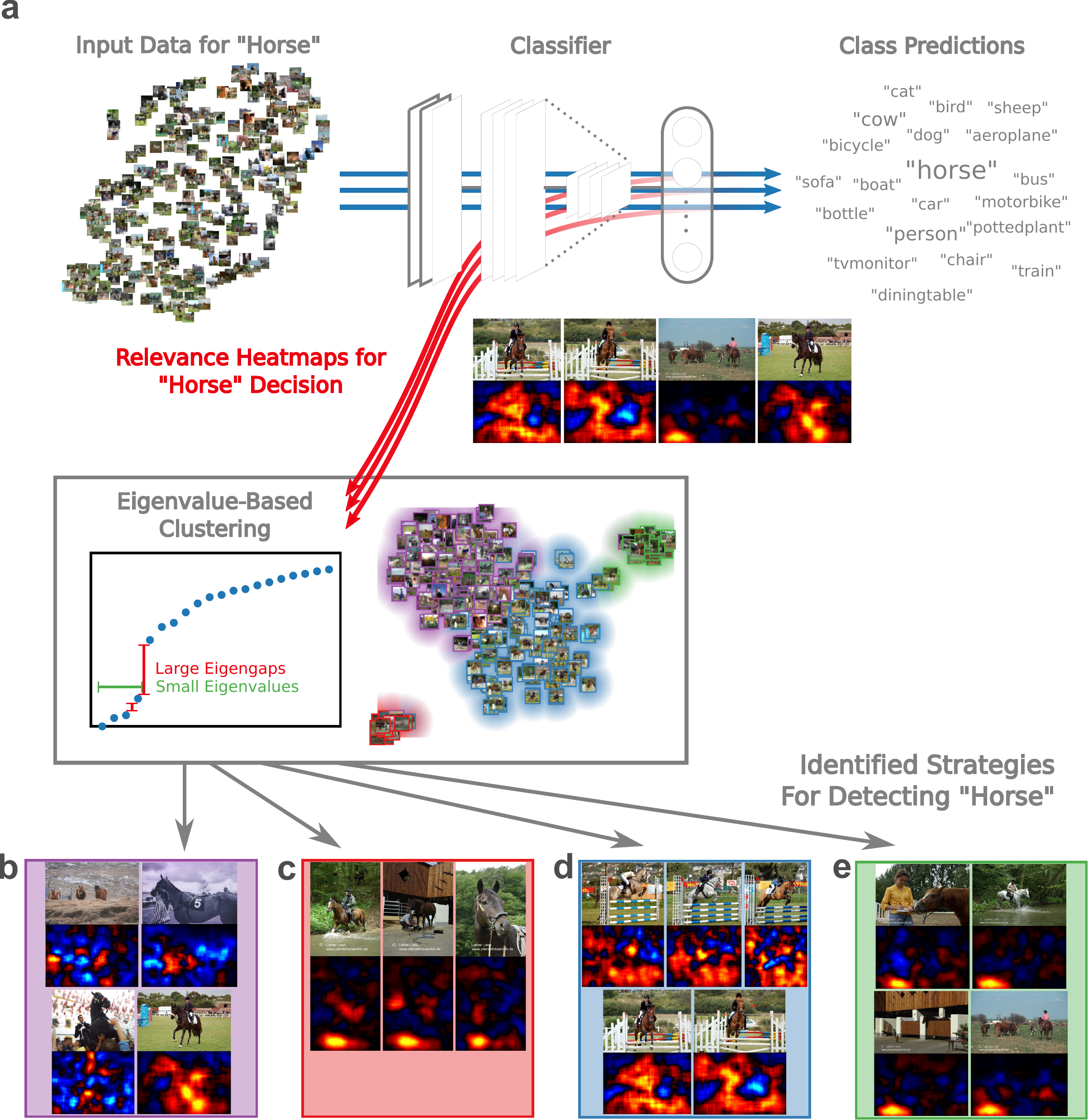}
\caption{The workflow of Spectral Relevance Analysis. (\textbf{a}) First, relevance maps are computed for data samples and object classes of interest, which requires a forward and a LRP backward pass through the model (here a Fisher vector classifier). Then, an eigenvalue-based spectral cluster analysis is performed to identify different prediction strategies within the analyzed data. Visualizations of the clustered relevance maps and cluster groupings supported by t-SNE inform about the valid or anomalous nature of the prediction strategies. This information can be used to improve the model or the dataset. Four different prediction strategies can be identified for classifying images as ``horse'': (\textbf{b}) detect a horse (and rider), (\textbf{c}) detect a source tag in portrait oriented images, (\textbf{d}) detect wooden hurdles and other contextual elements of horseback riding, and (\textbf{e}) detect a source tag in landscape oriented images.}
\label{fig:main4}
\end{figure}

\section{Results}
\subsection{Identifying valid and invalid problem-solving behaviors}
In this section we will investigate several showcases that demonstrate the effectiveness of explanation methods like LRP and SpRAy for understanding and validating the behavior of a learned model.

First, we provide an example where the learning machine exploits an unexpected spurious correlation in the data to exhibit what humans would refer to as ``cheating''. The first learning machine is a model based on Fisher vectors (FV) \cite{perronnin2010improving,sanchez2013image} trained on the PASCAL VOC 2007  image dataset \cite{everingham2010pascal} (see Section \ref{sec:pascal}). The model and also its competitor, a pretrained Deep Neural Network (DNN) that we fine-tune on PASCAL VOC, show both excellent state-of-the-art test set accuracy on categories such as `person', `train', `car', or `horse' of this benchmark (see Table \ref{tab:cvprresults}). Inspecting the basis of the decisions with LRP, however, reveals for certain images substantial divergence, as the heatmaps exhibiting the reasons for the respective classification could not be more different. Clearly, the DNN's heatmap points at the horse and rider as the most relevant features (see Figure \ref{fig:CVPR}). In contrast, FV's heatmap is most focused onto the lower left corner of the image, which contains a source tag. A closer inspection of the data set (of 9963 samples \cite{everingham2010pascal}) that typically humans never look through exhaustively, shows that such source tags appear distinctively on horse images; a striking artifact of the dataset that so far had gone unnoticed \cite{lapuschkinCVPR16}. Therefore, the FV model has `overfitted' the PASCAL VOC dataset by relying mainly on the easily identifiable source tag, which incidentally correlates with the true features, a clear case of `Clever Hans' behavior. This is confirmed by observing that artificially cutting the source tag from horse images significantly weakens the FV model's decision while the decision of the DNN stays virtually unchanged (see Figure \ref{fig:CVPR}). If we take instead a correctly classified image of a Ferrari and then add to it a source tag, we observe that the FV's prediction swiftly changes from `car' to `horse' (cf.\ Figure \ref{fig:main2}\textbf{a}) a clearly invalid decision (see Section \ref{sec:pascal} and Figures \ref{fig:aeroplane}-\ref{fig:coocurrence}  for further examples and analyses).

The second showcase example studies neural network models (see Figure \ref{fig:atariNetwork} for the network architecture) trained to play Atari games, here Pinball. As shown in \cite{mnih2015human}, the DNN achieves excellent results beyond human performance. Like for the previous example, we construct LRP heatmaps to visualize the DNN's decision behavior in terms of pixels of the pinball game. Interestingly, after extensive training, the heatmaps become focused on few pixels representing high-scoring switches and loose track of the flippers. A subsequent inspection of the games in which these particular LRP heatmaps occur, reveals that DNN agent firstly moves the ball into the vicinity of a high-scoring switch without using the flippers at all, then, secondly, ``nudges'' the virtual pinball table such that the ball infinitely triggers the switch by passing over it back and forth, without causing a tilt of the pinball table (see Figure \ref{fig:main2}\textbf{b} and Figure \ref{fig:games} for the heatmaps showing this point, and also Supplementary Video 1). Here, the model has learned to abuse the ``nudging'' threshold implemented through the tilting mechanism in the Atari Pinball software. From a pure game scoring perspective, it is indeed a rational choice to exploit any game mechanism that is available. In a real pinball game, however, the player would go likely bust since the pinball machinery is programmed to tilt after a few strong movements of the whole physical machine.

The above cases exemplify our point, that even though test set error may be very low (or game scores very high), the reason for it may be due to what humans would consider as cheating rather than valid problem-solving behavior. It may not correspond to true performance when the latter is measured in a real-world environment, or when other criteria (e.g.\ social norms which penalize such behavior \cite{DBLP:conf/iros/ChenELH17}) are incorporated into the evaluation metric. Therefore, explanations computed by LRP have been instrumental in identifying this fine difference.

Let us consider a third example where we can beautifully observe learning of strategic behavior: A Deep Neural Network playing the Atari game of Breakout \cite{mnih2015human} (see Table \ref{tab:archCompTable} for the investigated network architectures). We analyze the learning progress and inspect the heatmaps of a sequence of DNN models in Figure \ref{fig:main2}\textbf{c}. The heatmaps reveal conspicuous structural changes during the learning process. In the first learning phase the DNN focuses on ball control, the handle becomes salient as it learns to target the ball and in the final learning phase the DNN focuses on the corners of the playing field (see Figure \ref{fig:main2}\textbf{c}). At this stage, the machine has learned to dig tunnels at the corners (also observed in \cite{mnih2015human}) -- a very efficient strategy also used by human players. Detailed analyses using the heatmap as a function of a single game and comparison of LRP to sensitivity analysis explanations, can be found in the Figures \ref{fig:breakoutFailsToRestart}-\ref{fig:gradPulse} and in the Supplementary Video 2. Here, this objectively measurable advancement clearly indicates the unfolding of strategic behavior.

Overall, while in each scenario, reward maximization as well as incorporating a certain degree of prior knowledge has done the essential part of inducing complex behavior, our analysis has made explicit that (1) some of these behaviors incorporate strategy, (2) some of these behaviors may be human-like or not human-like, and (3) in some case, the behaviors could even be considered as deficient and not acceptable, when considering how they will perform once deployed. Specifically, the FV-based image classifier is likely to not detect horses on the real-world data; and the Atari Pinball AI might perform well for some time, until the game is updated to prevent excessive nudging.

All insights about the classifier behavior obtained up to this point of this study require the analysis of individual heatmaps by human experts, a laborious and costly process which does not scale well.

\subsection{Whole-dataset analysis of classification behavior}
Our next experiment uses SpRAy to comprehend the predicting behavior of the classifier on large datasets in a semi-automated manner. Figure \ref{fig:main4}\textbf{a} displays the results of the SpRAy analysis when applied to the horse images of the PASCAL VOC dataset (see also Figures \ref{fig:eigenvaluespectrum-horse} and \ref{fig:tsne+labels-horse}). Four different strategies can be identified for classifying images as ``horse'': 1) detect a horse and rider (Figure~\ref{fig:main4}\textbf{b}), 2) detect a source tag in portrait oriented images (Figure~\ref{fig:main4}\textbf{c}), 3) detect wooden hurdles and other contextual elements of horseback riding (Figure~\ref{fig:main4}\textbf{d}), and 4) detect a source tag in landscape oriented images (Figure~\ref{fig:main4}\textbf{e}). Thus, without any human interaction, SpRAy provides a summary of what strategies the classifier is actually implementing to classify horse images. An overview of the FV and DNN strategies for the other classes and for the Atari Pinball and Breakout game can be found in Figures \ref{fig:eigenval-barchart}-\ref{fig:tsne+labels-aeroplane} and \ref{fig:eigenvalue-spectra-atari}-\ref{fig:tsne+labels-pinball}, respectively.

The SpRAy analysis could furthermore reveal another `Clever Hans' type behavior in our fine-tuned DNN model, which had gone unnoticed in previous manual analysis of the relevance maps. The large eigengaps in the eigenvalue spectrum of the DNN heatmaps for class ``aeroplane'' indicate that the model uses very distinct strategies for classifying aeroplane images (see Figure \ref{fig:eigenval-barchart}). A t-SNE visualization (Figure \ref{fig:tsne+labels-aeroplane}) further highlights this cluster structure. One unexpected strategy we could discover with the help of SpRAy is to identify aeroplane images by looking at the artificial padding pattern at the image borders, which for aeroplane images predominantly consists of uniform and structureless blue background. Note that padding is typically introduced for technical reasons (the DNN model only accepts square shaped inputs), but unexpectedly (and unwantedly) the padding pattern became part of the model's strategy to classify aeroplane images. Subsequently we observe that changing the manner in which padding is performed has a strong effect on the output of the DNN classifier (see Figures \ref{fig:aeroplane-vs-none-border-artefact-barplots}-\ref{fig:vert-vs-horz-border-artefact-barplots}).

We note that while recent methods (e.g.\ \cite{rajalingham2018large}) have characterized whole-dataset classification behavior based on decision similarity (e.g.\ cross-validation based AP scores or recall), the SpRAy method can pinpoint divergent classifier behavior even when the predictions look the same. The specificity of SpRAy over previous approaches is thus its ability to ground predictions to input features, where classification behavior can be more finely characterized. A comparison of both approaches is given in Section \ref{sec:pascal} and Supplementary Figures \ref{fig:dicarlo-vs-spray} and \ref{fig:hurdles}.

\section{Discussion}
Although learning machines have become increasingly successful, they often behave very differently from humans \cite{lake2016building, tsividis2017human}. Commonly discussed ingredients to make learning machines act more human-like are e.g.\ compositionality, causality, learning to learn \cite{winston1975psychology,smith2002object,lake2015human}, and also an efficient usage of prior knowledge or invariance structure (see e.g.\ \cite{anselmi2016unsupervised,chmiela2017machine,Chmiela2018}). Our work adds a dimension that has so far not been a major focus of the machine intelligence discourse, but that is instrumental in verifying the correct behavior of these models, namely explaining the decision making.  We showcase the behavior of learning machines for two application fields: computer vision and arcade gaming (Atari), where we explain the strategies embodied by the respective learning machines. We find a surprisingly rich spectrum of behaviors ranging from strategic decision making (Atari Breakout) to `Clever Hans' strategies or undesired behaviors, here, exploiting a dataset artifact (tags in horse images), a game loophole (nudging in Atari Pinball), and a training artifact (image padding). These different behaviors go unnoticed by common evaluation metrics, which puts a question mark to the current broad and sometimes rather unreflected usage of machine learning in all application domains in industry and in the sciences.

With the SpRAy method we have proposed a tool to systematize the investigation of classifier behavior and identify the broad spectrum of prediction strategies. The SpRAy analysis is scalable and can be applied to large datasets in a semi-automated manner. We have demonstrated that SpRAy easily finds the misuse of the source tag in horse images, moreover and unexpectedly, it has also pointed us at a padding artifact appearing in the final fine-tuning phase of the DNN training. This artifact resisted a manual inspection of heatmaps of all 20 PASCAL VOC classes, and was only later revealed by our SpRAy analysis. This demonstrates the power of an automated, large-scale model analysis. We believe that such analysis is a first step towards confirming important desiderata of AI systems such as trustworthiness, fairness and accountability in the future, e.g.\ in context of regulations concerning models and methods of artificial intelligence, as via the General Data Protection Regulation (GDPR)~\cite{GDPR2016, goodman2017bryce}. Our contribution may also add a further perspective that could in the future enrich the ongoing discussion, whether machines are truly ``intelligent''.

Finally, in this paper we posit that the ability to explain decisions made by learning machines allows us to judge and gain a deeper understanding of whether or not the machine is embodying a particular strategic decision making. Without this understanding we can merely monitor behavior and apply performance measures without possibility to reason deeper about the underlying learned representation. The insights obtained in this pursuit may be highly useful when striving for better learning machines and insights (e.g.\ \cite{schutt2017quantum}) when applying machine learning in the sciences.

\section{Methods}
\label{section:methods}

\subsection{Layer-wise relevance propagation}
\label{section:lrp}
Layer-wise relevance propagation (LRP) \cite{10.1371/journal.pone.0130140} is a method for explaining the predictions of a broad class of ML models, including state-of-the-art neural networks and kernel machines. It has been extensively applied and validated on numerous applications \cite{ArrPLOS17, lapuschkinCVPR16, StuJNM16, HorArXiv18, yang2018explaining, ThoArXiv18}. The LRP method decomposes the output of the nonlinear decision function in terms of the input variables, forming a vector of input features scores that constitutes our `explanation'. Denoting $\x = (x_1,\dots,x_d)$ an input vector and $f(\x)$ the prediction at the output of the network, LRP produces a decomposition $\R = (R_1,\dots,R_d)$ of that prediction on the input variables satisfying
\begin{align}
\textstyle \sum_{p=1}^d R_p = f(\x).
\label{eq:decomposition}
\end{align}

Unlike sensitivity analysis methods \cite{Gevrey2003249}, LRP explains the output of the function rather than its local variation \cite{montavon2017methods} (see Section \ref{sec:understanding} for more information on explanation methods). 

The LRP method is based on a backward propagation mechanism applying uniformly to all neurons in the network: Let $a_j = \rho(\sum_i a_i w_{ij} + b_j)$ be one such neuron. Let $i$ and $j$ denote the neuron indices at consecutive layers, and $\sum_i$,$\sum_j$ the summation over all neurons in these respective layers. The propagation mechanism of LRP is defined as
\begin{align}
R_i = \sum_j \frac{z_{ij}}{\sum_i z_{ij}} R_j.
\label{eq:lrprule}
\end{align}
where $z_{ij}$ is the contribution of neuron $i$ to the activation $a_j$, and typically depends on the activation $a_i$ and the weight $w_{ij}$. The propagation rule is applied in a backward pass starting from the neural network output $f(\x)$ until the input variables (e.g.\ pixels) are reached. Resulting scores can be visualized as a heatmap of same dimensions as the input (see Figure \ref{figure:lrp}).

LRP can be embedded in the theoretical framework of deep Taylor decomposition \cite{MonPR17}, where some of the propagation rules can be seen as particular instances. Note that LRP rules have also been designed for models other than neural networks, in particular, Bag of Words classifiers, Fisher vector models, and LSTMs (more information can be found in the Section \ref{sec:understanding} and  Table \ref{tab:LRP}).

\subsection{Spectral relevance analysis}
\label{section:systematic}
Explanation techniques enable inspection of the decision process on a single instance basis. However, screening through a large number of individual explanations can be time consuming. To efficiently investigate classifier behavior on large datasets, we propose a technique: Spectral Relevance Analysis (SpRAy). SpRAy applies spectral clustering \cite{von2007tutorial} on a dataset of LRP explanations in order to identify typical as well as atypical decision behaviors of the machine learning model, and presents them to the user in a concise and interpretable manner.
 
Technically, SpRAy allows one to detect prediction strategies as identifiable on frequently reoccurring patterns in the heatmaps, e.g., specific image features. The identified features may be truly meaningful representatives of the object class of interest, or they may be co-occurring features learned by the model but not intended to be part of the class, and ultimately of the model's decision process. Since SpRAy can be efficiently applied to a whole large-scale dataset, it helps to obtain a more complete picture of the classifier behavior and reveal unexpected or `Clever Hans' type decision making.

The SpRAy analysis is depicted in Figure \ref{fig:main4} (see also Section \ref{sec:analysis}) and consists of four steps: 
Step 1: Computation of the relevance maps for the samples of interest. The relevance maps are computed with LRP and contain information about where the classifier is focusing on when classifying the images. 
Step 2: Downsizing of the relevance maps and make them uniform in shape and size. This reduction of dimensionality accelerates the subsequent analysis, and also makes it statistically more tractable.
Step 3: Spectral cluster analysis (SC) on the relevance maps. This step finds structure in the distribution of relevance maps, more specifically it groups classifier behaviors into finitely many clusters (see Figure \ref{fig:eigentutorial} for an example).
Step 4: Identification of interesting clusters by eigengap analysis. The eigenvalue spectrum of SC encodes information about the cluster structure of the relevance maps. A strong increase in the difference between two successive eigenvalues (eigengap) indicates well-separated clusters, including atypical classification strategies. The few detected clusters are then presented to the user for inspection.
Step 5 (Optional): Visualization by t-Stochastic Neighborhood Embedding (t-SNE). This last step is not part of the analysis strictly speaking, but we use it in the paper in order to visualize how SpRAy works.

Since SpRAy aims to investigate classifier behavior, it is applied to the heatmaps and not to the raw images (see Figures \ref{fig:tsne+labels-horse}, \ref{fig:tsne+labels-boat} and \ref{fig:tsne+labels-aeroplane} for comparison).

\subsection*{Code availability}
Source code for LRP and sensitivity analysis is available at \url{https://github.com/sebastian-lapuschkin/lrp_toolbox}.
Source code for Spectral Clustering and t-SNE as used in the SpRAy method is available from the scikit-learn at \url{https://github.com/scikit-learn/scikit-learn}.
Source code for the Reinforcement-Learning-based Atari Agent is available at \url{https://github.com/spragunr/deep_q_rl}.
Source code for the Fisher Vector classifier is available at \url{http://www.robots.ox.ac.uk/~vgg/software/enceval_toolkit}.\\
Our fine-tuned DNN model can be found at\\ 
\url{https://github.com/BVLC/caffe/wiki/Model-Zoo#pascal-voc-2012-multilabel-classification-model}.

\subsection*{Data availability} The datasets used and analyzed during the current study are available from the following sources.\\ 
PASCAL VOC 2007: \url{http://host.robots.ox.ac.uk/pascal/VOC/voc2007/#devkit}.\\ 
PASCAL VOC 2012: \url{http://host.robots.ox.ac.uk/pascal/VOC/voc2012/#devkit}.\\ 
Atari emulator: \url{https://github.com/mgbellemare/Arcade-Learning-Environment}.

\section*{Appendix}
\appendix
\section{Introduction}
AI systems are able to solve an increasing number of complex tasks. They occasionally exceed human performance spectacularly in tasks as diverse as face recognition\cite{lu2014surpassing}, traffic sign classification \cite{cirecsan2011committee}, reading subway plans \cite{graves2016hybrid}, understanding quantum many-body systems \cite{schutt2017quantum, chmiela2017machine, brockherde2017bypassing, Chmiela2018} or playing games such as Atari 2600 games \cite{Mnih2013NIPS,mnih2015human}, Go \cite{silver2016mastering, silver2017mastering}, Texas hold'em poker \cite{MorScience17} or Super Smash Bros.\ \cite{firoiu2017beating}. 
This impressive progress is largely owed to recent advances in {\it deep learning}, i.e., a class of end-to-end trainable, brain-inspired models with multiple (deep) layers of unfolding representations. Not surprisingly, these high-performance methods have quickly found their way out of research labs and are attracting much attention in the industry and the media. 
While these algorithms seem to clearly exhibit outstanding performance in the respective tasks, there is an ongoing debate on ``intelligence'' in AI systems in general \cite{turing1948intelligent, turing2009computing, mccorduck2009machines, lake2016building, bostrom2017superintelligence}, how to measure it and what could be the limits thereof. Amongst others this discussion aims to elucidate the defining ingredients for truly human-like learning such as compositionality, representation and how to include prior knowledge about the world \cite{lake2016building}. With this work we would like to add interpretability to this discussion as it is instrumental in judging and validating the behavior of these systems.

Let us first briefly roll out aspects of the current discussion. The authors of \cite{lake2016building, tsividis2017human} analyze key aspects of human-like learning. 
A major difference between machine and human learning is that humans (even infants) are equipped with a ``start-up'' software \cite{wellman1992cognitive, wellman1998knowledge} consisting of a priori representation about objects and physics \cite{spelke1990principles, baillargeon2004infants} and about other agents \cite{spelke2007core, csibra2003one}. This enables them to quickly learn to interact with the environment, to reason about future world states (e.g, position of objects in space) and to understand other people's goals, beliefs and strategies. Since such priors or also the knowledge about invariances \cite{poggio2016visual} are usually not available to a machine unless coded (see e.g.\ \cite{poggio2016visual,chmiela2017machine}), learning them takes significantly longer, requires orders of magnitude more examples and the learned concepts are less transferable to other tasks. 
Recently, Lake et al.\ \cite{lake2015human} proposed a learning machine that implements some of the human learning abilities and is able to learn new concepts from just a single example. The model brings together three important learning principles which have been intensively studied in cognitive science \cite{winston1975psychology, smith2002object} and are regarded as indispensable ingredients of human learning. These principles are (1) {\it compositionality}, i.e., the ability to build rich concepts from simpler primitives, (2) {\it causality}, i.e., the ability to model and infer the causal structure of the world, and (3) {\it learning to learn}, i.e., the ability to develop hierarchical priors which simplify learning new concepts from previous experience. Lake et al.\ \cite{lake2015human} incorporated these principles into a Bayesian program learning framework and showed on a restricted problem set, e.g., one-shot classification of handwritten characters, that their model achieves human-level performance and outperforms recent deep learning algorithms. Future generations of neural networks incorporating above principles may be able to approach human learning even in more realistic scenarios.

A further important aspect of human intelligence termed {\it computational rationality} was discussed in \cite{gershman2015computational}. Humans are adapted to interact with a dynamically changing, partially unknown and uncertain environment and thus are able to identify decisions with high expected utility while taking trade-offs in effort, precision, and timeliness of computations. Finding appropriate trade-offs and optimally allocating scarce resources is a difficult task which requires intelligence on a ``meta-level''. 
Recently, researchers started to bring these aspects to deep learning by incorporating mechanism such as attention \cite{xu2015show} or memory \cite{graves2016hybrid} into neural network models. Future AI systems that aim to interact with the real-world to a greater extent, may possibly have to fully implement computational rationality in order to be able to make meta-level decisions to better and more efficiently regulate base-level inferences.

A crucial aspect of human behavior, and the main focus of this paper, is the ability to {\it explain} \cite{lombrozo2009explanation, williams2010role}. It facilitates practical learning, e.g., in a student-teacher context, by not only providing the solution to a problem but also the description of how to solve the problem, what features to rely on etc. Furthermore, explanations have an important social role, because they enable to comprehend and rationalize other individuals' decisions, they also help establishing trust (e.g., doctor explains therapy to patient) and they are indispensable when the correctness of a decision needs to be verified. 
Until recently, deep neural networks and other complex, non-linear learning machines have been mainly used in a black-box manner, providing little information on what aspect of the input data supports the actual prediction for a single sample. This black-box behavior can amount to a major disadvantage and prevent the application of state-of-the-art AI technology in critical application domains. For instance, in medical diagnosis the ability to verify a decision made by an AI system is crucial for a medical professional as a wrong decision can cause threats to human life \cite{caruana2015intelligible}. Additionally, black-box systems are of limited value in the sciences, where it is crucial to ensure that the learned model is biologically, chemically or physically plausible or ideally contributes to a better understanding of the scientific problem \cite{schutt2017quantum}. In practice, if the learning machine is {\it interpretable}, we can visualize what it has learned, how it arrives at its conclusions and whether its task-solving strategy is meaningful, sensible and comprehensible from a human point of view. Also it can help confirming other important desiderata of AI systems such as fairness or accountability \cite{weller2017challenges, doshi2017roadmap, doshi2017accountability}. 

We will demonstrate in this work that state-of-the-art AI systems exhibit unexpected and to human standards not necessarily meaningful problem solution strategies. In particular, we would like to argue that this undesired behavior often goes unnoticed if AI models are not interpreted. Therefore, we will introduce in Section \ref{sec:understanding} recent techniques which allow to explain individual predictions of black-box AI models. Then, these techniques will be showcased in Section \ref{sec:agents} and \ref{sec:pascal}, by performing a detailed analysis of the behavior of Atari agents and image classifiers. Finally, we contribute in Section \ref{sec:analysis} a novel procedure for identifying erratic behavior of these systems in a semi-automated manner. Overall, our results demonstrate that interpretability provides a very effective way of assessing the validity and dependability of AI systems and we are convinced that it will also help us design more human-like AI systems in the future.

\section{Background}
We will now specifically focus on deep convolutional networks\cite{lecun1998gradient, krizhevsky2012imagenet}, which incorporate the principles of hierarchy and shift-invariance. They have scored commanding successes in the venues of strategic games \cite{silver2016mastering, mnih2015human, firoiu2017beating, DBLP:journals/corr/DosovitskiyK16}, image classification \cite{krizhevsky2012imagenet}, face \cite{DBLP:conf/cvpr/LeviH15, LapAMFG17} recognition, speech recognition \cite{hinton2012deep} and e.g.~physics \cite{schutt2017quantum,schutt2017moleculenet}. In the following we briefly introduce how neural networks work and how they can be combined with reinforcement learning to play arcade games.

\paragraph{Neural Networks}
Neural networks are a highly nonlinear, modular and hierarchical approach to learning\cite{lecun2012learning, lecun2015deep, bishop1995neural}. The advent of GPU accelerated training \cite{oh2004gpu} and the availability of performant deep learning frameworks~\cite{team2016theano, collobert2011torch7, jia2014caffe, abadi2016tensorflow,chen2015mxnet}, as well as large datasets \cite{russakovsky2015imagenet, everingham2010pascal, netzer2011reading, krizhevsky2009learning, garofolo1993darpa, miller2003movielens} lead to unprecedented accuracy in a variety of tasks\cite{silver2016mastering, stanley2002evolving, mnih2015human, krizhevsky2012imagenet, DBLP:conf/cvpr/LeviH15, BosTIP18}. Neural networks are composed of multiple layers, each consisting of a linear and a non-linear transformation,
\begin{equation}
 x^{(l+1)}_j = g\kl{\sum_j w^{(l)}_{j,k}x^{(l)}_k + b^{(l)}_j},
 \label{eq:nnlayer}
\end{equation}
where $g$ is a non-linear function, e.g.\ sigmoid function, hyperbolic tangent or a rectifier function\cite{glorot2011deep}. The universal approximation theorem ensures that any continuous function on a compact interval of $\mathbb{R}^n$ can be approximated by a neural network to an arbitrary precision\cite{hornik1989multilayer, funahashi1989approximate, cybenko1989approximation}.

Although only one hidden layer is strictly necessary, having a deeper structure allows this approximation to be more efficient in the number of neurons required \cite{bengio2009learning}. State-of-the-art deep neural networks achieve astounding results with many hidden layers, some numbering eight \cite{krizhevsky2012imagenet}, twenty-two \cite{szegedy2015going} or even more than one hundred \cite{he2016deep} layers. The deep structure allows the networks to operate on conceptual levels with each layer processing more complex features (e.g.\ edges $\rightarrow$ corners $\rightarrow$ rectangles $\rightarrow$ boxes $\rightarrow$ cars) mirroring abstraction levels similar to the human brain \cite{serre2007robust, goodfellow2016deep}. It can be shown that from layer to layer, networks compress information and enhance the signal-to-noise ratio (see \cite{montavon2011kernel, shwartz2017opening}).

There are three major learning paradigms for DNN's, each corresponding to a particular abstract learning task, i.e.~supervised learning, unsupervised learning and reinforcement learning. In this work we focus on supervised and reinforcement learning to train the agents and evaluate their decisions for a better understanding.

\paragraph{Supervised Learning}
In supervised learning the goal is to infer a function $f$ from a labeled training dataset consisting of the samples $x_i$ and their label $y_i$. If the labels only take discrete values, then we refer to the learning problem as {\it classification problem}, otherwise it is called a {\it regression problem}. Often $f$ belongs to a parameterized class of functions, e.g., neural networks with a particular architecture, then the learning task reduces to finding the ``optimal'' parameters (w.r.t.\ a specific loss). In practice, algorithms such as stochastic gradient descent \cite{bottou1991stochastic} and variants of it such as Adam \cite{kingma2014adam} are used for this optimization. After training we can use function $f$ to estimate the predictions for unseen samples (i.e., testing phase) in order to approximate its generalization ability. In Section \ref{sec:pascal} we will analyze and compare the generalization performance and task-solving strategy of two powerful systems for image categorization --- a typical supervised learning task.

\paragraph{Reinforcement Learning}
In a typical reinforcement learning setting, an agent is trained without explicitly labelled data, via interaction with the environment only. At each time step $t$ of the training process, the agent to be trained performs a preferred action $a_t$ based on a state $s_t$ and receives a reward $r_t$ with the goal to maximize the long term reward $\sum_t r_t$ by adapting the model parameters determining which action is to be chosen given an input.

Recently, a convolutional neural network was trained to play simple Atari video games with visual input on a human-like level\cite{mnih2015human}. Using Q-learning \cite{Watkins:1989QLearn}, a model-free reinforcement learning technique, a network has been trained to fit an action-value function. The Q-learning setup consists of three elements:
\begin{enumerate}
 \item \textbf{the Arcade Learning Environment \cite{bellemare13arcade}} that takes an action as input, advances the game and returns a visual representation of the game and a game score,
 \item \textbf{the Neural Network} that predicts a long-term reward, the so-called Q-function $Q(s,a;\boldsymbol\theta)$, for a game visual for every possible action $a$, and, lastly,
 \item \textbf{the Replay Memory} that saves observed game transitions during the training as tuples \texttt{(state, action, reward, next\_state)}, from now on $(s_t, a_t, r_t, s_{t+1})$.
\end{enumerate}

To update the network, Mnih et al.\ \cite{mnih2015human} make use of the fact that the optimal Q-function, $Q^*$ must obey the Bellman equation\cite{Bellman1952}
\begin{equation}
 Q^*(s_t,a_t) = \mathbb{E}_{s_{t+1}}\ekl{r_t + \gamma \max_{a_{t+1}} Q^*(s_{t+1},a_{t+1}) \ \Big|\ s_t,a_t},
 \label{eq:Bellmann}
\end{equation}
which can be used to train the network by choosing the cost function as the squared violation of the Bellman equation. The expectation value is approximated by the sum over a batch of game transitions $B$ drawn uniformly at random from the replay memory.
\[
 C(\boldsymbol\theta) = \sum_{\kl{s,a,r,s'}\in B} \kl{Q(s,a;\boldsymbol\theta) - \ekl{r + \gamma \max_{a'} Q(s',a';\boldsymbol\theta)}}^2.
\]
The agent is trained by alternating three steps. First, it explores its environment using the actions that promise the highest long-term reward according to its own estimations. Second, it records the observations it encounters and saves the game transitions in the replay memory, potentially replacing older transitions. Third, one trains the network to predict the long-term reward by drawing batches from the replay memory and updating the cost functional.

The interplay of environment, network and memory is a convoluted process where convergence is not guaranteed \cite{kaelbling1996reinforcement}. Methods to explain the network decisions can be used to chart the development of the neural agent even at stages where it is not yet able to successfully interact with the environment. As we will see later, this will be useful to correct the network architecture early on.

\section{Understanding AI Systems}
\label{sec:understanding}
Nonlinear learning methods such as neural networks are often (but falsely) considered as black boxes. 
One approach to assess the quality of these learning machines is to observe the model's behavior on an independent test dataset and from that draw conclusions about it's problem solving strategies. Although theoretically valid, this approach may be very cumbersome or even practically impossible for AI systems, because evaluating the responses of the system to all possible variations of the input requires huge amount of testing (curse of dimensionality). In practice test datasets are often small and not representative. More advanced approaches exist for measuring or defining the capabilities of AI systems (e.g., \cite{turing2009computing, legg2007universal, hernandez2010measuring, hernandez2017evaluation}), however, these tests rather focus on assessing the system's performance in solving various tasks than on understanding the decision process (in particular the single decision) itself.

In this work we argue that to fully understand the quality of a learning machine, its intrinsic nonlinear decision making process needs to be made accessible to human judgment. 
Understanding the basis of decisions is desirable for many reasons:
\begin{itemize}
\item[(1)] In cases where we know what the decision should be based on, we can judge the soundness (according to human standards) of the decision process. 
\item[(2)] If we do not know what the decision should be based on but we can trust the system, we may infer new domain knowledge. For example, Vidovic et al.\ \cite{vidovic2015svm2motif} use an accomplished SVM classifier that predicts protein splice sites from large gene sequences to also explain which base pair motifs in genes were decisive for the classification. This can potentially increase our understanding or at least narrow down interesting domains for further, more informed, research. 
\item[(3)] Even if we can not make sense of the explanations, e.g., because we are not domain experts, it is still possible to use this extra information for detection of erratic behavior of the AI system (see Section \ref{sec:analysis}). We provide an example of such erratic behavior in Section \ref{sec:pascal}, where we train a Fisher vector classifier which unintentionally bases its decision on a spurious correlation. For images showing horses, the model has learned to dominantly decide based on the presence of a copyright watermark on one of the corners of the images. Of course, in the world portrayed by an artifactual dataset it is completely valid to assume that horses are connected to the existence of a source tag. But from human perspective this is the Clever Hans phenomenon \cite{pfungst1911clever} well known in comparative psychology, i.e., the system uses a spurious correlation\footnote{The Orlov Trotter horse claimed to perform arithmetic and other intellectual tasks, but actually was watching the reactions of his trainer.} to solve the problem without understanding the problem. It is obvious that the learning system in our case does not truly understand the concept of a horse.
\end{itemize}

There is also an interesting phenomenon described in the literature, the {\it AI effect} \cite{mccorduck2009machines}, saying that everytime an AI system solves a problem which has been regarded as an intelligence task, e.g., playing checkers or chess, it is is not regarded as being ``intelligent'' after some time, but solving the problem is regarded as rather computation. Thus AI systems (e.g., alphaGo \cite{silver2016mastering}, DeepStack \cite{MorScience17} or subway plan reading system \cite{graves2016hybrid}) which are regarded intelligent today, may be not regarded as intelligent anymore in the near future.

\subsection{Explaining Classification Decisions}
In the context of image recognition, classification decisions can be explained by tracing the model decision down to the input pixels. The resulting explanation takes the form of an image of the same format as the input image, for which the content provides visualizable feedback on the classification decision of the learned model (see e.g.~\cite{montavon2017methods}). In particular, it will reveal which part of the image is important for classification, and more precisely, which pattern in the image causes the neural network to decide.

The first approaches to extract interpretable visual patterns from a classifier's decision were based on Sensitivity Analysis \cite{Gevrey2003249,DBLP:journals/jmlr/BaehrensSHKHM10,DBLP:journals/corr/SimonyanVZ13}. These methods rely on the gradient of the decision function and  identify input variables (e.g., pixels) which maximally strengthen or weaken the classifier decision signal when changed. Strictly speaking these methods do not explain the prediction (``what made the classifier arrive at it's decision''), but analyze the local variation of the classification function (see \cite{montavon2017methods} for more discussion). Other more recent techniques \cite{DBLP:conf/eccv/ZeilerF14, DBLP:conf/cvpr/FangGISDDGHMPZZ15, DBLP:journals/corr/SpringenbergDBR14,  DBLP:conf/eccv/MahendranV16, DBLP:conf/kdd/Ribeiro0G16, fong2017interpretable} extract visual patterns associated to the decision of deep convolutional neural networks by using different heuristic criteria (e.g., occlusion, perturbation, sampling or deconvolution). A principled analysis technique for explaining decisions of complex models such as deep neural networks or bag-of-words-type classifiers has emerged with Layer-wise Relevance Propagation (LRP) \cite{10.1371/journal.pone.0130140, MonPR17}. We will describe this technique below and use it throughout the experiments of this paper. A theoretical foundation of LRP, called deep Taylor decomposition, can be found in \cite{MonPR17}. Explanation methods were successfully applied to a wide set of complex real-world problems such as the analysis of faces \cite{DBLP:conf/dagm/ArbabzadahMMS16, LapAMFG17, SeiArXiv18}, EEG and fMRI data \cite{StuJNM16, ThoArXiv18}, human gait \cite{HorArXiv18}, videos \cite{AndArXiv18}, speech \cite{BecArXiv18}, reinforcement learning \cite{DBLP:conf/icml/ZahavyBM16}, biological data \cite{kraus16}, text \cite{DBLP:conf/naacl/LiCHJ16, ArrACL16, ArrPLOS17}, or comparing human and algorithm behavior in the context of visual question answering \cite{DBLP:conf/emnlp/DasAZPB16}. 

\subsection{Pixel-Wise Decompositions}
A natural way of explaining a classifier's decision is to decompose its output as a sum of pixel-wise scores representing the relevance of each pixel for the decision. More precisely, denoting by $R_f$ the output of the neural network, and $R_p$ the relevance of pixel $p$, one requires that the following conservation property holds
$$
\textstyle \sum_p R_p = R_f,
$$
where the sum runs over all pixels in the image. Several methods, including LRP, perform such decomposition \cite{DBLP:conf/aaai/PoulinESLGWFPMA06, DBLP:conf/cidm/LandeckerTBMKB13, 10.1371/journal.pone.0130140, MonPR17}. An advantage of the decomposition framework is versatility: If needed, one can directly recompute the analysis at a coarser level by aggregating relevance scores according to a partition of the pixel space \cite{SamTNNLS16}. For example, denoting by $\mathcal{P}$ a particular region of the image (e.g.\ its center), we can compute the total relevance for that region as
$$
\textstyle R_\mathcal{P} = \sum_{p \in \mathcal{P}} R_p.
$$
while still satisfying the coarser conservation property when summing over all regions of the partition: $\sum_\mathcal{P} R_\mathcal{P} = R_f$. This aggregation property is useful in our study to be able to produce height-aggregated heatmaps, that we can plot as a function of time, thus, allowing to monitor the evolution of the agent attention for the considered Atari games. The aggregation property is also used implicitly to sum relevance over the RGB channels of the pixels, and thus produce a single score per pixel.

Decomposition methods can be subdivided in three categories: (1) structure approaches, where the function itself has the particular summation structure, and where individual pixel-wise contributions can be identified as the summands of the function \cite{DBLP:conf/aaai/PoulinESLGWFPMA06}, (2) analytic approaches, where a local analysis of the function is performed, and where pixel-wise contributions can be identified as linear terms of some local function expansion \cite{DBLP:journals/corr/SimonyanVZ13, 10.1371/journal.pone.0130140}, (3) propagation approaches \cite{10.1371/journal.pone.0130140, MonPR17, DBLP:conf/cidm/LandeckerTBMKB13, DBLP:conf/eccv/ZhangLBSS16}, where the model is assumed to describe a directed computational graph, where the output score is propagated through the network under local conservation constraints until the pixels are reached.

\subsection{Layer-Wise Relevance Propagation}
\begin{figure}
\centering\includegraphics[width=0.85\textwidth]{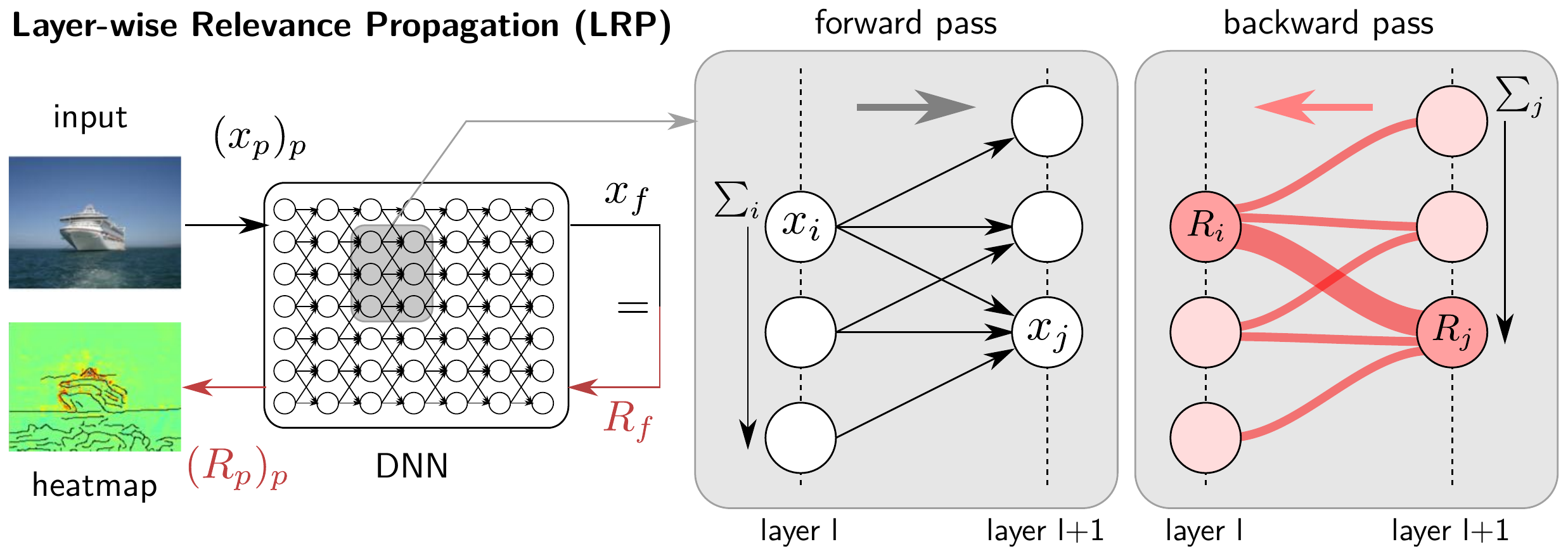}
\caption{\emph{Left}: Overview of the LRP method. The input image is propagated in a neural network, which classifies it as ``boat''. The classification score is backpropagated in the network, leading to individual pixel-wise relevance scores that can be visualized as a heatmap. \emph{Right}: Details of the forward and backward passes for a small portion of the neural network.}
\label{figure:lrp}
\end{figure}
We present here the Layer-wise Relevance Propagation (LRP) approach proposed by \cite{10.1371/journal.pone.0130140} for obtaining a pixel-wise decomposition of the model decision and use this technique throughout all experiments of the paper. LRP uses a propagation approach and is general enough to apply to most of the state-of-the-art architectures for image classification or neural reinforcement learning, including in particular AlexNet \cite{krizhevsky2012imagenet}, or Atari-based neural networks \cite{mnih2015human}, and can also be applied to other types of models such as Fisher vector with SVM classifiers or LSTMs \cite{hochreiter1997long,ArrWASSA17}. LRP assumes that the decision function can be decomposed as a feed-forward graph of neuron computations of type
$$
x_j = g \big({\textstyle \sum_i} x_i w_{ij} + b_j\big),
$$
where $g$ is some monotonically increasing nonlinear function (e.g.\ the ReLU nonlinearity), $x_i$ are the neuron inputs, $x_j$ is the neuron activation, and where $w_{ij}$ and $b_j$ are learned weights and bias parameters. The propagation behavior of LRP can be characterized by looking at a single neuron: The relevance $R_j$ received by neuron $j$ from the upper layers must be redistributed to its incoming neurons in the lower layer. For this, we produce messages $R_{i\leftarrow j}$ that satisfy a local conservation property:
$$
\textstyle \sum_i R_{i \leftarrow j} = R_j
$$
The exact definition of LRP rule is dependent on the neuron type, and its position in the architecture, however, the message can generally be written as
$$
R_{i \leftarrow j} = \frac{q_{ij}}{\sum_i q_{ij}} R_j,
$$
where $q_{ij}$ is a measure of contribution of neuron $i$ to activating neuron $j$. The relevance score assigned to the neuron $i$ is subsequently obtained by pooling all relevance messages coming from the higher-layer neurons to which neuron $i$ contributes:
$$
R_i = {\textstyle \sum_j} R_{i \leftarrow j} = \sum_j \frac{q_{ij}}{\sum_i q_{ij}} R_j
$$
The intuition behind this propagation method is that a neuron is defined as relevant if it contributes to neurons that are relevant themselves. The relevance propagation procedure is iteratively applied from the top layer down to the pixel layer, at which point the procedure stops. The method is illustrated in Figure \ref{figure:lrp}; for theoretical background see \cite{MonPR17}. LRP toolboxes are described in \cite{LapJMLR16, AlbArXiv18}.

A possible propagation rule results from defining contributions as the positive part of the product between the neuron input and the weights $q_{ij}~=~(x_i w_{ij})^+$ where $()^+$ indicates the positive part. $q_{ij}$ can be interpreted as the amount by which neuron $i$ excites neuron $j$. This rule was advocated by \cite{10.1371/journal.pone.0130140, MonPR17, DBLP:conf/eccv/ZhangLBSS16}, is easy to implement and particularly suitable for neural networks with ReLU nonlinearities. In this context, this rule is an instance of the more general $\alpha\beta$-rules proposed by \cite{10.1371/journal.pone.0130140}. In practice, the rule can be replaced by other rules (e.g.\ the $\alpha\beta$-rules, the $\epsilon$-rule, the $w^2$-rule, or the ``flat''-rule) all based on the same local relevance conservation principles, but with different characteristics such as sparsity, amount of negative evidence, or domain of applicability. An overview of these rules is given in Table \ref{tab:LRP}, where we use the shortcut notation $z_{ij} = x_i w_{ij}$ and where we define the map $\sigma : t \rightarrow t + \epsilon \cdot \text{sign}(t)$. Finally, max-pooling layers are treated in this paper by redirecting all relevance to the neuron in the pool that has the highest activation. The Caffe reference model \cite{jia2014caffe}, as used for image categorization in Section \ref{sec:pascal}, employs local renormalization layers. These layers are treated with an approach based on Taylor expansion \cite{BinICANN16}. Pixel-wise relevance is obtained by pooling relevance over the RGB components of each pixel.
\begin{table}[h]
\centering
 \caption{Formula and usage of various LRP-rules. The $\alpha\beta$-rule is used together with flat-weight- or $w^2$-method for the neural agents. The $\epsilon$-rule is used for the Fisher vectors (FV).}
\small
\begin{tabular}{|c|c|l|}\hline
 LRP rule 		& Formula & Used for \\ \hline\hline
 $\alpha\beta$-rule 	& $R_{i \leftarrow j} = \bigg.\Big( \alpha\frac{z^+_{ij}}{\sum_i z^+_{ij}} + \beta\frac{z^-_{ij}}{\sum_i z^-_{ij}} \Big) R_j$ &
 \parbox{5cm}{
 - DNN in \cite{10.1371/journal.pone.0130140,lapuschkinCVPR16,BacICIP16}\\
 - Atari agents.
 }\\\hline
 $w^2$-rule 		& $R_{i \leftarrow j} = \bigg.\frac{w_{ij}^2}{\sum_i w_{ij}^2} R_j$ &
 \parbox{5cm}{- Atari agents (bottom layers)}\\\hline\hline
  $\epsilon$-rule 	& $R_{i \leftarrow j} = \bigg.\frac{z_{ij}}{\sigma(\sum_i z_{ij})} R_j$ 	&
 \parbox{5cm}{- FV top layer in \cite{lapuschkinCVPR16,BacICIP16}}\\\hline
  flat rule 	& $R_{i \leftarrow j} = \bigg.\frac{1}{\sum_i 1} R_j$ &
  \parbox{5cm}{- FV local descriptors \cite{lapuschkinCVPR16,10.1371/journal.pone.0130140}\\
  - Atari agents (bottom layers)} \\\hline
\end{tabular}
 \label{tab:LRP}
 \end{table}

\subsection{Experiments}
We analyze the decision process of convolutional neural networks in two different machine intelligence tasks, reinforcement learning for arcade video games and supervised image classification. In the first task a convolutional neural network is trained to play simple computer games. We explain the decisions of the network using LRP and visualize the relevance for different game objects. We visualize which game objects the network focuses on and how this corresponds to the current game situation and the learned strategy for the games of \textit{Breakout} and \textit{Video Pinball} for the Atari 2600. Taking this approach further, we monitor the importance attributed by the learning machine to different game objects as a function of the training time and quantify changes in the model's behavior that are not apparent by monitoring the game score.

Second, we analyze machine behavior for a popular computer vision task; here two ML models are trained to classify images of the Pascal VOC dataset \cite{everingham2010pascal}. Interestingly and anecdotally, one of them, a Fisher vector-based classifier trained to classify images as horses or non-horses, is shown to rely its decisions on a spurious correlation: a copyright watermark that accidentally and undetected by the computer vision community persisted in this highly popular benchmark. For the classification of ships the classifier is mostly focused on the presence of water in the bottom half of an image. Removing the copyright tag or the background results in a drop of predictive capabilities. A deep neural network, pre-trained in the ImageNet dataset \cite{russakovsky2015imagenet}, instead shows none of these shortcomings.

\section{Task I: Playing Atari Games}
\label{sec:agents}
The first intelligence task inspected studies neural network agents playing simple Atari video games. The training of these models has been performed using a Python- and Theano-based implementation, which is publicly available from \texttt{https://github.com/spragunr/deep\_q\_rl} and implements the system described in \cite{mnih2015human}. The method uses a modification to Equation \ref{eq:Bellmann} to calculate the long-term reward of the next action and considers both the current set of model parameters $\theta$, as well as an older,
temporarily fixed version thereof $\theta^*$, where after every $k$ steps, $\theta^*$ will be updated to the values of $\theta$. Having a different set of parameters for the target and for the prediction stabilizes the training process. The resulting update step for the network parameters is
\begin{equation}
 \theta_{n+1} = \theta_n + \alpha~ \sum_{\mathclap{\kl{s,a,r,s'}\in B}}~ {\nabla_{\boldsymbol\theta} Q\kl{s,a;\boldsymbol\theta} D\kl{Q(s,a;\boldsymbol\theta) - \ekl{r + \gamma \max_{a'} Q(s',a';\boldsymbol\theta^*)}}},
 \label{eq:update}
\end{equation}
where $\alpha$ is the learning rate of the training process. Following the approach of \cite{mnih2015human}, the function $D$ clips the difference term at the end of Equation \ref{eq:update} between $[-1\ 1]$ to curb oscillations in updates where the network is far from satisfying the Bellman equation. This is equivalent to employing a quadratic error term until the value 1 and a constant error term beyond.

\begin{figure}
\centering\includegraphics[width=0.7\textwidth]{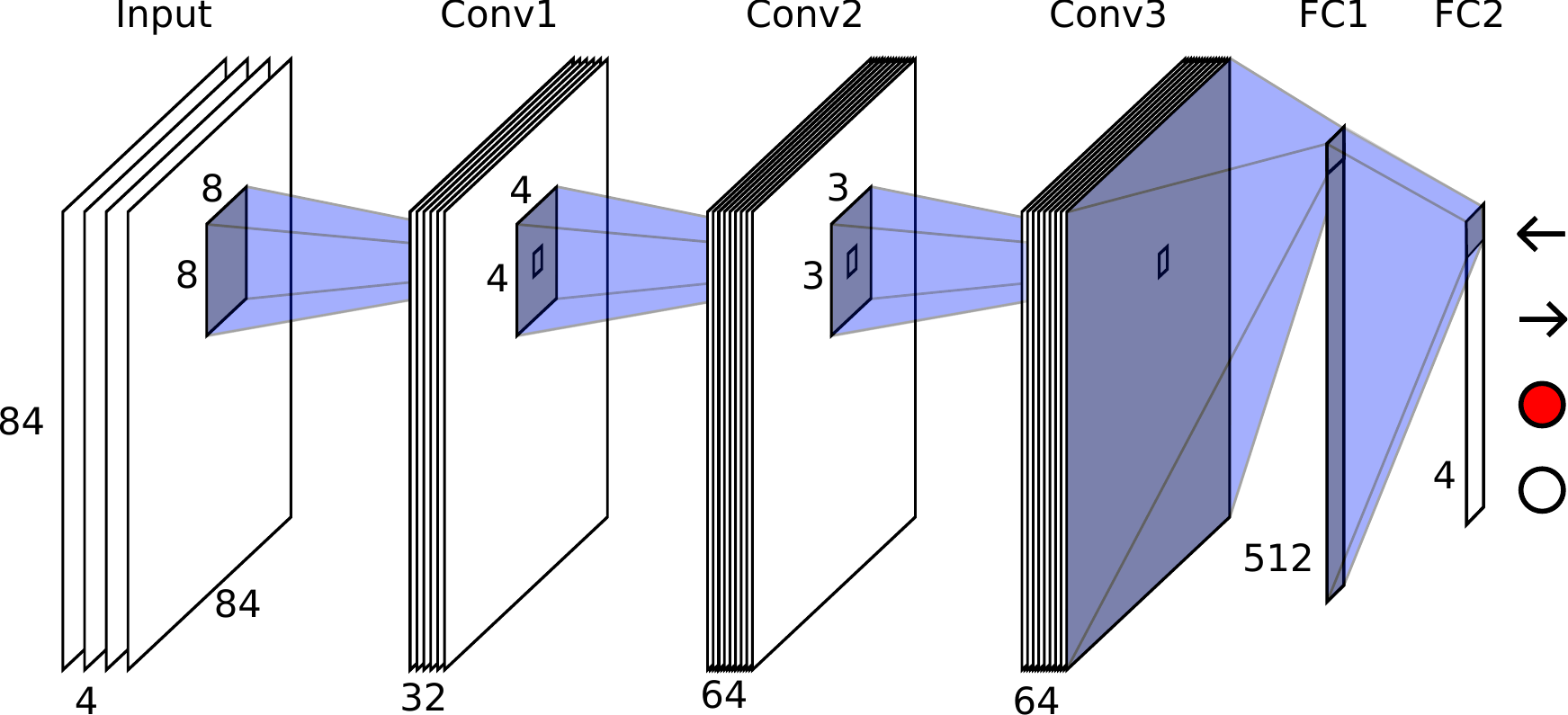}
\caption{The Atari Network architecture. The network consists of three convolutional layers (Conv) and two fully connected layers (FC). All layers comprise a ReLU-nonlinearity except for the last FC layer. The network output corresponds to a vector that predicts the Q-function for every possible action, in this example ``go left'', ``go right'', ``fire'' and ``do nothing''.}
\label{fig:atariNetwork}
\end{figure}

The network consists of three convolutional layers and two inner product layers. The exact architecture from \cite{mnih2015human} is described in Figure \ref{fig:atariNetwork} and in Table \ref{tab:archCompTable}. An input state corresponds to the last four frames of game visuals as seen by a human player, transformed into gray-scale brightness values and scaled to $84 \times 84$ pixels in size, is fed to the network as a $4 \times 84 \times 84$-sized tensor with pixel values rescaled between $0$ (black) and $1$ (white).
The network prediction is a vector of the expected long-term reward for each possible action, where the highest rated action is then passed as input to the game. Every action is repeated for four time steps (i.e.~every four frames the model receives a new input spanning four frames and makes a decision which is used as input for the next four frames) which corresponds to the typical amount of time it takes for a human player to press a button. With a probability of $10\%$, the trained agent will choose a random action instead of using the action predicted to be the most valuable option; as we will see in our analysis this randomness is essential. Upon training a completely fresh model, the probability of picking actions at random is initiated at $100\%$ and linearly over time lowered to $10\%$ with ongoing training.
As suggested in \cite{mnih2015human}, we use stochastic gradient descent with a batch size of 32 and a learning rate of $\alpha = 2.5\cdot10^{-4}$. Every fourth update step, the parameters from $\theta$ are copied to $\theta^*$. During training, a replay memory is maintained as an active training set. The memory is updated as a FIFO-style buffer structure, as soon as its capacity of $10^6$ time steps or observations is exhausted. Model weight updates are performed as soon as $5 \cdot 10^4$ observations have been collected.

We use LRP to explain how the network decides which action to take. Specifically, we use the $\alpha\beta$-rule, for LRP (see Table \ref{tab:LRP}) with parameters $\alpha=1$ and $\beta=0$. It distributes relevance proportionally to the positive forward contribution in every layer. In the input layer, the black background (black) of the game corresponds to a gray-scale value of zero and would receive no relevance. To be able to distribute relevance onto all pixels, in the lowest layer, we resort to the $w^2$-rule described in \cite{MonPR17} which was designed to appropriately address this situation.

\subsection{Visualization and Comparison to Game Play}
\label{sec:vis_and_comparison_gameplay}
LRP is used to back-propagate the dominant action through the network onto the input which results in a $4 \times 84 \times 84$-sized tensor of relevance values. To create a heatmap, we sum over the first axis and normalize the values to a value range of $[-1\ 1]$ per pixel --- $R_i \leftarrow {R_i}/(\max_j |R_j|)$ --- and rescale the image via linear interpolation to the original size of 210$\times$160 pixels. To give an intuition for interpreting the heatmaps, we demonstrate how they correspond to different situations and strategies for two examples, the games \textit{Breakout} and \textit{Video Pinball}, both of which show above-human performance for the trained learning machines
\cite{mnih2015human}.
\begin{figure}[!t]
	\centering
 \includegraphics[width=0.7\textwidth]{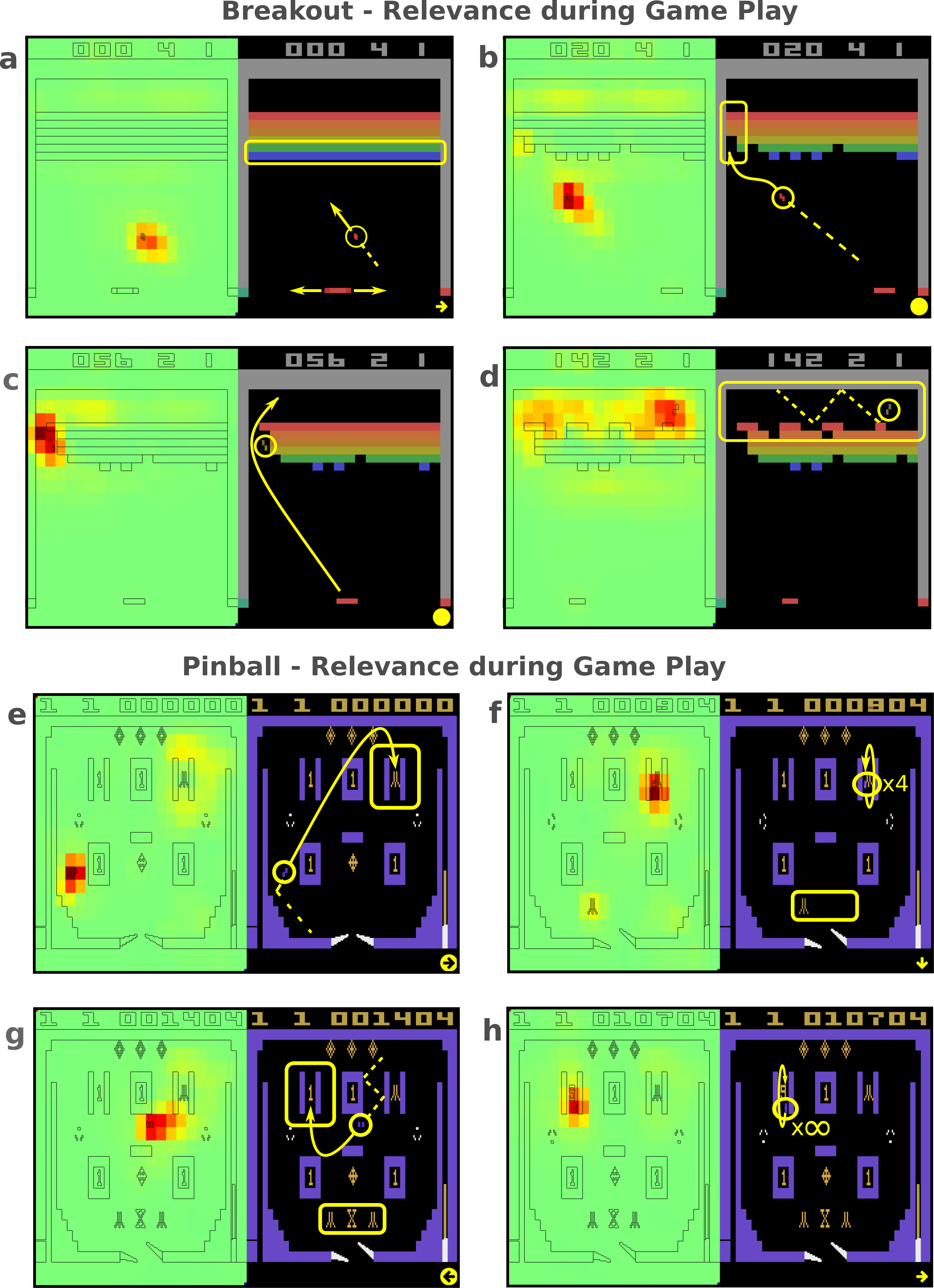}
 \caption{ \textbf{Breakout}: a) DNN tracks the ball, relevance is focused almost exclusively on the ball. b) The agent tries to build a tunnel. Relevance allocation concentrates on tunnel and ball. c) The ball moves through the tunnel. d) The agent tries to keep the ball above the brick wall. The lingering cloud of positive relevance above the bricks clearly reveals the agent's strategy. \textbf{Pinball}: e) The DNN targets the bonus life trigger, with positive relevance focusing on the ball and trigger area. f) The ball is moved through the trigger area four times, with the DNN tracking the progress in the icons at the bottom. g) The ball is maneuvered via ``nudging'' towards the scoring trigger. h) The DNN tries to move the ball through scoring trigger indefinitely.}
 \label{fig:games}
\end{figure}
Figure \ref{fig:games} displays game visuals captured at different stages of the Atari games \textit{Breakout} and \textit{Video Pinball} alongside the corresponding heatmaps produced by a trained network.

\paragraph{DNN Agent Gameplay in Breakout:}
We observe: In the early stages of the game, the neural agent concentrates on tracking the ball and follows its movements with the paddle on the bottom of the screen. The relevance is almost exclusively focused on the ball, as illustrated in Figure \ref{fig:games} a). Only when the ball is in downwards movement and close to the paddle, strong positive relevance is allocated on the paddle for a very brief moment. During this game play stage, also a weak uniform relevance is attributed to the area around the colored bricks at the top, where the model aims to shoot the ball.

Each time the ball hits a block, the block disappears, and the score increases. When the ball hits an exposed block from the third row of blocks, the ball speeds up to twice its initial movement speed. The neural network agent recognizes the increase in ball velocity and transitions into the second phase of its learned game play strategy. It focuses on targeting the ball towards the leftmost column of bricks, as seen in Figure \ref{fig:games} b), where it has learned to create a vertical tunnel\footnote{This behavior has been also observed by the authors of \cite{mnih2015human}.} through the colored wall. Relevance allocation indicates that the model still tracks the ball, but also is aware of the area where it tries to build the tunnel.

After finishing the tunnel, the model attempts to position the ball above the brick wall (Figure \ref{fig:games} c) and d)) and keep it in this area, where a quick accumulation of score rewards is possible. A permanent cloud of strong positive relevance perseveres above the brick wall, as seen in Figure \ref{fig:games} d), including those times the ball is below the brick wall, clearly reveals the agent's strategy. All of the above clearly embodies an appropriate ``understanding'' of the game and its strategic targets. 

\begin{figure}[t]
\centering\includegraphics[width=0.8\textwidth]{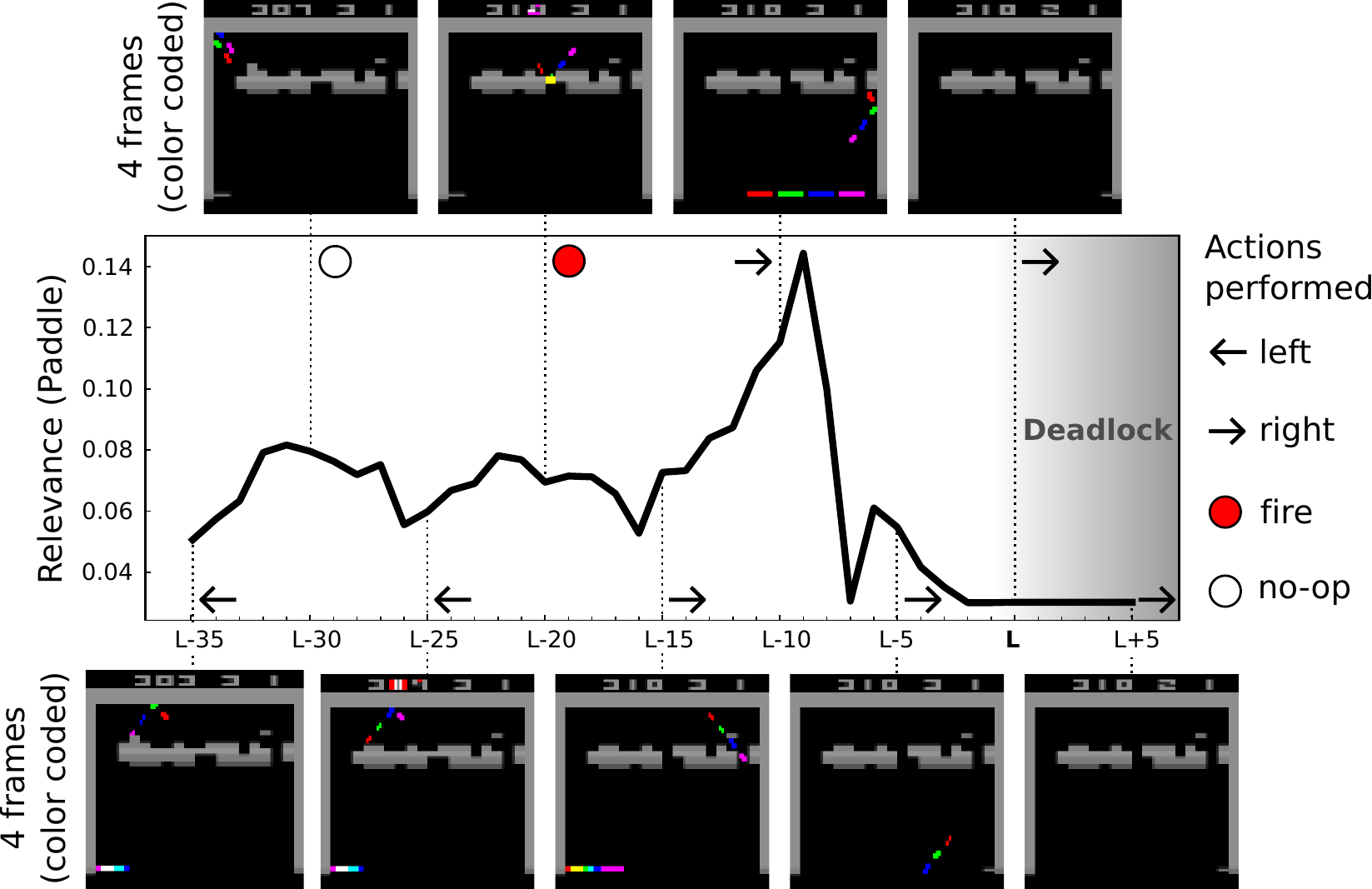}
\caption{Interactions between the DNN model and the Atari Breakout in a series of time steps around an occurring deadlock situation. The model has been trained using the configuration from \cite{mnih2015human}. For this game play sequence no randomly chosen actions have been used as input for the game interface. The deadlock situation firstly occurs at time step \textbf{L}=764.
The line plot in the middle shows the total amount of relevance allocated to the paddle controlled by the DNN agent.
The images above and below the line plot visualize network inputs (i.e., each image represents four consecutive frames) at the time of the deadlock \textbf{L}, as well as before and after in intervals of 5 time steps alongside the actions chosen by the model for those inputs. Each color used in the shown frames corresponds to one layer of the DNN input tensor, with magenta colored pixels showing the most recent game state (i.e., most recent frame) provided to the network and red pixels show the oldest game state (i.e., oldest frame) in the input tensor. Pixels which do not change at all are shown as brightness values from black to white. The relevance allocated to the paddle peaks around time step L-10, where the agent moves the paddle to the right in an attempt to intercept the ball in its downward movement. The second, smaller peak around L-4 corresponds to the ball missing the paddle and exiting the screen. From here on the model loses track of the paddle --- which can be measured by the now minimal amount of relevance in the area of interest --- and keeps on predicting the command to move further to the right. Since the paddle will not fully reappear from its halfway hidden state by moving to the right --- on the contrary it will not move at all --- no change in input visuals will be provided to the model, resulting in a deadlock.}
\label{fig:breakoutFailsToRestart}
\end{figure}

As a side remark, we would like to add a caveat: Should the current ball be lost at this stage, then the agent will stall and not enter a new ball (should there still have been any in reserve) voluntarily if the paddle is not approximately centered. This behavior is somewhat surprising and asserts that the system still lacks some deep understanding of the game and the system's possibilities to interact with it. A restart (i.e., the agent receives a new ball) may occur, however, as a randomly chosen command. If actions are chosen solely based on the Q-function (without any randomness), then the agent will not restart and continue playing the game. Figure \ref{fig:breakoutFailsToRestart} visualizes a sequence of interactions between the trained model and the Atari interface, until such a deadlock situation occurs and the game stalls. We conclude that this ``don't known what to do'' situation occurs, because the model was not exposed to sufficient amounts of training data covering a game environment almost cleared of bricks. Humans are much better in coping with such new infrequent situations, because they understand the rules of the game and have the ability to plan. Although the AI agent outperforms humans in terms of measured performance and reflexes, it fails to cope with such simple game situations, in fact the system has to rely on random action selection in this situation as it does not come at any additional costs.

\paragraph{DNN Agent Gameplay in Video Pinball:}
Also in the game {\em Video Pinball} we observe a behavior of the neural agent which is surprising and which sheds an interesting new light on the overall capabilities of the system. As a result of the learning process, the agent plays solely by ``nudging'' the table and thus completely ignores the paddles designed to prevent the ball to fall out at the bottom. Interestingly, the agent has learned to balance the use of the ``nudging'' input well enough to effectively steer the ball without ever being penalized as ``tilt'' (which happens after ``nudging'' too frequently or tilting the pinball table to steeply\footnote{See game manual at \mbox{\texttt{https://atariage.com/manual\_html\_page.php?SoftwareLabelID=588}}}) It has identified this peculiar control approach as the optimal strategy to play the game. Let us analyze this unexpected result closer with LRP. Initially, the ``pipe'' structure at the top right of the pinball table holds a high concentration of relevance. The agent tracks the location of the ball and pipe and tries to move the ball towards the pipe element, as shown in Figure \ref{fig:games} e). Once the ball arrives at its destination, the agent uses ``nudging'' and collisions with other table elements to let the ball pass through the pipe exactly four times. For each time passing the pipe, a score increase is rewarded to the player and after the fourth passing, an extra ball is added to the player's reserve. Afterwards, no more extra balls are awarded until a ball has been lost. The agent is apparently tracking its progress by focusing on the items appearing at the bottom of the screen each time the ball moves through the pipe (Figure \ref{fig:games} f)).

After the fourth change in icon appearance due to passing the top right pipe (after one Atari icon appearing each time, the middle one changes to an X), the screen flashes white once and LRP shows that all relevance focus is removed from the bottom icons and accumulated on the top left pipe element instead. The agent uses the ``nudging'' strategy again to move the ball towards the left, as can be seen in Figure \ref{fig:games} g).
The agent now bounces the ball through the pipe and off the block below/the wall indefinitely, while using the ``nudging'' strategy to control the movement. Figure \ref{fig:games} h) visualizes this behavior through the LRP heatmap. For each time passing the left pipe, the agent is awarded with a score increase --- identical to the reward earned by passing the right pipe. Passing the left pipe, however also increases a counter, which further increases a reward added to the current score, as soon as the ball is lost.

Notably, the overall strategy followed by the agent --- first ensuring to secure an extra life per round to prolong the game, and then switching sides to gain an additional score increase in case the current ball is lost --- shows highly strategic behavior, targeted at maximizing long term reward. The agent has learned to stay within the rule boundaries of pinball tricking/cheating only to an extent that no tilt ends the game. Thus, the machine has found an optimal strategy to continuously increase the score at only minimal risk of hazardous ball movement, perhaps not the one targeted by the original game designers.

\subsection{Is the AI Agent Cheating ?}
\label{sec:cheating}
In both examples the neural agent learns a strategy that leads to a maximized long-term reward within the limited scope of possible game states.
Thus, it is doing what it is supposed to do according to Equation \ref{eq:update}. Still the behavior is to a large extend unexpected and may be regarded as {\it cheating}, because it does not aligned with our idea of how to play these games. The agent is playing according to the rules of the game, but it exploits some weak points of the system. Such behavior may be not acceptable for humans in real life (although in the case of Atari games it is certainly fine).
Human players often have an implicit agreement what is regarded as fair play, e.g., in sports \cite{simon2018fair}. Not following these unwritten rules may have negative consequences up to social isolation. Learning machines are trained to solve a given task and do not have to fear these negative consequences, unless explicitly included into the game design or the agent's objective function. Thus, from the agent's point of view the ``tunnel'' and ``nudging'' strategies are not cheating, but a clever way to maximize long-term reward. 

Since unintended and unexpected behavior of the AI system may pose severe risks in real applications (e.g., in medical domain), it is important to detect it. Although it may be possible to draw conclusions about the agent's intentions and strategy by just observing the behavior of the agent, we argue that it is much easier when visualizing the focus of the networks' decision making with LRP. With the heatmap it is clearly evident from just one game that, e.g., the agent is tracking the ball, purposely wants to build a tunnel or completely ignores the paddles in pinball. Drawing such conclusions from sole observations of the agent is much more difficult (e.g., it is not obvious that the agent ignores the paddles) and certainly requires watching a lot more games. But there is also a limit to what can be practically deduced solely from the observations. Section \ref{sec:analysis} presents a technique for automated detection erratic behavior of the model, which only works with heatmaps as they are much more informative than raw observations.
Also the fact that the artifacts in the Pascal VOC dataset have not been detected for almost a decade and probably still would not have been if methods such as LRP were not available, is clear evidence that model analysis from observations does not work.

In the next section, we take the LRP analysis even one step further and quantify how the neural agent's strategy is being refined during training by monitoring the relevance attributed to certain game regions; this will permit automated quantitative analysis without the need to manually watch long sequences of game play.

\subsection{Quantifying Strategy Shifts}
\label{sec:breakout}
During training of neural agents, different stages of behavior can be observed. This is mirrored by our analysis of the networks focusing on different game aspects. Using LRP, we can measure the relevance of input regions and changes in relevance allocation with ongoing training, allowing us to quantify changes in the neural network agent's focus of attention. As a case study, we will now analyze the complex tunnel-building strategy that the agent follows in a game of \textit{Atari Breakout}. Here the network training progresses from initially just being able to recognize the ball and later the paddle with which the network learns to control the game, until finally acquiring long-term strategies as subsequent behavioral stages. Specifically, we are able to observe that the network recognizes certain game objects reliably before it is even able to play the game rudimentarily well. We take that as an indicator that early in training the convolution filters have adapted to respond to the moving game elements such as the ball and paddle well, while the top layers of the network are not yet able to use this information to form an appropriate strategic response for the current state of the game, other than correlating ball movements to changes in the score. That is, the model has learned that the ball is relevant for the game. We note that analysis like this can help to investigate shortcomings of a network-based predictor, e.g.\ to distinguish between problems within the lower level architecture, responsible for tracking, from shortcomings in the top layers, responsible for devising a decision.

\subsubsection{Quantifying the Focus of the Deep Networks over Training}
\label{sec:attn_over_training}
To consolidate our observations from watching the network play at different levels of training progression, we analyze how the focus of the neural agent changes during training using the various heatmaps computed with LRP. For our analysis, a neural network agent has been trained for 200 epochs to play \textit{Atari Breakout}, with each epoch spanning 100,000 parameter update steps. At the end of each training epoch, a snapshot of the current model state is saved, resulting in 200 models of different levels of expertise at playing the game.

To get a sample set of game states to be analyzed, we let the fully trained network play a game of \textit{Atari Breakout} for a series of 2000 frames, where every input state, Q-value and action passed to the game interface are recorded. From this set of 2000 frames, 500 frames from the early game phase and another 500 from the late game phase are taken and used as inputs for each of the 200 network training stages obtained earlier.
To reach comparability across networks, the same inputs are used for all the networks, instead of letting each network generate its own input by playing. Clearly, for a model reflecting an earlier state of training it is quite unlikely that complex strategies such as tunnel building could be observed, thus leaving the focus on the tunnel for this model undefined. With our protocol, we intend to see whether networks in earlier learning stages are already able to recognize the usefulness of a tunnel or not, although this knowledge would only be exploited in the strategy at a later stage.

\begin{figure}[ht]
	\centering
 \includegraphics[width=0.8\textwidth]{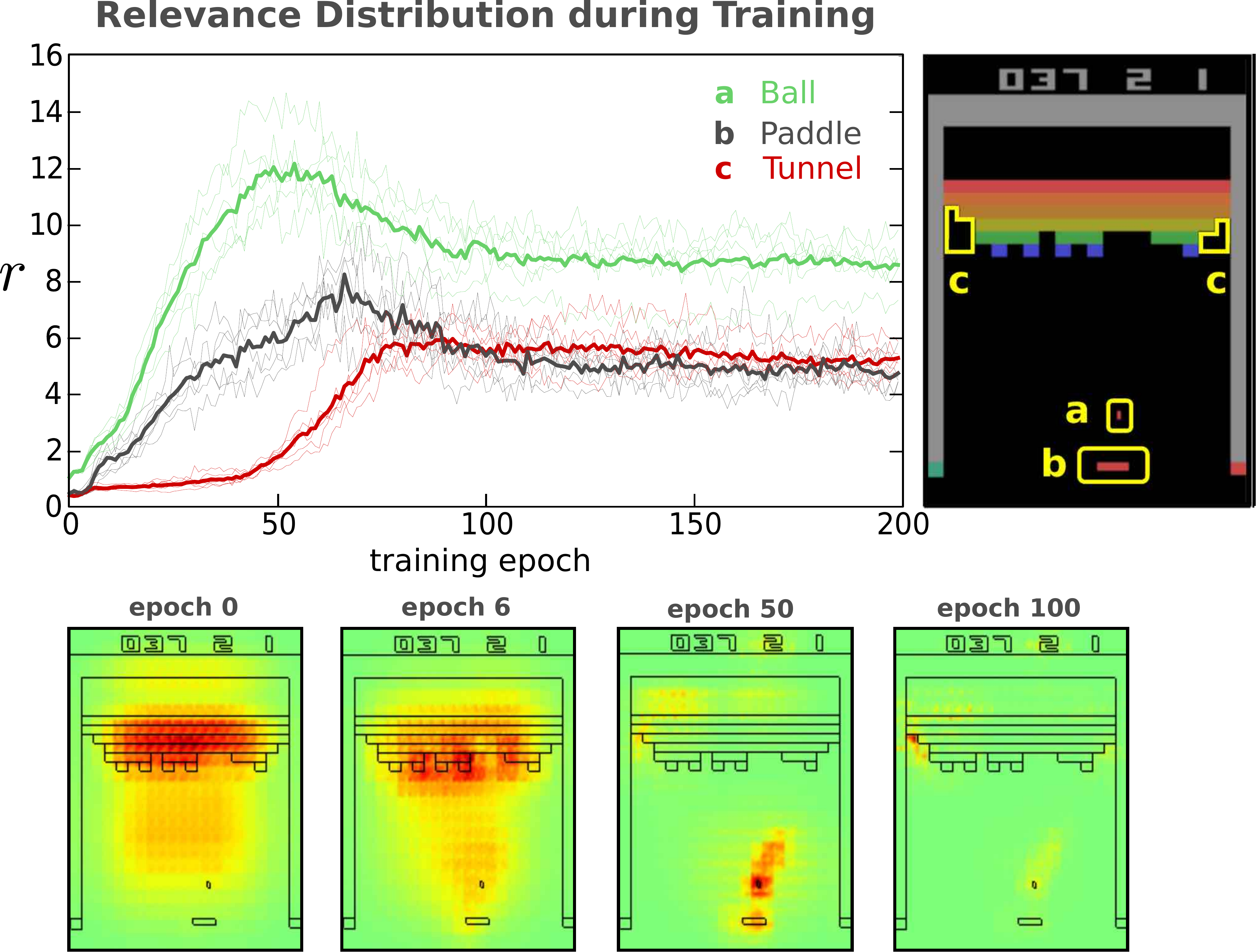}
	\caption{A neural network was trained on the game Breakout. We measured the relative relevance on different objects during the training epochs, evaluated on a game sequence of 500 states. Thin lines: 6 individual runs of training. Thick lines mean over the 6 individual runs. The attention on the ball grows first, followed by the paddle. After 50 training epochs the tunnel regions come into focus, which indicates a shift to a tunnel building strategy.}
	\label{fig:hmEvolution}
\end{figure}

For every model and input sample, relevance maps are computed with respect to the model decision. For a set of important game objects we measure the total amount of relevance allocated to their respective pixel regions. To minimize the influence of randomization effects from the training process, the experiment is repeated over six different training runs (i.e.\ we have trained 6 networks in total, with 200 network snapshots created in regular intervals for each network), with the individual network responses and the mean thereof being reported in Figure \ref{fig:hmEvolution}. The tracked game objects are:

\vspace{0.1cm}
{\bf The ball:} A rectangular region enclosing the ball with two pixels of padding in each direction. The relevance response is only measured if the ball is below the brick area, in order to distinguish the amount of relevance attributed to the ball from the amount of relevance score covering the bricks.
  
{\bf The paddle:} A rectangular region around the paddle with two pixels of padding in each direction. The relevance response for the paddle is only measured when the measurement regions of ball and paddle do not overlap, to avoid confusion in the relevance allocation measurements taken.
  
{\bf The tunnel:} A region defined by all the pixels in the leftmost and rightmost columns of the brick wall, which have been cleared.
\vspace{0.1cm}

For every frame, the relative amount of relevance attributed to an object is calculated by summing over the pixel region defined for each object as described above, divided by the number of pixels per object, to compute a measurement of relevance per pixel score for each object. Additionally, this value is then divided again by the average pixel relevance score taken from the full frame to mitigate the effect of better trained networks generally predicting higher Q-values (and thus attributing higher relevance scores). The resulting relative relevance $r$ per game object is then
\[
 r = \frac{\text{relevance on object}}{\text{total relevance in frame}}\cdot\frac{\text{area of frame}}{\text{area of object}}.
\]

The object to be recognized earliest during training is the ball, since its position and movement direction are a direct indicator of a change in score or an expected penalty due to losing the ball. This is a prediction task that can be solved without knowledge about future frames. In fact, the networks are able to recognize the ball before grasping the controls well enough to play the game. The importance of the paddle manifests itself only a few training epochs later. The paddle indirectly influences the rewarded score by preventing the ball from leaving the screen, allowing the game to accumulate additional reward in future frames. Connecting an action to a delayed reward or penalty is a considerably more complex strategic task than just recognizing an immediate score increase which is connected to an event.

After roughly 40 epochs of training, the networks attempt to build tunnels at the sides of the brick wall, quickly breaking through the bricks. After the breakout, the ball ricochets between the borders limiting the game area and the highest valued blocks at the top for longer series of frames, limiting the need for interaction while maximizing the received reward. The relevance plot in Figure \ref{fig:hmEvolution} mirrors the discovery of this strategy of the learning machine. It shows increasing amounts of relevance attributed to the tunnel region over time until about the 80th training epoch, from which on the attributed proportion of relevance remains almost constant.

On the one hand, building a tunnel can be interpreted as an advanced strategic task, since the AI has to plan ahead many steps in advance to create a favorable game state that will ultimately result in high rewards. On the other hand, one could argue that the only \textit{intelligence} to be found is in the \textit{design} of the training process, since the use of reinforcement learning encourages delayed gratification to increase the overall reward. The latter view is backed up by the observation that network agents having seen 200 epochs of training curiously fail to restart a new ball (see Figure \ref{fig:breakoutFailsToRestart}) due to the apparent lack of transfer knowledge, a \textit{key element} of true intelligence. In this case the agent fails to transfer knowledge to later stages of the game, namely that the ``fire'' button needs to be pressed if the ball is lost. This was learned in beginning stages of the game, but is not transferred since the game situation is only observed seldomly in the advanced game phases.

\subsubsection{Varying Depth of Architecture}
In this section we compare the relevance responses of neural network agents with different model architectures, namely the architecture from \cite{mnih2015human} (denoted as Nature architecture), the architecture of its predecessor in \cite{Mnih2013NIPS}, (denoted as NIPS architecture) and a third network with an architecture similar to \cite{mnih2015human} (Small architecture). The NIPS architecture has one convolutional layer less than its successor, whereas the Small architecture has the same number of convolutional layers as the Nature network, but is limited in its decision making capacity by only one fully connected layer instead of two. An overview in the different network architectures is given in Table \ref{tab:archCompTable}.

\begin{table}[h]
\centering
 \caption{A comparison of the investigated network architectures. To describe a layer, we use the notation $(I){\rightarrow}(O), [S]$ where $I$ is the shape of the (convolutional) weights,
$O$ are the output responses and $S$ is the stride for a convolutional layer. For the Nature network, C1 $(4{\times}8{\times}8){\rightarrow}(32), [4{\times}4]$ describes the first convolutional layer with a learned filter bank counting 32 convolutional 8$\times$8 filter tensors of 4 input channels each, applied at a 4$\times$4 stride across the layer input. Similarly, F2 $(512){\rightarrow}(4)$ describes the second and in this case last fully connected layer with 512 input nodes and 4 outputs.}
\small
\begin{tabular}{ l|l|l}
 NIPS architecture & Nature architecture & Small architecture \\
 \hline
 C1 $(4{\times}8{\times}8){\rightarrow}(16), [4{\times}4]$ & C1 $(4{\times}8{\times}8){\rightarrow}(32), [4{\times}4]$ & C1 $(4{\times}8{\times}8){\rightarrow}(32), [4{\times}4]$\\
 C2 $(16{\times}4{\times}4){\rightarrow}(32), [2{\times}2]$ & C2 $(32{\times}4{\times}4){\rightarrow}(64), [2{\times}2]$ & C2 $(32{\times}4{\times}4){\rightarrow}(64), [2{\times}2]$\\
  & C3 $(64{\times}3{\times}3){\rightarrow}(64), [1{\times}1]$ & C3 $(64{\times}3{\times}3){\rightarrow}(64), [1{\times}1]$\\
 F1 $(2592){\rightarrow}(256)$			    &F1 $(3136){\rightarrow}(512)$ & F1 $(3136){\rightarrow}(4)$\\
 F2 $(256){\rightarrow}(4)$			    & F2 $(512){\rightarrow}(4)$ 	&
 \end{tabular}
 \label{tab:archCompTable}
\end{table}

\begin{figure}[ht]
	\centering
 \includegraphics[width=0.65\textwidth]{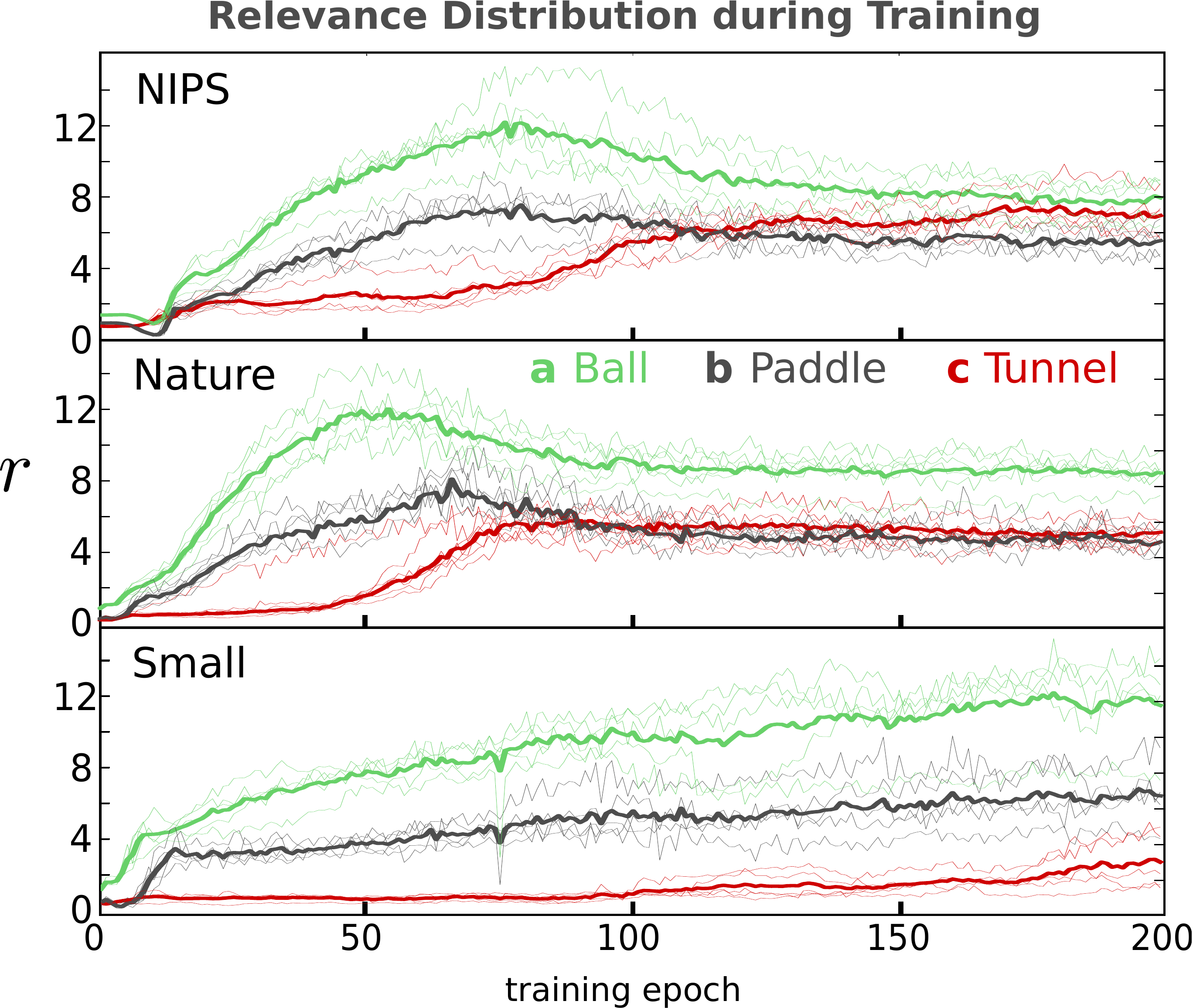}
	\caption{Comparison of the network attention in terms of relevance responses of three different model architectures to important game elements (ball, paddle and tunnel) as a function of training time. An overview of the compared model architectures is given in Table \ref{tab:archCompTable}. All three architectures show a shift in strategy towards tunnel building. For the NIPS architecture the shift appears in two stages resulting in a larger shift in total, whereas for the Small architecture the tunnel does not become a relevant objective during the training period.}
	\label{fig:architecture}
\end{figure} 

For the NIPS architecture, the focus on ball and paddle is delayed, emerging approximately 5 training epochs later when compared to the Nature network (see Figure \ref{fig:architecture}). The relevance on the tunnel region shifts in two stages and finally rises to an amount higher than that recorded for the Nature architecture. A possible interpretation is that the shallower convolutional architecture slows down the recognition of the important game elements, but once recognized the development of higher level strategies is possible.

For the Small architecture, the focus switches very early to ball and paddle. The three convolutional layers allow for a quick recognition of the important game objects. However, the game play performance remains at a sub-human level and more advanced strategies like tunnel building remain out of focus. The agent learns to follow the ball with the paddle, but is only able to successfully reflect the ball in three out of four interactions.

\subsubsection{Varying Size of Replay Memory}
We investigate the influence of replay memory size on the development of a neural network agent. The replay memory is a dataset of previous game situations described as the tuple \texttt{(state, action, reward, next state)}. After each interaction with the Atari emulator one new game situation is added to the memory. Once the memory capacity is reached, the earliest entries are replaced with newer ones. For training, a random batch is drawn from the memory.
\begin{figure}[!h]
	\centering
 \includegraphics[width=0.67\textwidth]{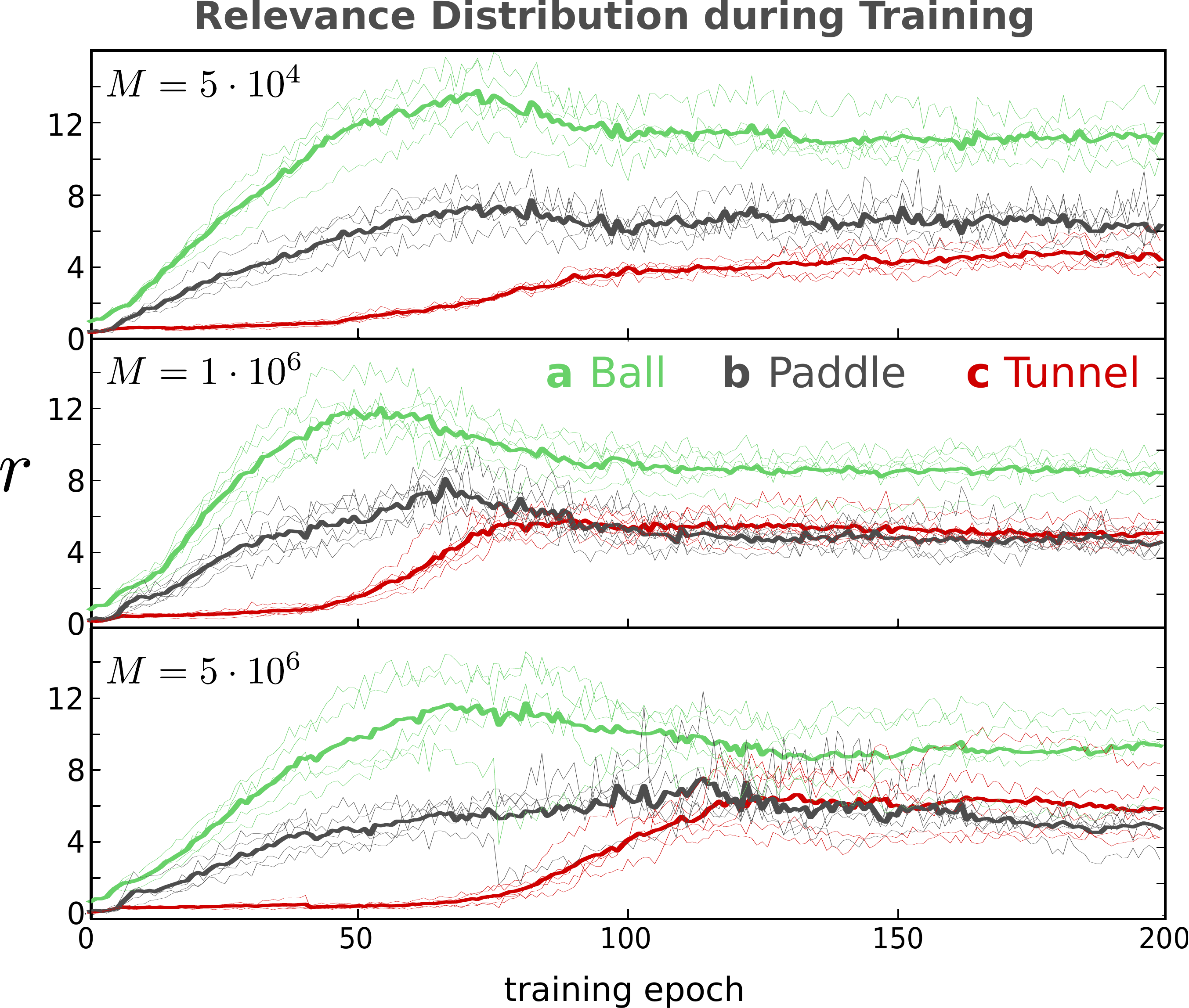}
	\caption{Evolution of network attention on the game elements ball, paddle and tunnel for varying replay memory sizes $M$, with $M=5\cdot10^4$, $M=10^6$ and $M=5\cdot10^6$. All three memory settings allow for a shift of strategy towards tunnel building.}
	\label{fig:memory}
\end{figure}

In Figure \ref{fig:memory} we present the development of the relevance of the ball, paddle and tunnel to neural network agents, for memories of capacities $5\cdot10^4$, $10^6$ and $5\cdot10^6 $. For all memory size settings, the corresponding models shift their attention to the tunnel area with ongoing training. For the smallest memory size of $5\cdot10^4$ game states we note a more gradual shift in strategy that starts earlier in training and leads to less concentrated model attention registered on the tunnel area. The largest memory of $5\cdot10^6$ game states shifts to the advanced strategy at a later time, but more abruptly so. Relevance scores concentrate more on the tunnel area when compared to the other models. A later onset of strategically recognizing the tunnel during training as an effect of replay memory size can be attributed to the vast amount of early game states still populating the memory. Many more training epochs are needed to fully eliminate the effect of those game states that have resulted from many short games with uncontrolled game play in early training epochs. A smaller replay memory might replace those game state recordings faster, but does not allow for enough memorized observations of later phases of the game (or full games played more expertly) to learn the tunnel building strategy successfully.

\subsubsection{Changes of Model Attention during Game Play}
\label{sec:model_attention_during_game_play}
\begin{figure}[!tb]
	\centering
 \includegraphics[width=0.8\textwidth]{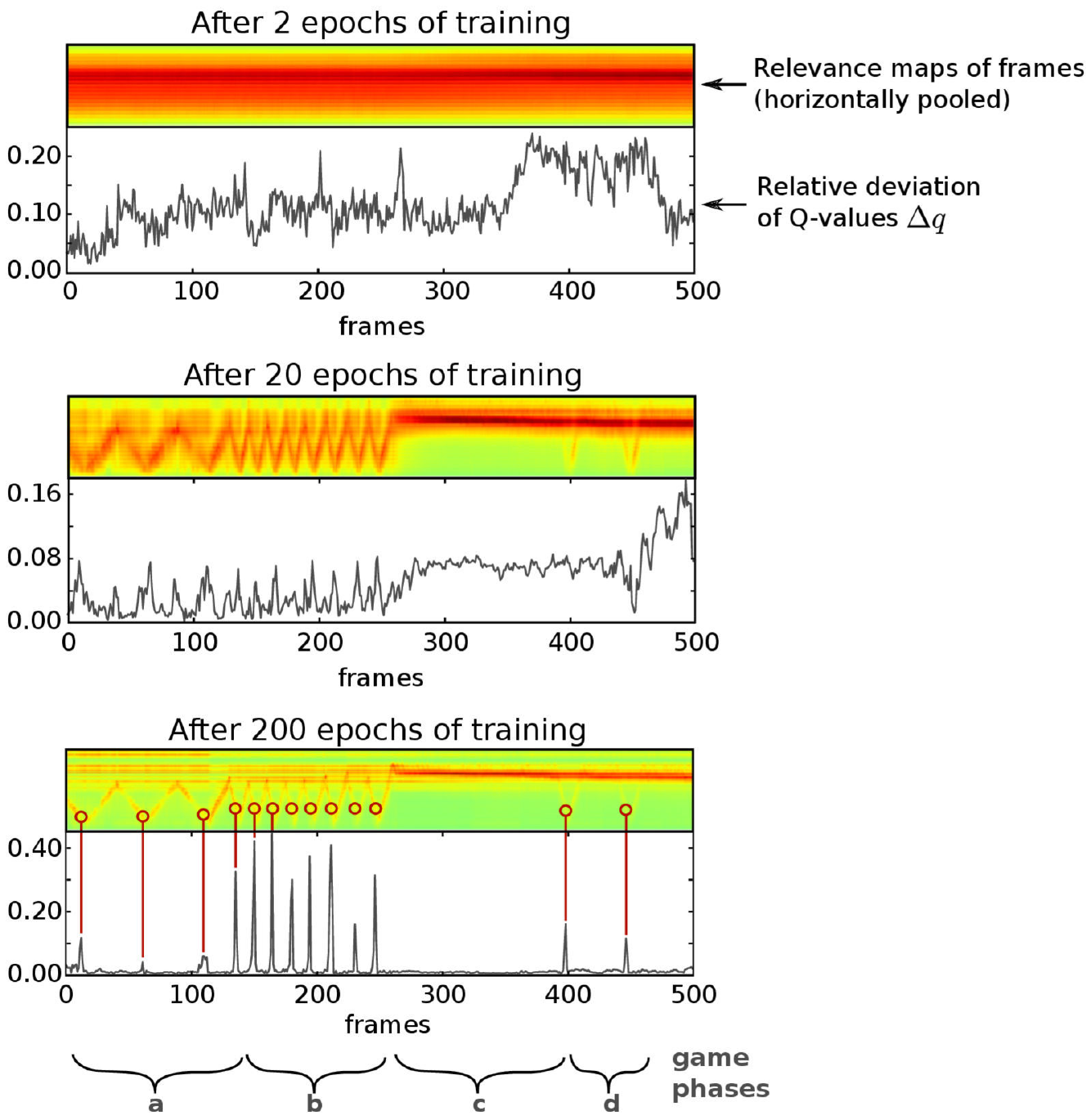}
	\caption{Divergence of $Q$-values over game time, contrasted with horizontally pooled heatmaps. a) The ball moves slowly in this early phase of the game, only small spikes in $\Delta q$ are recorded. b) The ball increases in speed after hitting a brick in a layer beyond the lowest two. Spikes in $\Delta q$ increase as recovery time after a mispredicted action is decreased. c) The ball breaks through and stays above the wall. Temporarily no interaction from the agent is necessary. d) The ball moves back down twice and is reflected by the paddle. To increase visual readability of the network attention, gamma correction with $\gamma = 0.5$ has been applied after normalizing the heatmaps, increasing the contrast of the color coded heatmaps responses.\
	We compare the $\Delta q$ and the horizontally pooled heatmaps between three selected training epochs, i.e.\ after 2, 20 and 200 epochs of training. At certain frames in the game, $\Delta q$ becomes much larger than zero, signifying the expected reward of one action dominating the others. When it makes no difference which action is taken, since the ball is far from the paddle, the correct long-term reward of all actions will be close and $\Delta q$ small. Late-stage networks appear to incorporate this fact -- they become more focused on certain frames. After two epochs of training, $\Delta q$ fluctuates strongly with no apparent focus. After 20 epochs, peaks of diverging q-values emerge. After 200 epoch, $\Delta q$ is generally close to zero except for frames in which a falling ball has reached a height were action is necessary. Complementing the temporal focusing is a spatial focusing of the heatmaps on important game objects, see also Figure \ref{fig:hmEvolution}. We illustrate this showing the horizontally pooled heatmaps over the game time. Over training, a narrower portion of the game frame gets an increasing portion of the relevance, making the focus on the ball apparent.}
	\label{pulseAndHeatmaps}
\end{figure}

Over the course of a running game of \textit{Atari Breakout}, we investigate how and where relevance allocation changes over time. To visualize heatmaps over time, the relevance map computed for each frame is summed over the horizontal axis, resulting in a relevance vector with an entry for each vertical pixel location. Then, the compacted frames from all time steps are concatenated column-wise with the time now being the new horizontal axis. Relevance values attributed to the ball, the paddle and the tunnel can still be distinguished since they have different vertical positions for most of the time. We complement this analysis with information about at which time steps or events, certain actions are predicted significantly stronger than the other options. For a trained network, the Q-values of all possible actions generally only differ by relatively small quantities. If the ball is far from the paddle, for example, the next action does not influence the predicted score and game play much. At points in time when single actions do have a large impact, the predicted Q-values diverge. We measure this relative deviation of Q-values as
\[
 \Delta q = \frac{q_{max} - q_{min}}{q_{max}}
\]
where $q_{max}$ and $q_{min}$ are the maximum and minimum predicted Q-value for a given input, respectively. Over the course of training, the networks become more decisive. Figure \ref{pulseAndHeatmaps} shows $\Delta q$ during a game sequence of 500 frames for three different stages in training: After two, twenty and two hundred epochs of training, with 100.000 training steps per epoch. We observe that the fully trained network's prediction output focuses on single output neurons at times when it has to interact with the ball, while earlier versions of the same network are not yet fully able to decide with a comparable amount of certainty.

The apparent \textit{zig-zag} structure in the heatmap visualizations in Figure \ref{pulseAndHeatmaps} is the up-and-down movement of the ball, as tracked by the neural network agent over time. Applying LRP allows us to track the ball without having to design a specific tracker. Combining this with the temporal focus of the network we can discern at which height of the ball the decision for an action is made. Figure \ref{pulseAndHeatmaps} illustrates these more important action predictions for a fully trained network as red circles and visualize the change in relevance allocation over the time of a game. Note that the network itself becomes more focused over training time (cf.~the discussion in Section \ref{sec:vis_and_comparison_gameplay}).

\subsubsection{Comparing Relevance Maps to Gradient Map}
Another popular method to analyze neural network type models or reinforcement learning systems is Sensitivity Analysis (cf.~\cite{DBLP:journals/jmlr/BaehrensSHKHM10,DBLP:journals/corr/SimonyanVZ13, DBLP:conf/icml/ZahavyBM16}), a gradient-based saliency method. In a nutshell, Sensitivity Analysis computes the network gradient at a specific input point and then applies the $\ell_2$-norm over the input channels at each pixel. The magnitude of this gradient response at input level then yields information about the sensitivity of the model to changes in certain
regions of the input.

We repeat our experiments from Section \ref{sec:attn_over_training} and \ref{sec:model_attention_during_game_play} with the gradient maps computed via Sensitivity Analysis to compare both methods. As we will observe, LRP provides clearer results.
\begin{figure}[t!]
	\centering
 \includegraphics[width=0.8\textwidth]{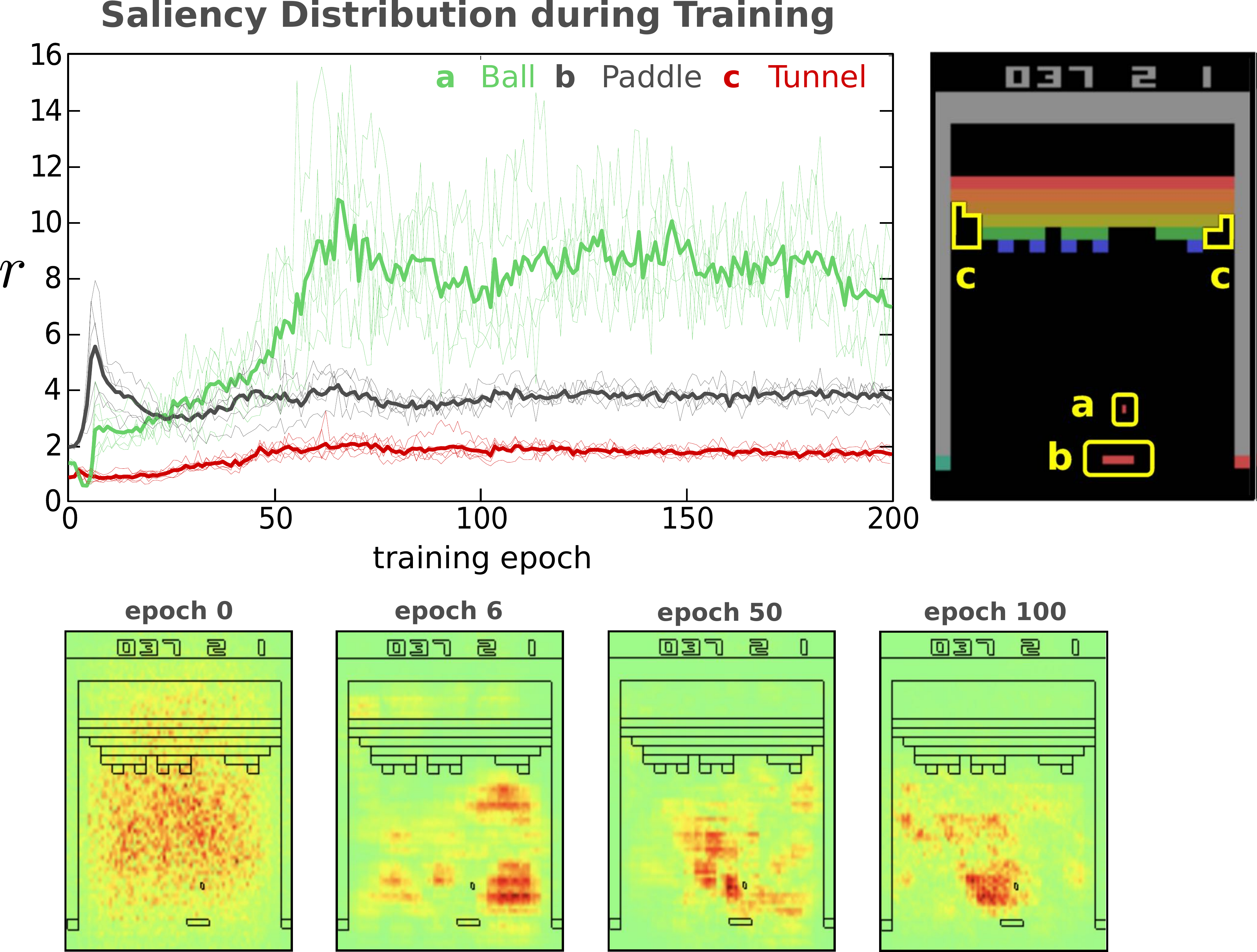}
	\caption{With the same neural network classifier trained to play \textit{Atari Breakout} as used in Section \ref{sec:attn_over_training}, gradient maps are computed using Sensitivity Analysis. We measure the relative saliency on different objects with ongoing model training, evaluated on a game sequence of 500 states. The plot shows the gradient magnitude allocation on the ball, paddle and tunnel elements for all training epochs for all six models. Bold lines show the average result over the six models.
Gradient maps are less local and sparse compared to Relevance maps computed by LRP. Relevance maps do also align better with the game elements in pixel space. More saliency is attributed to the region around the ball than the ball itself. The gradient map does not respond to the tunnel area. }
	\label{fig:gradEvol}
\end{figure}
Figure \ref{fig:gradEvol} presents the results obtained when using Sensitivity Analysis instead using LRP for the setup of Figure \ref{fig:hmEvolution}. Observing the gradient maps during game play reveals that the saliency follows the general area where the ball is located, but is {\em not} focussed on the ball itself. The response obtained with Sensitivity Analysis, when given a model and an input data point, measures how much a change in the input would affect the model prediction. Here, peaking gradient values in close proximity to the ball essentially indicate where the ball should be to cause the steepest change in predicted Q-values, compared to where it actually is. Furthermore, for each of the neural network agents trained, gradient maps did not attribute weights on the pixels covering the tunnel element. The changes in game play strategy are not observable through Sensitivity Analysis.

\begin{figure}[ht]
	\centering
 \includegraphics[width=0.8\textwidth]{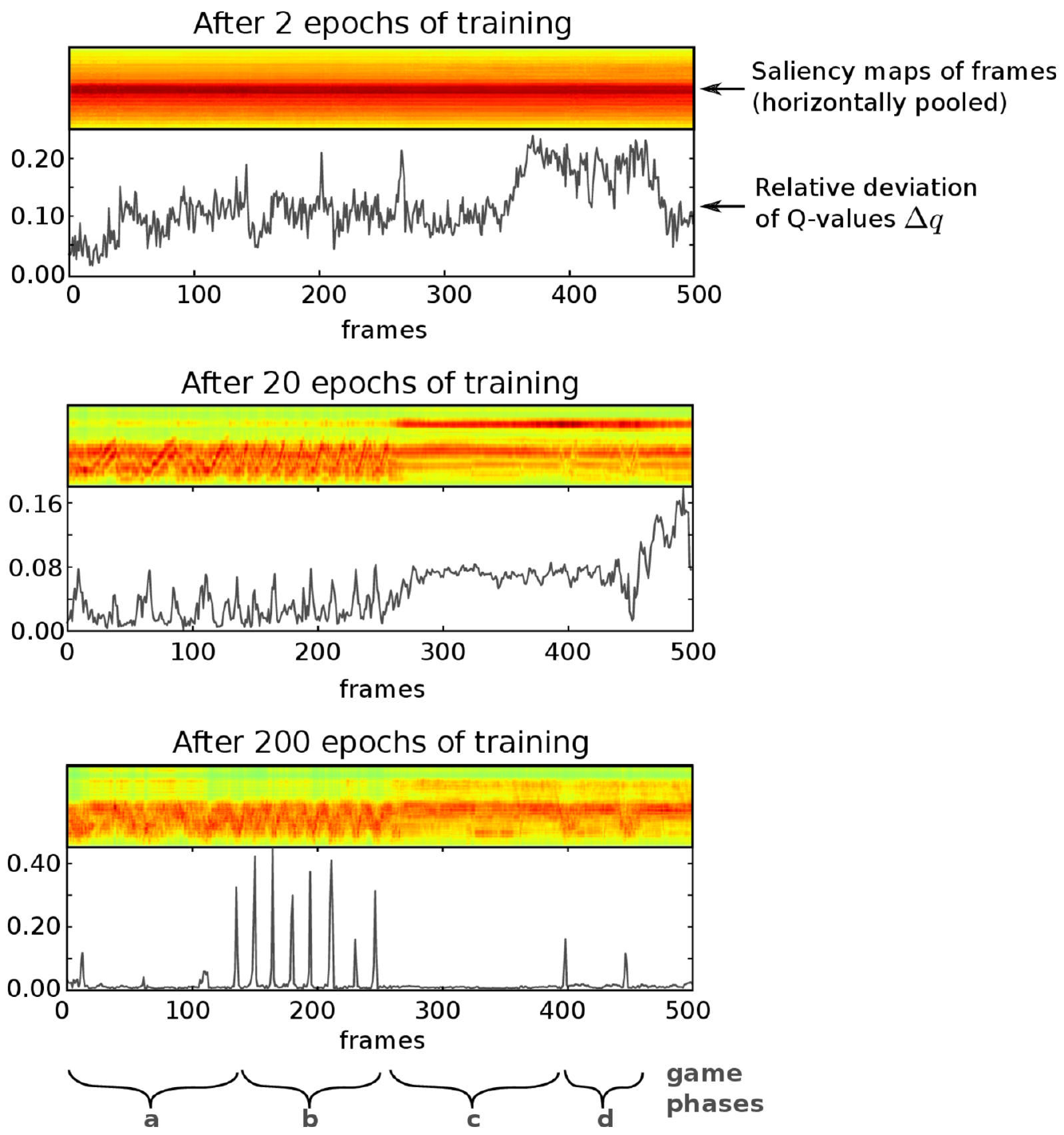}
	\caption{a)--d): Compare to Figure \ref{pulseAndHeatmaps}. The
     gradient map obtained by Sensitivity Analysis gives a less
     clear picture than LRP.}
	\label{fig:gradPulse}
\end{figure}

In Figure \ref{fig:gradPulse}, we repeat the analysis from Figure \ref{pulseAndHeatmaps} with gradient maps instead of LRP maps. Similarly, gradient maps are summed up and then concatenated along the horizontal axis, providing a representation of gradient responses over time and vertical game coordinate. The same $\Delta q$ plot line is drawn below the heatmap time line to contrast importance prediction with pixel-wise responses. While the vertical movement of the ball is still distinguishable in the gradient map, the response is by far more blurred and less local compared to the heatmaps computed using LRP. This can be explained with the visualizations described in Figure \ref{fig:gradEvol}, demonstrating strong gradient responses in areas around the ball --- where the ball has to be in order to maximally change the predictor output. The gradient map, however, is unable to indicate situations or game elements that are per se of importance to the predictor. It lacks, next to interpretability, in a clear relationship between game elements and gradient responses. We refer the reader to \cite{montavon2017methods} for a more detailed discussion on the advantages of LRP over gradient-based explanation methods.

\section{Task II: Image Categorization}
\label{sec:pascal}

As emphasized in \cite{lapuschkinCVPR16}, generalization error rates and accuracy ratings are often not sufficient for assessing the reliability of a trained classifier and the appropriateness of its problem solution strategy. This aspect is especially important for application scenarios where the usage of the machine learning model in question is for e.g.~safety critical data or in medical analysis. Explanation algorithms such as LRP can help in various ways, i.e.\ for identifying flaws in predictors and also for detecting weaknesses and artifacts in the used datasets rsp.\ benchmarks.

Even though huge data collections with millions of samples such as ImageNet \cite{russakovsky2015imagenet} exist, their size is finite and models trained based on said datasets will in most cases only ever see a small subset of all possible configurations an object might be represented in. In some image dataset for example, a ship might regularly be depicted floating on a body of water but seldomly under maintenance on a dry dock. In the latter case a human observer would still recognize a ship, but the unusual background it has been captured in might cause a classifier trained to recognize aquatic vehicles to incorrectly reject the sample. This would indicate that the concept of ``boat'' or ``ship'' has not been understood as such but heavily depends on the co-occurrence of a suitable background environment.

Therefore, analysis should be extended beyond simply calculating test set error rates and also explain the behavior of a classifier. In this section, two
state-of-the-art model architectures in their respective predictor families --- a Fisher Vector (FV) classifier and a Deep Neural Network --- are briefly described and subsequently their prediction strategies are contrasted with the help of LRP.
 
\subsection{The Fisher Vector Classifier}
\label{sec:fv_model}

The FV model evaluated in this section follows the configuration and implementation from \cite{chatfield2011devil, perronnin2010improving} and can easily be replicated by using the Encoding Evaluation Toolkit \cite{chatfield2011devil}. Twenty one-vs-all object detection models are trained on the \texttt{train\_val} partition of the Pascal VOC 2007 dataset --- one for each object class. The FV model pipeline consists of a local feature extraction step, computing dense SIFT descriptors \cite{lowe1999object} from an input image, which are then mapped to Fisher Vector representations via a Gaussian mixture model (GMM) fit to the set of local descriptors of the whole training set. The FV representations of each image are then used for training a linear SVM \cite{cortes1995support} model. The parameters used in each step of the pipeline are as follows:

The images are normalized in size to a side length of at most $480$ pixels, while maintaining the original aspect ratio. From a gray-scale version of the image, SIFT features are computed with spatial bin sizes of $\lbrace 4,6,8,10 \rbrace$ pixels with fixed feature mask orientation at spatial strides of $\lbrace 4,6,8,10 \rbrace$ pixels respectively. Using PCA, the descriptors are then projected from the originally $128$-dimensional feature space to an $80$-dimensional subspace, onto which a GMM with $k=256$ Gaussian mixture components is fit. The trained GMM is then used as a soft visual vocabulary for mapping the projected local descriptors to a FV. The mapping process involves a spatial pyramid mapping scheme, subdividing an input image into $1 \times 1$, $3 \times 1$ and $2 \times 2$ grid areas. For each of the in total 8 image areas, a FV is computed based on the corresponding descriptor set, which are then concatenated to form one FV descriptor per image. After the application of power- and $\ell_2$-normalization steps, reducing the sparsity of the vector and benefitting the model quality \cite{chatfield2011devil} this \emph{Improved Fisher Vector}\cite{perronnin2010improving} representation is used for training and prediction together with a linear SVM \cite{cortes1995support} classifier.

Evaluating the model in the \texttt{test} partition of the Pascal VOC 2007 dataset results in a $61.69\%$ mAP score in \cite{chatfield2011devil} and $59.99\%$ mAP in \cite{lapuschkinCVPR16}.

\subsection{The Multilabel Deep Neural Network}
In order to contrast a Deep Neural Network to the previously introduced FV classifier, a pre-trained model has been fine-tuned on the Pascal VOC data and classes to achieve a comparable evaluation setting. For fine-tuning, we use the Caffe Deep Learning Framework \cite{jia2014caffe} and the pre-trained BVLC Caffe reference model available from \texttt{http://dl.caffe.berkeleyvision.org} as a starting point. The model has been adjusted to match the number of Pascal VOC classes by setting the number of output neurons of the topmost fully connected layer to 20. The softmax layer has been replaced with a standard hinge loss layer to accommodate for the appearance of multiple object classes per input sample as it is the case with the Pascal VOC data.

The training dataset consists of the \texttt{train\_val} partition of the Pascal VOC 2012 dataset, augmented by first resizing each image to 256 pixels on its longest side and padding the image by repeating the border pixels to a size of $256{\times}256$, then cropping $227{\times}227$ sized subimages at the corners and the image center. Together with a horizontally mirrored version for each of the subimages, the tenfold augmented training data then counts $115,300$ training samples. Fine-tuning is performed for $15,000$ iterations with training patches holding $256$ samples per iteration and a fixed learning rate of 0.001. The resulting model achieves an mAP score of $72.12\%$ on the Pascal VOC 2007 dataset. We have made this fine-tuned multi-label model available as part of the Caffe Model Zoo\footnote{\texttt{https://github.com/BVLC/caffe/wiki/Model-Zoo\#pascal-voc-2012-multilabel-classification-model}}.

\subsection{Comparing Fisher Vector and Deep Network Prediction Strategies} 
\label{sec:FVvsDNN}
We compare the reasoning of two different classifiers --- an Improved Fisher Vector SVM classifier from \cite{chatfield2011devil, perronnin2010improving} as a state-of-the-art Bag of Words-type model to the much deeper convolutional neural network architecture described before --- by analyzing the relevance feedback obtained via LRP on pixel level. Both models are trained / fine-tuned to predict the $20$ object classes defined in the Pascal VOC datasets. In order to compute pixel-level relevance scores for the FV model, we employ the $\epsilon$-rule to obtain a numerically stable decomposition of the FV mapping step and the ``flat'' rule to project local descriptor relevances to pixels. The relevance decompositions for the SVM predictor layer are computed using the ``simple'' ($\epsilon=0$) rule (cf.~\cite{lapuschkinCVPR16} and \cite{BacICIP16} for further technical details). Since the fine-tuned DNN model structure is composed of a repeating sequence of convolutional or fully connected layers interleaved with pooling and ReLU activation layers, we uniformly apply the $\alpha\beta$-rule with $\alpha=2,\beta=-1$ throughout the network, which complements the ReLU-activated inputs fed into hidden layers especially well for explanation. Since the output of ReLUs is strictly $\geq 0$, positive weights in the succeeding convolutional or linear layer do carry the input activation to further layers, whereas negatively weighted connections serve as inhibitors to a layer's output signals. Using the  $\alpha\beta$ decomposition rule allows for a numerically stable relevance decomposition, where the parameter $\alpha$ controls the balance of neuron exciting and neuron inhibiting patterns for the explanation.

Observing the selection of results presented in Figure \ref{fig:CVPR}, the lower granularity of the relevance maps computed for the FV classifier is apparent. This effect results from the application of the ``flat'' decomposition rule for computing pixel level relevance scores from local descriptor relevances for the FV model: Whereas for DNN type predictors pixel to neuron and neuron to neuron relationships are given via distinguished (weighted) connections, such a relationship is often obscured (or does outright not exist) in the process of local feature extraction, e.g.\ especially when features based on histograms or quantiles computed from a set of pixels are used to describe parts of an image. The ``flat'' decomposition rule bridges this gap between two layers of computation at the cost of potential heatmap resolution, by uniformly distributing the relevance value of a top layer neuron or feature onto the neuron or feature's receptive field. The receptive field of e.g.\ a standard SIFT descriptor as used for the FV model is the image area the descriptor output is computed from. Analogously, the receptive field of an output neuron in a neural network context then is the set of input neurons connected and contributing to the considered output.

\subsubsection{Detecting Boats \& Horses} 
The relevance maps computed for both predictors have been compared qualitatively as well as quantitatively in \cite{lapuschkinCVPR16}. On a qualitative level we see that the DNNs prediction is dominantly based on the objects in the images themselves, whereas the FV model often resorts to using background information or co-occurring objects and image elements for support. This observation also reflects the reported prediction performances of both models, shown in Table \ref{tab:cvprresults}.
\begin{table*}
\centering
\caption{Prediction performances per model and class on the Pascal VOC 2007 test set in percent average precision.}
\begin{tabular}{|l||c|c|c|c|c|c|c|}
\hline
&  { \bf aer} & { \bf bic} & { \bf bir} & { \bf boa} & { \bf bot} & { \bf bus} & { \bf car}\\ 
{\bf FV} & 79.08 & 66.44 & 45.90 & 70.88 & 27.64 & 69.67 & 80.96 \\ 
{\bf DNN} & 88.08 & 79.69 & 80.77 & 77.20 & 35.48 & 72.71 & 86.30 \\\hline 
&  { \bf cat} & { \bf cha} & { \bf cow} & { \bf din} & { \bf dog} & { \bf hor} & { \bf mot}\\ 
{\bf FV} & 59.92 & 51.92 & 47.60 & 58.06 & 42.28 & 80.45 & 69.34 \\ 
{\bf DNN} & 81.10 & 51.04 & 61.10 & 64.62 & 76.17 & 81.60 & 79.33 \\ \hline 
&  { \bf per} & { \bf pot} & { \bf she} & { \bf sof} & { \bf tra} & { \bf tvm} & { \bf mAP}\\ 
{\bf FV} & 85.10 & 28.62 & 49.58 & 49.31 & 82.71 & 54.33 & 59.99 \\ 
{\bf DNN} & 92.43 & 49.99 & 74.04 & 49.48 & 87.07 & 67.08 & 72.12 \\ \hline
\end{tabular}
\vspace{10pt}
\label{tab:cvprresults}
\end{table*}

\begin{figure}[t]
	\centering
 \includegraphics[width=0.9\textwidth]{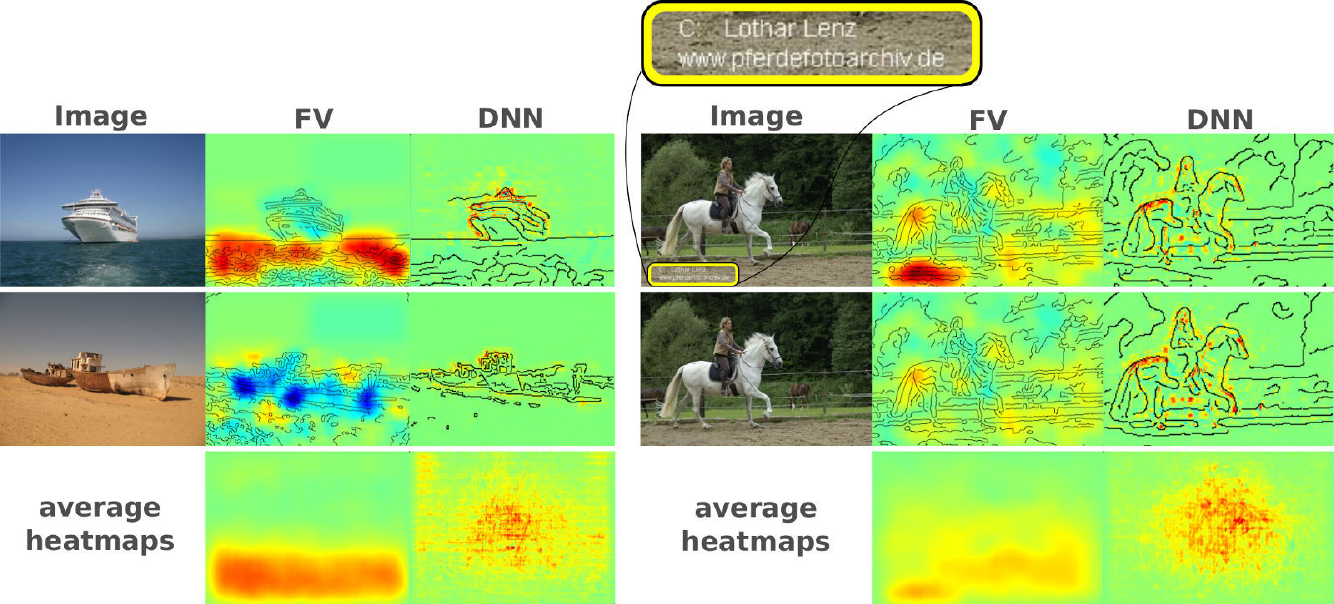}
 \caption{\emph{Left}: Example heatmaps for both models and input images showing boats. \emph{Right}: A sample belonging to class ``horse'', which is categorized correctly by the FV model only due to the presence of a copyright watermark regularly occurring in the Pascal VOC dataset. The bottom row shows the average heatmaps for each object class per model. The aspect ratios of relevance maps have been normalized before averaging.}
 \label{fig:CVPR}
 \end{figure}

\begin{figure}[t]
\centering
\includegraphics[width=0.9\textwidth]{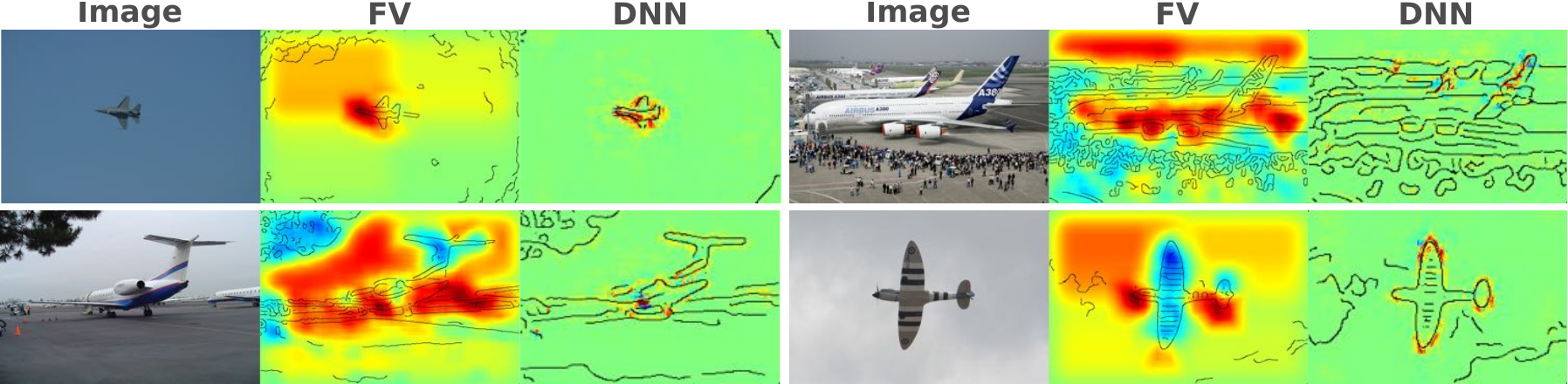}
\caption{Example heatmaps for class ``aeroplane'', for which both the DNN and the FV model predict well. The attention of the DNN model generally concentrates on the body of the airplanes. The FV model also has learned to recognize the airplanes themselves reliably, presumably due to the pronounced edges and the contrast-rich geometry, but still uses the sky in both topmost quadrants of the image for support. Negative relevance scores are attributed to areas of the top half of the image where the clear sky obstructed by an object, such as a tree or parts of the airplane itself.}
\label{fig:aeroplane}
\end{figure}

Of special interest are our findings related to the categories ``boat'' and ``horse'' (\textbf{boa} and \textbf{hor} in Table \ref{tab:cvprresults}): Both classes are amongst those for which both models predict with high confidence. No statistically significant difference can be seen for the class ``horse''. Figure \ref{fig:CVPR} presents results for both object classes. To the left, input images show boats alongside pixel-level relevance maps computed for the corresponding predictors, visualized as heatmaps. We observe that for the class ``boat'', the FV model dominantly decides depending on the presence of a body of water in the bottom of the image. The DNN however has based its decision on the silhouette of the boat itself, in line with human intuition. When an image of a shipwreck on land is passed as input to both classifiers, the relevance maps computed for class ``boat'' show that the DNN still finds positive evidence in the outline of the wreck, whereas the FV model detects evidence strongly contradicting the target class ``boat'', predicting the label ``aeroplane'' instead.
The observation that the FV model consistently predicts boats based on the presence of water holds for almost all samples in the dataset. For other object classes, the FV model has learned a similarly biased contextualized prediction rule. The prediction of object category ``aeroplane'', for example, relies heavily on sky as background in the top left and top right quadrants of the image. Figure \ref{fig:aeroplane} shows some examples for class ``aeroplane'', for which both models predict with high average precision (AP) rating.

Most notably, the FV model has also developed a strong bias towards making use of a copyright watermark to detect samples of class ``horse'', as shown in the right half of Figure \ref{fig:CVPR} --- notably this artifact had gone undetected when establishing the benchmark. Removing the watermark in the second row of the figure results in a false negative prediction of the FV model. In the explanatory heatmap computed using LRP, the aggregation of strong positive relevance in the area of the copyright watermark vanishes as well, while the remainder of the heatmap remains unchanged. The neural network model is invariant to this input change in prediction and heatmap response. Upon closer inspection of the Pascal benchmark data on which the FV model has been trained and evaluated, similar copyright watermarks were located in approximately $20\%$ of all images containing horses in the bottom, or bottom left of the image. A large part of the remaining benchmark images with horses were taken in a horse show tournament setting, where obstacles and hurdles being jumped over are identified as decisive image features representing the class ``horse''. The presence of simple and easy to learn visual features which are not part of the animal horse explain the classifier's strong focus on contextual information at prediction time, and the relatively low amounts of relevance attributed to the pixels showing the horses themselves (cf.\ also the following Section \ref{sec:importanceofcontext}
for an investigation of the importance of image context for both the considered FV and the DNN models).

Such artifactual learning has also been described in other work, e.g, in \cite{DBLP:conf/kdd/Ribeiro0G16} where the model leaned to distinguish wolves and Huskies by the presence of snow, or in the context of reinforcement learning \cite{amodei2016faulty} where the RL agent finds an isolated lagoon where it can turn in a large circle and repeatedly knock over three targets, timing its movement so as to always knock over the targets just as they repopulate due to the misspecify of the reward function. All these are concrete problems on AI safety, which has been recently studied in \cite{leike2017ai}.

\begin{figure}[t]
	\centering
 \includegraphics[width=0.9\textwidth]{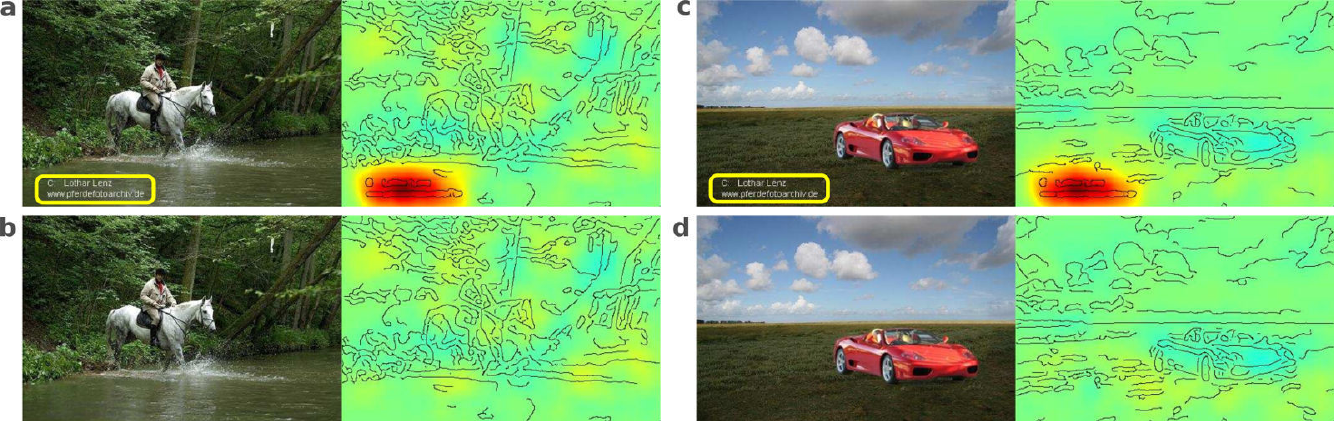}
 \caption{Images used as input for a FV model trained to detect horses, next to the corresponding relevance map. a) shows an image from the Pascal VOC 2007 dataset containing a copyright watermark, causing a strong response in the model. In b), the watermark has been edited out. The artificially created images c) and d) show a sports car on a lush green meadow with and without an added copyright watermark. In samples a) and c) the presence of class ``horse'' is detected, whereas in samples b) and d) this is not the case.}
  \label{fig:horseCar}
\end{figure}

We provide another example for the extent of the influence the learned bias has for prediction. Figure \ref{fig:horseCar}a shows an image belonging to class ``horse'' as it originally occurs within the Pascal VOC 2007 data. Using the FV model to predict the class of the image results in a true positive prediction for class ``horse''. Again, the corresponding relevance map identifies the copyright watermark in the bottom left corner as the most important feature for the classifier to make its decision. When the copyright tag is removed (Figure \ref{fig:horseCar}b), however, the model fails to recognize the sample as a member of its true class.

Conversely Figure \ref{fig:horseCar}d shows an artificially created image, where a sports car has been placed on a lush green meadow, causing a true negative prediction by the FV model. By adding the copyright watermark corresponding to class ``horse'' to the image (Figure \ref{fig:horseCar}c), the model recognizes the tag and false positively classifies the image as ``horse''. That the learned biases for the classes ``boat'' and ``horse'' are global phenomena for the FV model reflects well in the bottom row of Figure \ref{fig:CVPR}. For both object classes, average heatmaps for the FV classifier (left) and the DNN (right) are shown. Whereas for the FV model, positive relevance tends to aggregate in the same area --- the bottom of the image for class ``boat'' and the bottom as well as the bottom left corner for class ``horse'' where the copyright watermark is located --- the average relevance response computed for the DNN classifier shows distinguishable localized patterns around the image center, recapitulating our observations on a class-wide scale and underlining that the DNN model predicts based on the true object appearance, which shifts and differs from image to image.

\subsubsection{Importance of Context} 
\label{sec:importanceofcontext}
\begin{figure}[t]
\centering
\includegraphics[width=0.9\textwidth]{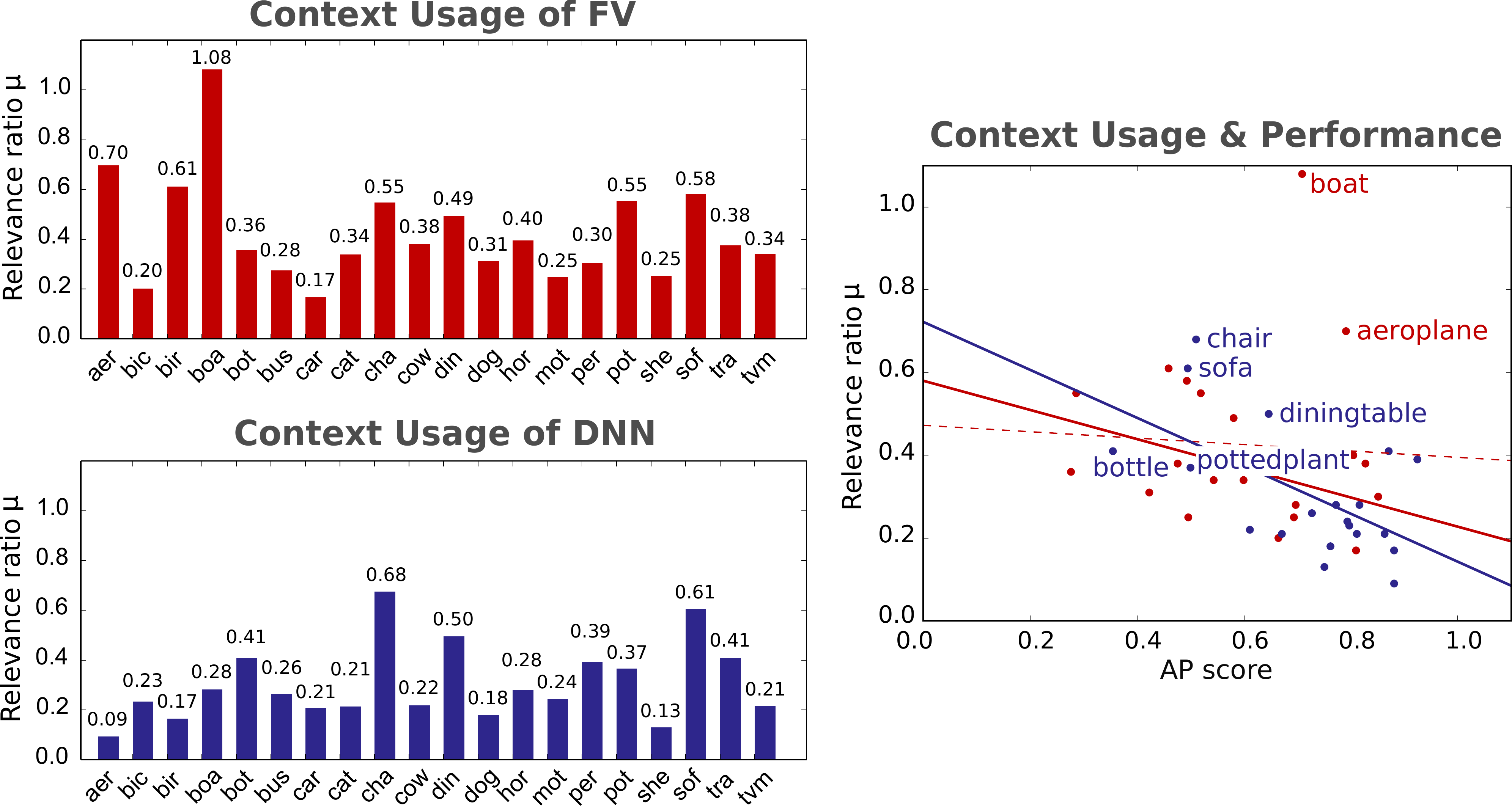}
\caption{{\it Left}: Measurements reflecting the importance of image context per class and model. Higher values correspond to on average large amounts of positive relevance being located outside the object bounding boxes. {\it Right}: Red dots show the relationship between AP score and importance of context per class for the FV model. Blue dots show the same results for the DNN. The red solid line tracks the linear relationship between AP score and importance of context $\mu$ for the FV model when ignoring the outliers (classes ``aeroplane'' and ``boat''). The dashed red line shows the trend for all $20$ classes. The blue line shows the trend for all classes and the DNN model. The named blue dots are classes for which the DNN model uses relatively high amount of context information due to feature sharing, caused by exceptionally high co-occurrence rates in the class labels.}
\label{fig:context}
\end{figure}

To quantify above observations, we measure image context importance against object importance for prediction, per class and for both the FV and the fine-tuned DNN model. The image annotations of the Pascal datasets provide coordinates for tightly fit object bounding boxes for each image. For each object class of the dataset, we collect all positively labelled samples and compute an average outside-inside relevance ratio per image as
\begin{equation}
\mu = \frac{|P_\text{out}|^{-1}\sum\limits_{q\in P_\text{out}}R_q}{|P_\text{in}|^{-1}\sum\limits_{p\in P_\text{in}}R_p}
\end{equation}
where $P_\text{in}$ is the set of pixels inside the class-specific object bounding boxes (multiple instances of an object class can appear simultaneously) with positive relevance scores, $P_\text{out}$ covers the set of pixels with positive relevance scores outside of the bounding boxes and $R_p$ and $R_q$ are relevance values for pixel coordinates $p$ and $q$. Large values indicate model decisions being largely based on background information, whereas low values speak for object-centric predictions. Figure \ref{fig:context} reports the average ratios per class over all samples representing that class.

\begin{figure}[ht]
\centering
\includegraphics[width=0.895\textwidth]{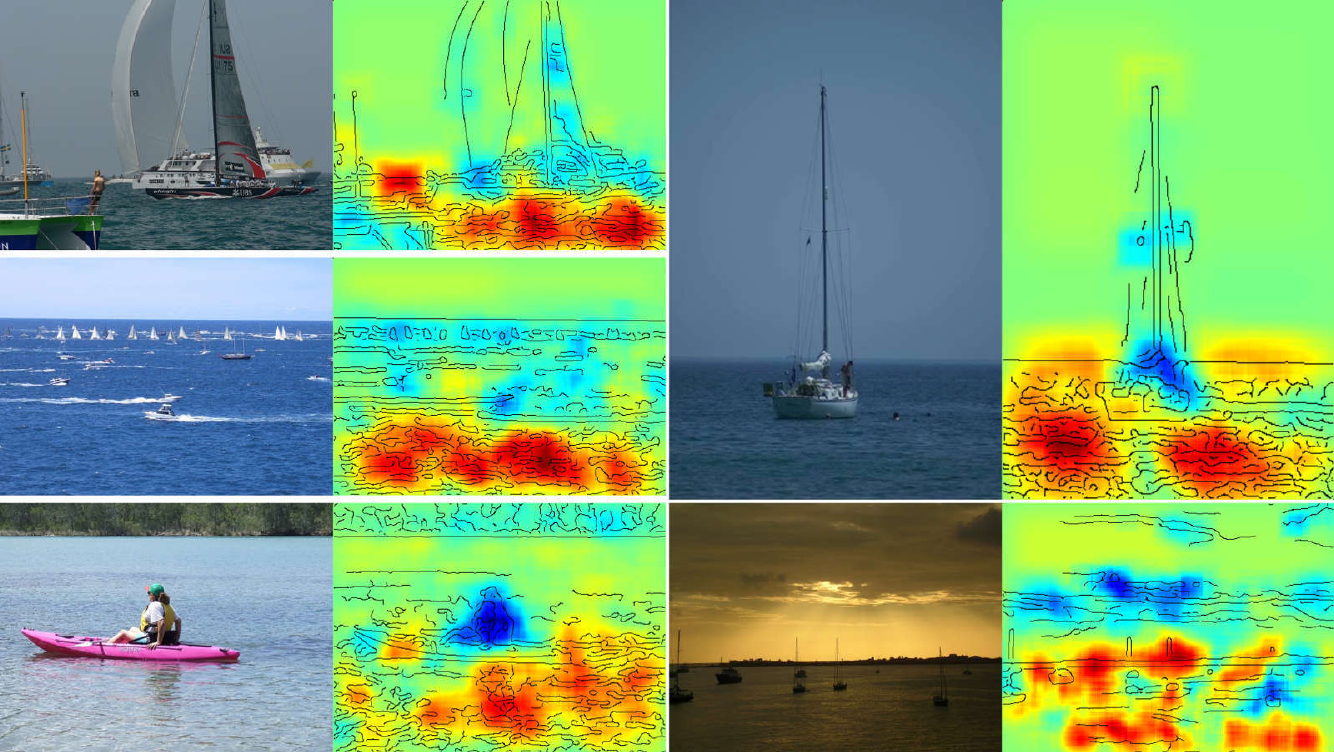}
\caption{Several images showing boats below the horizon line, with predictions explained for the ``boat''-class model. In all images, the water itself receives strong positive relevance scores. \textit{Top Left} and \textit{Top Right}: Both sail boats are located in front of the horizon, receiving negative relevance scores in the respective image areas. Unobstructed horizon lines in both images yield information indicating the presence of class ``boat''. \textit{Middle Left}: Distant sails and motor boats receive negative relevance scores, while the water surface in the bottom part of the image strongly indicates the target class. \textit{Bottom Left}: The features extracted for the human visually blocking the water receive strong negative relevance values, even though class ``person'' frequently appears together with class ``boat'' (See Figure \ref{fig:coocurrence}). The visual information obtained from the image area showing the person strongly contradicts the learned concept of class ``boat''. \textit{Bottom Right:} In this scene, masts crossing the horizon and the water itself count towards the concept of ``boat'', while some vessels below the horizon --- both boats to the middle left aligned with the camera taking the image and both boats to the right --- are yielding disruptive visual features. See also Figure \ref{fig:evenmoreboats} for complementary heatmaps computed for class ``boat'' based on the DNN model.}
\label{fig:moreboats}
\end{figure}

\begin{figure}[ht]
\centering
\includegraphics[width=0.895\textwidth]{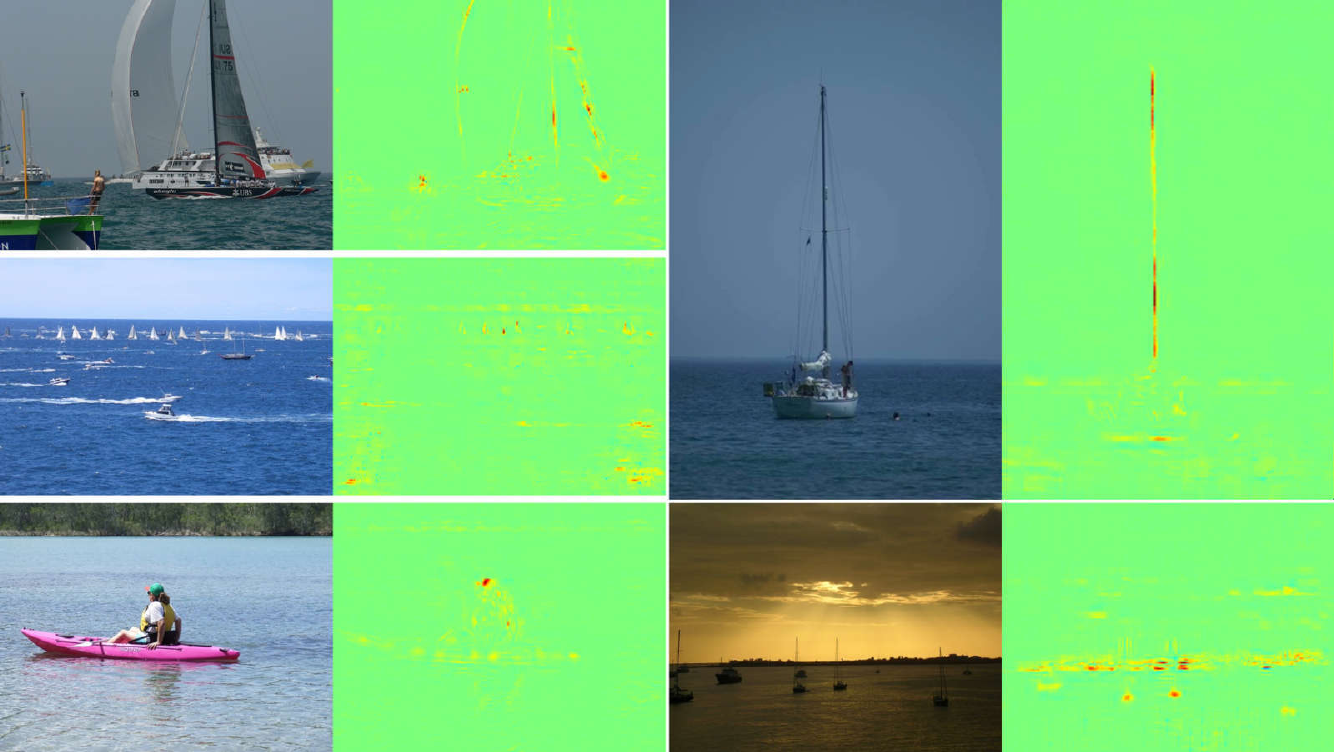}
\caption{Figure as a counterpart for Figure \ref{fig:moreboats}, showing the additional relevance responses for class ``boat'' for the fine-tuned DNN model. Due to the filigree character of the relevance response no edge maps have been drawn on top of the heatmaps. The relevance scores follow the boat-like features very closely, rendering the heatmaps well-readable and the relevance scores easy to localize. In contrast as shown by the heatmaps for the FV model in Figure \ref{fig:moreboats}, the DNN does not regard boats below the horizon line as disrupting objects. Frequently, even people sitting or working (sail) on boats are rates as contributing factors to class ``boat'', which can be attributed to the label ``person'' appearing in almost every third ``boat'' image (see Figure \ref{fig:coocurrence}).}
\label{fig:evenmoreboats}
\end{figure}

Both bar charts confirm that the DNN predicts by using visual information provided by the class-representing objects themselves, while the FV model resorts more to using contextual information. Well-predicted classes such as ``aeroplane'' and especially ``boat'' are predicted based on a consistently occurring contextual bias. For these classes the model might predict the label correctly in a majority of trials, i.e.\ the model performs well \textit{in number}, but it has not learned to generalize the concept well and will most likely fail when deployed for a task outside the controlled benchmark laboratory setting. Figure \ref{fig:moreboats} for example shows boats in water, partially below the horizon line, yielding negative relevance values, since their presence occludes the water. Figure \ref{fig:evenmoreboats} demonstrates, that the DNN classifier does not show this effect.

\subsubsection{Why do FV and DNNs exhibit different strategies ?} 
We attribute this effect to a combination of circumstances: For one, the FV mapping step includes the subdivision of the input image into multiple image sub-areas, a technique known to considerably boost the predictive performance of Bag of Words models due to the incorporation of weak geometric information. In this mapping scheme, one FV representation is computed for each image sub-area, spanning a separate set of dimensions in the final image representation in vector form. This encourages the classifier to optimize by concentrating on the subset of input dimensions providing information which correlates most with the expected label. Secondly, the Pascal VOC 2007 training dataset is comparatively small and lacks diversity. Some classes are under-represented or class labels frequently co-occur. The lack in diversity is common cause for the development of prediction biases, while object co-occurrence will cause the model to also learn features from other classes, located outside the target class object's bounding box area.

The DNN predicts dominantly based on the objects themselves. Therefore, most class context values reported in Figure \ref{fig:context} (left) are much lower for the DNN than for the FV model. The DNN has been fine-tuned on Pascal VOC data and moreover many of the 1000 classes from the ImageNet challenge semantically overlap with the classes present in the Pascal VOC data. The class ``horse'' for example corresponds to the (semantic) subcategories ``zebra'' and ``sorrel'' in the ImageNet label set. While Pascal VOC provides the class ``cat'', ImageNet has a range of labels describing subtypes of cats (``tabby'', ``burmese cat'', ``manx'',~$\dots$). This visually and semantically similar overlap allows the network to resort to robust prior knowledge accumulated during training for the comparatively larger and more diverse ImageNet. This prior knowledge --- the learned set of filters insensitive to the biases, but sensitive to the appearance of the objects in the Pascal VOC data --- benefits the fine-tuning and results in a model which --- loosely speaking --- merely has to adapt its outputs to the new semantic feature groupings in the Pascal VOC label set.

\begin{figure}[!tb]
\centering
\includegraphics[width=0.7\textwidth]{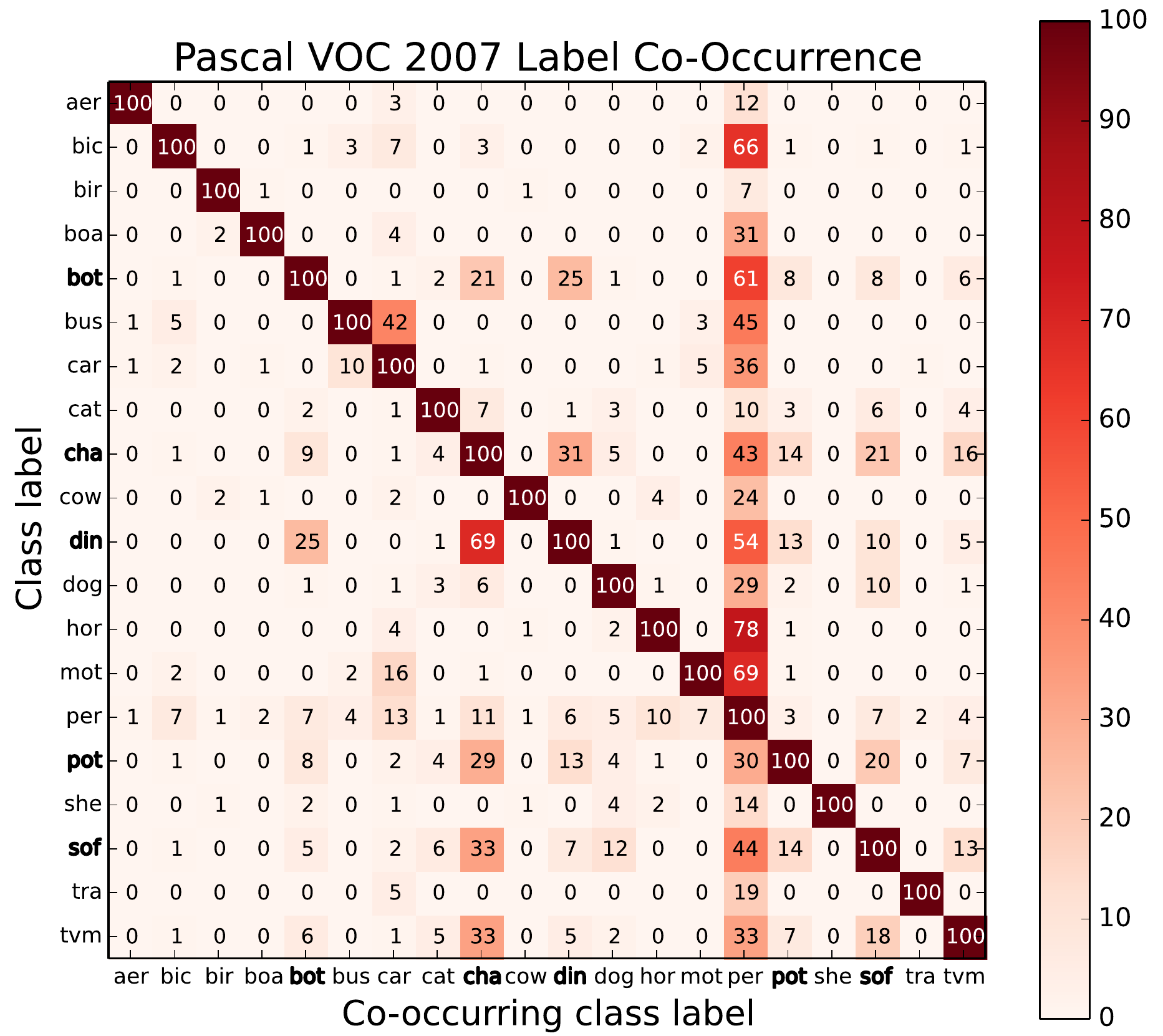}
\caption{A matrix showing class label co-occurrences for the Pascal VOC 2007 data in percent. The entries visualize the rate at which the classes at the columns appear together with the classes indexing the rows of this matrix. Most notably is the frequent appearance of class ``person''. Also the ``living room'' classes (``bottle'', ``chair'',``diningtable''.``pottedplant'',``sofa'') often share images, explaining the high use of contextual information reported in Figure \ref{fig:context}. For the Pascal VOC 2012 dataset, differences in co-occurrence rates are only marginal.}
\label{fig:coocurrence}
\end{figure}

There are, however, outlying classes in the measurements shown in Figure \ref{fig:context} (left) for the DNN model, namely the classes ``chair'', ``diningtable'' and ``sofa'' and other classes describing interior items. This set of labels often co-occurs in the images of the Pascal datasets, showing cluttered indoor scenes (see Figure \ref{fig:coocurrence}). The frequently and simultaneously present set of features makes it difficult for both models to distinguish between the present classes as individual categories. Furthermore, from the results in Figure \ref{fig:context} (left) and Table \ref{tab:cvprresults} we may conjecture that performance may indeed correlate with ``object understanding'', as shown in Figure \ref{fig:context} (right) which combines both results. Equally speculating, we consider the results for the neural network agent playing Atari games as shown in Figure \ref{pulseAndHeatmaps} suggestive for the hypothesis that tight groupings of relevance scores on intuitively important image elements indicate a well-performing predictor even beyond typical classification tasks.

\section{Semi-automated Analysis of Classifier}
\label{sec:analysis}

This section demonstrates the usefulness of relevance maps as features for analyzing classifier behavior automatically. The proposed approach constitutes a new tool for semi-automated model analysis -- based on the computation of relevance maps \cite{10.1371/journal.pone.0130140} and spectral analysis \cite{von2007tutorial} -- which closes the gap between
\begin{itemize} 
\item (classical) performance evaluations over whole datasets (e.g.\ error rates). Here, no human inspection is needed, but this approach does not provide any information about the inner working of the classifier. 
\item visual assessment of single predictions using methods such as LRP. Here, information about the decision process is provided,
but this analysis requires human attention for each data point and thus does not scale to whole datasets.
\end{itemize} 
We name our novel analysis method \emph{Spectral Relevance Analysis} or short SpRAy. 
To the best of our knowledge it is the first method to investigate the predictions strategies of the classifier on the whole dataset in a (semi-)automated manner. 

Technically, our novel technique SpRAy allows to detect predictions based on irregularly frequent reoccurrence of non-obvious and highly similar image features, which can assist in the uncovering of intriguing or suspicious patterns in either the data used for training (and testing), the prediction strategies learned by the model, or both. The identified features may be benign and truly meaningful representatives of the object class of interest, or they may be co-occurring features learned by the model but not intended to be part of the class, and ultimately of the model's decision process. In the latter case, SpRAy will assist researchers and software engineers to systematically find weak points in their trained models or training datasets, or even generate new knowledge about the data and problem solving strategies.

\subsection{Spectral Relevance Analysis}
\label{sec:spray}
The predictions made by the FV model for class ``horse'' are based on image features frequently appearing in a very structured manner in the object class' image data, despite not describing the animal in any way visually. The artifact in question is a copyright tag in approximately one fifth of the test and training images representing the object class, which has been picked up by the model as an even stronger indicator of horseness than the pictured horses themselves. Since the copyright watermark is already present in the input data, a \emph{very} attentive human observer might have observed it, yet since those samples found their way into the Pascal data, which also was widely used as a benchmark dataset for several years, we suppose ultimately no one did. Humans often solve tasks in certain ways predisposed by life (long) experience. Any assessors of the data might thus have directed their attention towards the intended objects (known through life experience) shown on the image samples instead of auxiliary and unexpected features. Machine learning algorithms on the other hand can only connect information which is available in the discrete samples of finite example datasets.

With the goal to systematically identify pitfalls in the data a model might -- or in fact does -- fall victim to, we establish a novel \emph{Spectral Relevance Analysis} (SpRAy) pipeline, which finds features correlated with a target class. Although those features appear during correlation analysis between pixel space and the target label, they might not be obvious to a human observer at all. If said features are very descriptive of a prediction target or object class, their presence is expected to result in concentrated relevance feedback within the corresponding models' and samples' relevance maps.

For our analysis, we employ the following steps: 
\begin{enumerate}
\item Computation of the relevance maps (potentially over the whole dataset).
\item Preprocessing (downsizing, padding, normalizing ...).
\item Spectral cluster analysis (SC) on the relevance maps.
\item Identification of interesting clusters by eigengap analysis.
\item Optional: Visualization by t-Stochastic Neighborhood Embedding (t-SNE).
\end{enumerate}

After computing the heatmaps for multiple samples from the training / test dataset, the initial preprocessing step (2) renders all relevance maps uniform in shape and size, creating viable inputs for the following application of SC and t-SNE. Some predictors such as the FV classifier can accept as inputs images of arbitrary size, making this step a necessity in some cases. Also, by downsizing the input relevance maps, e.g.\ by pooling relevance spatially wrt.\ a regular grid, the dimensionality of the analysis problem is decreased (in general, this step is practically useful but not absolutely necessary). This expedites the process considerably and produces more robust results. Suitable preprocessing choices can be used to focus the following steps on certain characteristics within the input relevance maps.

The SC \cite{meila2001random,ng2002spectral,von2007tutorial}  used in step (3) is a clustering algorithm with interesting analytic properties, which constitutes the foundation of the analysis pipeline. The method relies on the computation of a weighted \emph{affinity} or \emph{adjacency} matrix $W = (w_{ij})_{i,j=1,\dots,N}$, measuring the similarity $w_{ij}>=0$ between all $N$ samples $s_i$ and $s_j$ of a dataset. E.g.\ when building affinity matrices based on $k$-nearest-neighborhood relationships as we do in Section \ref{sec:spray-uncovering}, we compute $w_{ij}$ as
\begin{align}
w_{ij} =
\begin{cases}
1 	&~;~ s_j~\text{is among the}~k~\text{nearest neighbors of}~s_i  \\
0				&~;~\text{else}
\end{cases}
\end{align}
Since this is an asymmetrical relationship between $s_i$ and $s_j$ we follow \cite{von2007tutorial} and create a symmetric affinity matrix $W$ with $w_{ij} = \max(w_{ij}, w_{ji})$, describing an undirected adjacency graph between all samples. The evaluations in Section~\ref{sec:spray-uncovering} (which are based on binary neighborhood relationships) yielded highly similar results when using (euclidean) distances $||s_i - s_j||$ between connected neighbors instead of above binary neighborhood relationships with only small differences in eigenvalue spectra.

\begin{figure}[t]
\centering
\includegraphics[width=0.75\textwidth]{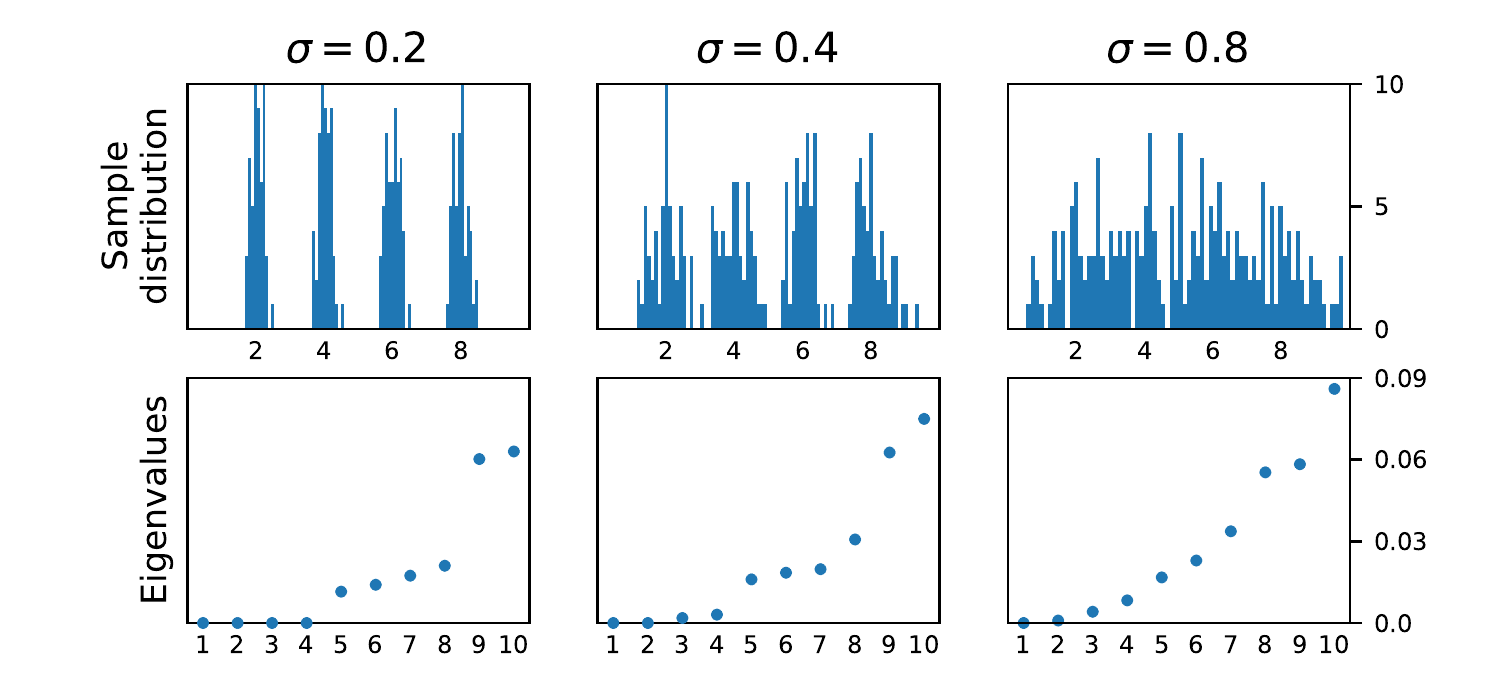}
\caption{Histograms of datasets containing samples from standard normal distributions with corresponding standard deviations $\sigma$, centered with $\mu \in \lbrace 2,4, 6, 8\rbrace$ and the 10 smallest eigenvalues each. Inspired by Figure~4 from \cite{von2007tutorial}.}
\label{fig:eigentutorial}
\end{figure}

From the matrix $W$, a graph Laplacian $L$ is then computed as 
\begin{align}
d_i = \sum\limits_j w_{ij}\\
D = \mathrm{diag}([d_1, \dots, d_N])\\
L = D - W
\end{align}
with $D$ being a diagonal matrix with entries $d_{ii}$ describing the \emph{degree} (of connectivity) of a sample $i$. Performing an eigenvalue decomposition on the Laplacian $L$ then yields $N$ eigenvalues $\lambda_i,\dots,\lambda_N$, with the guarantee of $\min_i \lambda_i = 0$, and a set of corresponding eigenvectors. Here, the number of eigenvalues $\lambda_i = 0$ identifies the number of (completely) disjoint clusters within the analyzed set of data. For non-empty datasets, there is at least one cluster and thus the guarantee for at least one eigenvalue to be zero-valued. The final step of cluster label assignment can then be performed using an (arbitrary) clustering method -- e.g.\ k-means -- on the $N$ eigenvectors. On non-synthetic data, however, there seldomly are cleanly disjoint groups of samples, and the (main) cluster densities usually are at least weakly connected. In that case,  the main densities can be identified by eigenvalues close to zero as opposed to exactly zero, followed by an \emph{eigengap}. The eigengap is a sudden increase in the difference between two eigenvalues in sequence $|\lambda_{i+1} - \lambda_i|$. In Figure~\ref{fig:eigentutorial} we recreate an example from \cite{von2007tutorial} which demonstrates the change in the eigenvalue spectrum with increasing overlap between the sampled distributions.

With $\sigma=0.2$, four non-overlapping sample sets have been generated. The four smallest eigenvalues of this dataset thus are exactly zero and the eigengap is distinctive. With increasing overlap of the sampled distributions ($\sigma=0.4$), the eigengap still indicates four clusters of data clearly, although the four smallest eigenvalues indicating well defined groups of samples are not exactly equal to zero anymore. Once the distributions overlap considerably ($\sigma=0.8$), there is no well-distinguished eigengap anymore to identify the number of densities sampled from truthfully. We can use the eigengap property of an analyzed dataset to identify structural similarities in the image input data of an object class, but also the corresponding relevance maps computed for each sample of the dataset wrt.\ to a trained model, reflecting the model's prediction strategy (step (4)).

In step (5) t-SNE \cite{maaten2008visualizing} is used to compute a two-dimensional (or in general, a human-interpretable lower dimensional) embedding of the analyzed data, based on pair-wise distances between samples. The resulting embedding is used to visualize groups of input images or relevance maps by similarity, aiding in the identification of interesting patterns in the models' prediction. Properties of the t-SNE are known to be connected to cluster assignments computed via SC \cite{linderman2017clustering}. Sample to sample relationships computed during the SC step can be re-used in the t-SNE embedding phase by transforming the affinity matrix $W$ to a distance matrix, e.g.\ as $\text{DIST} = \frac{1}{W + \epsilon}$ for above binary k-nearest-neighbor affinity matrix by adding a small $\epsilon$ to the denominator to encode high distances between (via k-nearest-neighbors) unconnected points.

\subsection{Uncovering Prediction Strategies of the Pascal VOC Classifiers}
\label{sec:spray-uncovering}
Searching to uncover distinct prediction strategies within the scope of single object classes, we analyze for all classes of the Pascal VOC 2007 test set separately. Given the extensive amount of options to configure the SC algorithm, we perform our analysis using the recommended parameters from \cite{von2007tutorial}, i.e.\ we build neighborhood graphs using the $k$-nearest neighbor algorithm with $k = log(n)$ and $n$ being the number of samples and use normalized, symmetric and positive semi-definite Laplacians \cite{ng2002spectral} $L_\text{sym}$ instead of the unnormalized $L$. The use of $L_\text{sym}$ also allows for meaningful comparisons between input sample sets of different sizes, which will be important in a moment with Figure \ref{fig:eigenval-barchart}. For t-SNE, we set the perplexity to 7, the parameter for early exaggeration to 6. The method is known to be insensitive to small changes in both input parameters. For downsizing images, we use standard interpolation algorithms for images and sum pooling over a regular grid of pixels for relevance maps to reduces the inputs to $20\times 20$ sized matrices, representing local relevance allocation behavior throughout the dataset. Our experiments have shown that the qualitative results as presented in below setting can also be reliably obtained for input sizes $5 \times 5$, $10 \times 10$ and $50 \times 50$.

\begin{figure}[t]
\centering
\includegraphics[width=0.8\textwidth]{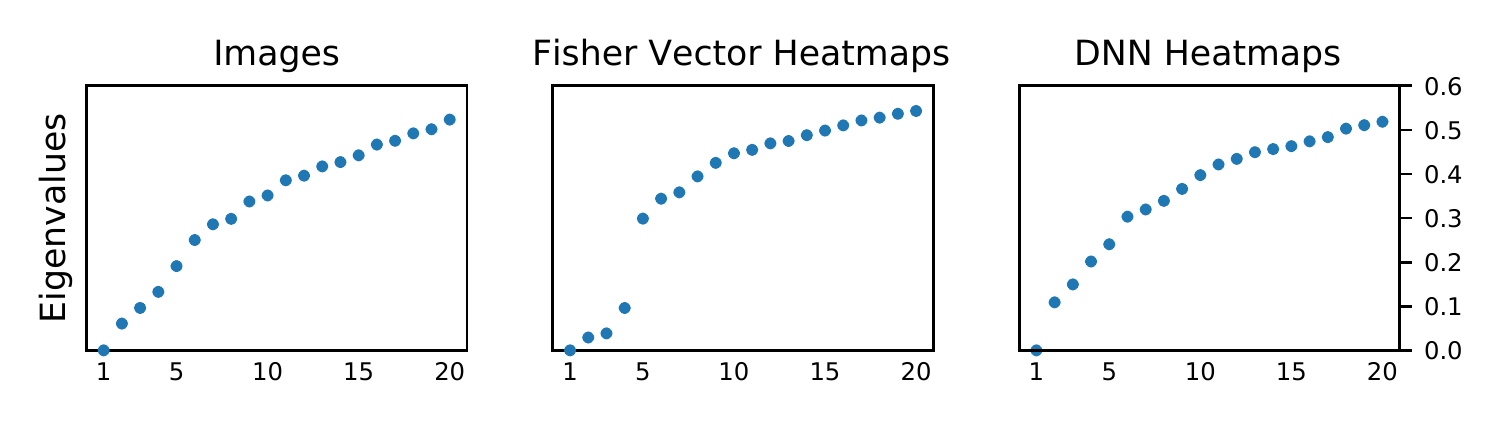}
\caption{The first 20 eigenvalues for class ``horse'' for images, relevance maps for the Fisher Vector predictor and the DNN respectively.}
\label{fig:eigenvaluespectrum-horse}
\end{figure}

In the following we inspect the eigenvalue spectra for all classes and both the FV and DNN model. Figure~\ref{fig:eigenvaluespectrum-horse} shows the eigenvalue spectrum for class ``horse'' for input images and relevance maps for both the FV and DNN model. Previous manual inspection has revealed that the FV predictor primarily relies on a copyright watermark embedded into the pixels of some images for predicting the class. The gaps after $\lambda_3$ and $\lambda_4$ in the eigenvalue spectrum of the FV model's relevance maps are apparent and indicate multiple well-separated clusters. In the eigenvalue spectra of the images and the relevance maps obtained with the DNN model, the largest gap between eigenvalues occurs after $\lambda_1$, indicating only one densely connected cluster over all test samples for that object category.

\begin{figure}[!h]
\centering
\includegraphics[width=\textwidth]{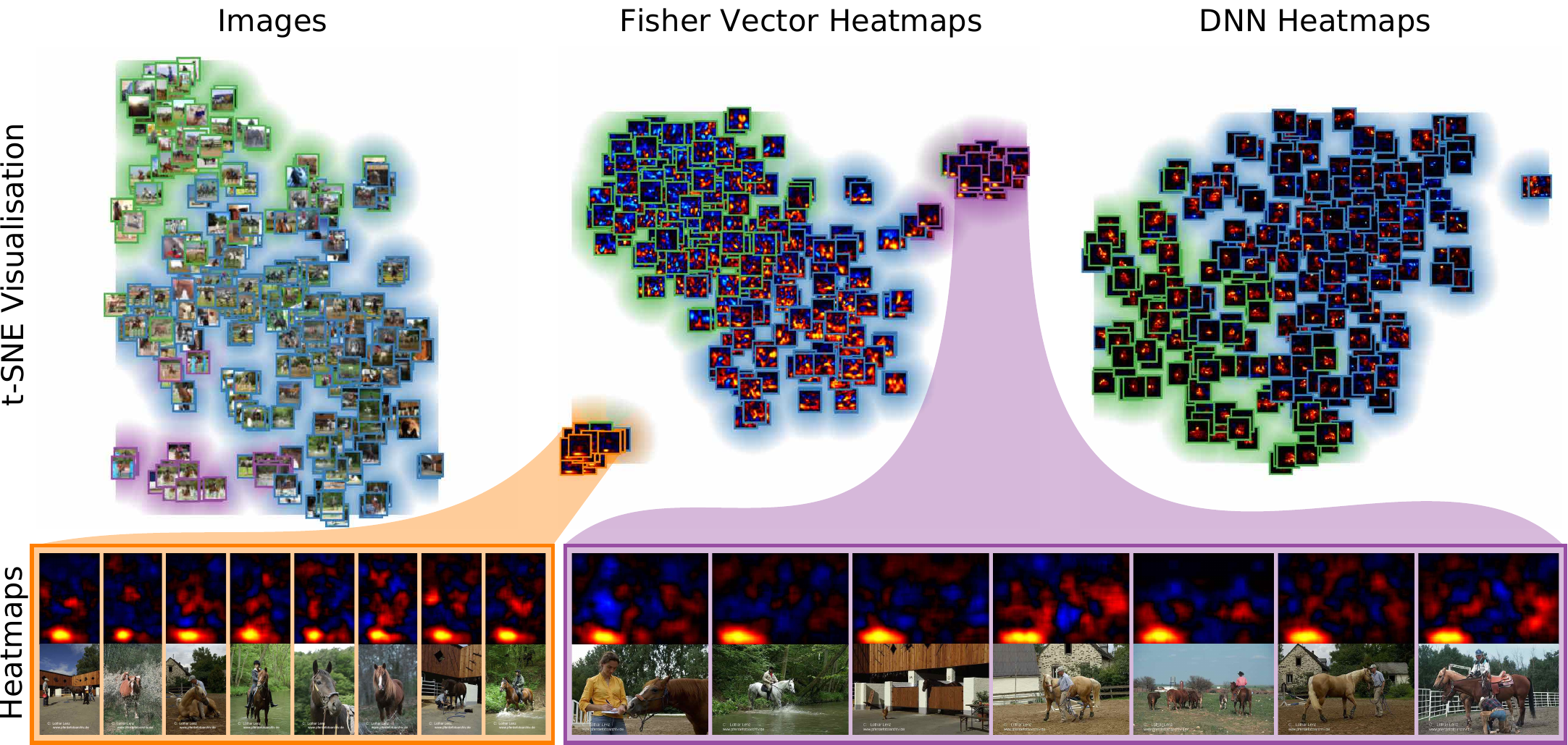}
\caption{Cluster label assignments for class ``horse'' via SC computed from input images (left), FV relevance maps (middle) and DNN relevance maps (right). Embedding coordinates in $\mathbb{R}^2$ for visualization have been computed on pair-wise distances derived from the weighted affinity matrix $W$ used for SC. \emph{Top row:} The data (images or relevance maps) used as input for the analysis at their respective two dimensional embedding coordinates. The colored borders and aurae around the input samples show cluster label assignments via SC. \emph{Bottom row:} The enlarged samples and FV heatmaps (without preprocessing) show examples grouped into the bottom left cluster (portrait oriented images) and the top right cluster (landscape oriented images), revealing the FV models' strong reaction to the copyright watermark.}
\label{fig:tsne+labels-horse}
\end{figure}

Likewise to the eigenvalue spectra in Figure \ref{fig:eigenvaluespectrum-horse}, the cluster label assignments computed via SC as displayed in Figure \ref{fig:tsne+labels-horse} clearly show point clouds of homogeneous density for both DNN relevance maps and input images and well separated clusters for the FV relevance maps. Figure \ref{fig:eigenvaluespectrum-horse} also demonstrates that the clusters found with SC (color coded) in the input data correspond well to the embeddings computed by t-SNE (embedding locations in $\mathbb{R}^2$) from the same input data with very high similarity. The connection between SC and t-SNE described in \cite{linderman2017clustering} can also be observed in all following results, making t-SNE a suitable tool for visually supporting the analysis via SC.

While for the DNN relevance maps the embedding and clustering results seem to be grouped by dominant combinations of horse and rider in the foreground vs.\ the presence of contradictory image features, the results for color images as input features are determined by visual scene similarity in structure and color composition (Figure \ref{fig:tsne+labels-horse}). For the Fisher Vector relevance maps, the presence of two well defined groups besides the main cluster can be identified by both the cluster labels assigned via SC and also the embedding locations computed with t-SNE. This result suggests that the FV model picks up on information which is apparently correlated strongly with the target label ``horse'', but not obvious from the input images alone.
Both clusters separated from the main point cloud contain relevance maps for image samples with the copyright watermark in question, which only covers a small number of pixels within the images originating from the same horse-themed online image repository.
The blue cluster groups samples which are predicted dominantly based on the presence of hurdles in a tourney setting. For a zoomed-in view of affected samples and their respective relevance maps from that cluster, please refer to Figure 3 in the main paper and Figure~\ref{fig:hurdles} below.

With the application of SpRAy on the input images and relevance maps computed for both models, we have gained a clue that the FV model picks up on input information  in a structured and consistent way, which is not obviously present in the image domain and not used by the DNN model (the copyright watermark).
Further results reveal predictions based on image features co-appearing with the target class ``horse'' (the hurdles), which are not intended to represent instances of the target object.
That is, on relevance maps we obtain clusters of samples which are predicted with a very similar localized strategy by the FV model, which we may not find when inspecting the raw input samples.
The model has learned prediction strategies suitable for small subsets of training and test samples based on image features which are hard to find on the images directly, or are not intended to represent the target class. The usage of such a classifier that is relying on spurious correlations would lead to overfitting when applied to real world data, where spurious correlations appear much more rarely.

A recent and interesting study~\cite{rajalingham2018large} compares behavioural patterns of primates (humans and rhesus macaque monkeys) and recent state-of-the-art DNNs in an image prediction setting.
The authors systematically investigate several image attributes and characteristics and their effect on prediction performance in a carefully controlled setting, in order to explain the performance gap between primates and the less well performing DNN models.
One component of the analysis presented in \cite{rajalingham2018large} explores the difference of {\em prediction signatures} between e.g.\ DNN and primate. Signatures may be built for each model from the class-level or example-level recognition performance. Another component of the analysis in \cite{rajalingham2018large} focuses on the effect of {\em predefined visual cues} on recognition performance.
Among the considered visual cues are pixel attributes (e.g.\ mean image luminosity) and object viewpoint attributes (e.g.\ object size and pose).
Results of the analysis in \cite{rajalingham2018large} suggest that these predefined cues do not exhaustively explain the whole prediction difference, and that there are potentially other (unknown) factors at play, that prevent the analyzed DNN models from attaining primate level image recognition performance.

\begin{figure}[t]
\begin{center}
\includegraphics[width=\textwidth]{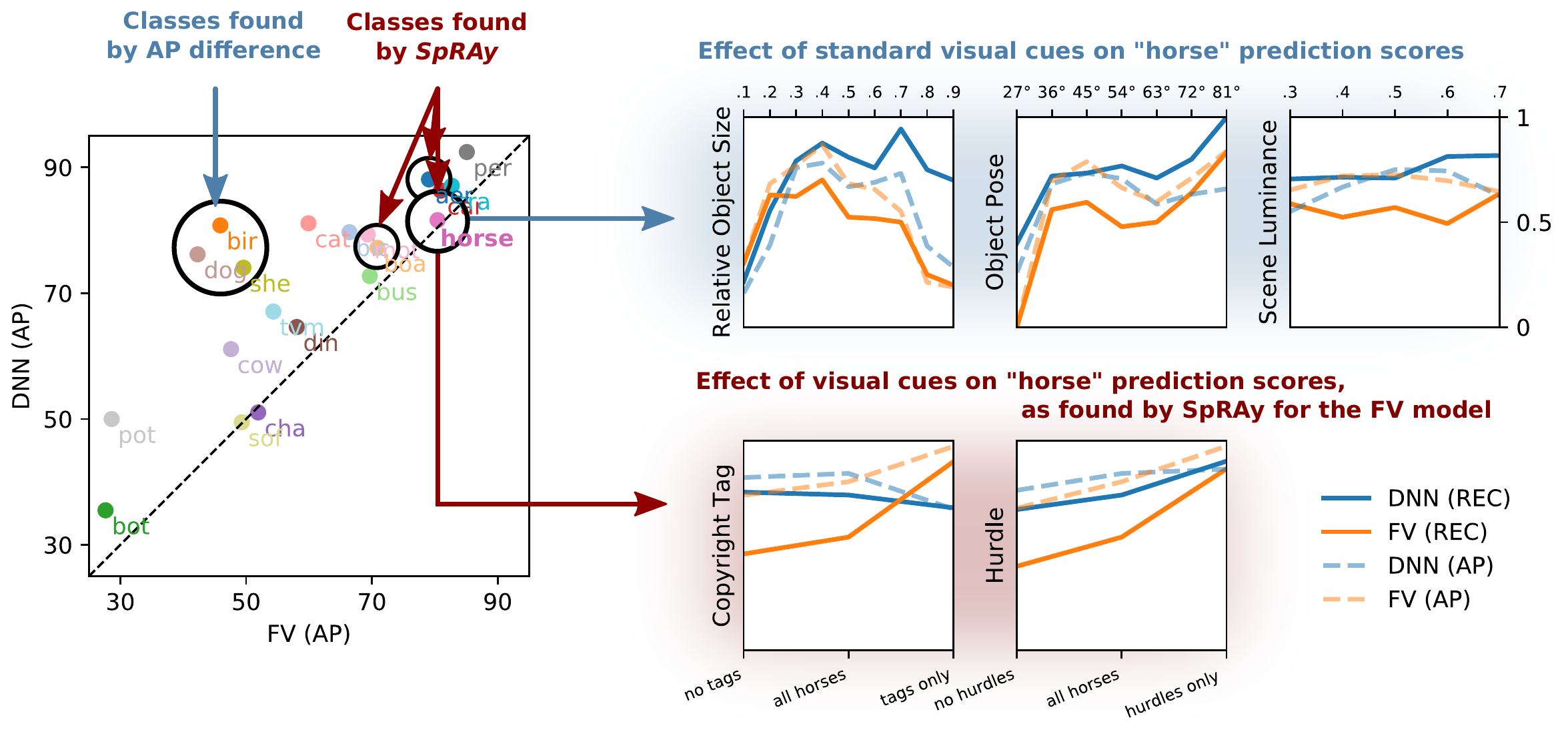}
\end{center}
\caption{\textit{Left:} An analysis based on model outputs (e.g.\ here prediction performance in AP) can be used to only identify classes which are predicted with different levels of success between considered models.
Via the analysis of the eigenvalue spectrum, SpRAy can identify classes and corresponding features which cause characteristic \emph{prediction behaviour} for a model, despite comparable model performances.
\textit{Right:}
Quantifying the effect of a feature to the predictor output \emph{in order to identify} whether the feature is indicative for model performance~\cite{rajalingham2018large} can not rule out the unintended measuring of the effect of another correlating feature.
Here, the features ``hurdle'' (identified by SpRAy for the FV model only) and ``pose'' seem to affect DNN prediction performance.
Figure~\ref{fig:hurdles} reveals that the DNN is focussing on the horse jumping over the hurdle, along with its rider, not the hurdle itself.
Object pose as estimated via the bounding box dimensions are a by-product of the horse frontally photographed mid-jump.
}
\label{fig:dicarlo-vs-spray}
\end{figure}

We recreate the experiment described in~\cite{rajalingham2018large} on our Pascal VOC setting to test if class-level prediction signatures are capable of identifying differences of classification behavior between the considered models, and whether the additional visual cues found by SpRAy (e.g.\ the source tags) are capable of further explaining this difference of classification behavior between models. We consider the FV model and compare it to the DNN model which here plays the role of the `ground truth' due to its higher overall performance. The class-level signature-based analysis, rendered as a scatter plot in Figure \ref{fig:dicarlo-vs-spray} (left), singles out classes such as `bird', `dog', `sheep' as strongly discrepant when considering the recognition performance of DNN and FV. On the other hand, classes `horse', `boat', and `airplane' previously identified by SpRAy all remain close to the diagonal line.
For the second part of the analysis relating visual cues to recognition performance, we focus on the class ``horse'', for which the SpRAy eigenvalue spectrum in Figure~\ref{fig:eigenvaluespectrum-horse} identifies four well-distinguishable prediction strategies for the FV model.
The cluster structure identifies the features ``copyright tag'' and ``tourney hurdle'' as strong indicators for class ``horse''.
Given our less controlled data setting (real photographic images, multiple labels per sample possible, no segmentation mask for all images), we evaluate the effects of different sample characteristics by splitting the test samples for class ``horse'' into groups of samples exhibiting the chosen characteristics at different strengths.
Like~\cite{rajalingham2018large}, we choose as image specific features mean pixel luminosity and group images into non-empty bins of luminosity values $]0.2,\, 0.3]$ to $]0.6,\, 0.7]$ in steps of $0.1$.
Similarly, we compute object specific features from the bounding box annotations available for the Pascal VOC data.
Firstly, as an approximate measure to relative object size, we compute the fraction of image pixels covered by class-relevant bounding box areas, in group ranges of $0.1$.
Secondly, we approximate object pose via the (mean) angle to ground of the diagonal line of the class-relevant bounding boxes, in degree.
That is, a value of $45^{\circ}$ corresponds to a square bounding box. Values close to $90^{\circ}$ indicate very slim and tall objects, while values close to $0^{\circ}$ identify images with very flat or wide objects.

Next to those intuitively expectable features, we add the features identified by SpRAy to our analysis, the ``copyright tag'' and ``hurdle''.
We arrange the test samples for class ``horse'' into three groups: One group is comprised of all samples without the considered feature.
A second group is comprised of only those samples showing the feature.
The last group combines both former groups and presents the mixture of samples as present in the test dataset.

We then measure the AP score\footnote{
In contrast to the carefully controlled evaluation setting of~\cite{rajalingham2018large}, the class populations of the Pascal VOC setting are highly imbalanced.
We therefore choose to report our results in Average Precision~\cite{salton1984introduction} (AP), which is in use for model evaluation in the Pascal VOC challenge since 2007~\cite{everingham2007pascal} and yields more representative measures in highly class-imbalanced object retrieval settings compared to accuracy.
}
for those groups (vs.\ all non-``horse'' test samples) and report the Recall (as detection performance within the groups sorted by feature expressiveness).
Results for both models and all considered features are shown in Figure~\ref{fig:dicarlo-vs-spray} and are similar to the ones reported in~\cite{rajalingham2018large} for image-specific and object-specific features:
Both models seem to  prefer reasonably sized samples for the prediction of class ``horse'' and perform better when bounding boxes are oriented upright (i.e.\ horses are expected to stand, not lie down on the ground).
However, the copyright tag feature identified earlier clearly is indicative for class ``horse'' as perceived by the Fisher vector model.
Its presence increases the reported Recall values considerably compared to the absence of the feature, whereas the DNN model is largely unaffected.
The results in Figure~\ref{fig:dicarlo-vs-spray} show that SpRAy is able to identify features crucial to the model's decision making, which can not be discovered by comparing (differences in) AP values.

Another interesting observation from that analysis is that seemingly  both models depend on the presence of tourney hurdles.
A closer inspection of the sample sets reveals that the presence of hurdles is indeed correlated with tall and slim object bounding boxes.
By looking at the corresponding relevance maps, we quickly recognize that neither of both features is the reason for the prediction of class ``horse'' for the DNN model:
Figure~\ref{fig:hurdles} shows images of horses from the group with the tallest bounding boxes, which obtain a perfect Recall value for the DNN model, alongside corresponding relevance maps for both the DNN and FV model.
As identified by SpRAy, the FV model indeed uses the hurdles as feature for predicting class ``horse''.
The DNN however is focusing on the horses themselves -- photographed  mid-jump in frontal view causing the characteristic bounding box shape -- as well as the knees of the rider.
The hurdles themselves, however, are not used for prediction (and thus do not light up in relevance maps and do not form detectable clusters of relevance maps).

Given our results we reason that the above analysis, replicating the execution of the experiment from~\cite{rajalingham2018large}, may be of use as a verification of already identified features relevant to concept recognition.
The analysis \cite{rajalingham2018large} might however prove insufficient for the identification of features themselves:
There is no guarantee that the effect of (only) the target feature expression is quantified, and not the expression of another correlating but unintended feature.

\begin{figure}[t]
\begin{center}
\includegraphics[width=\textwidth]{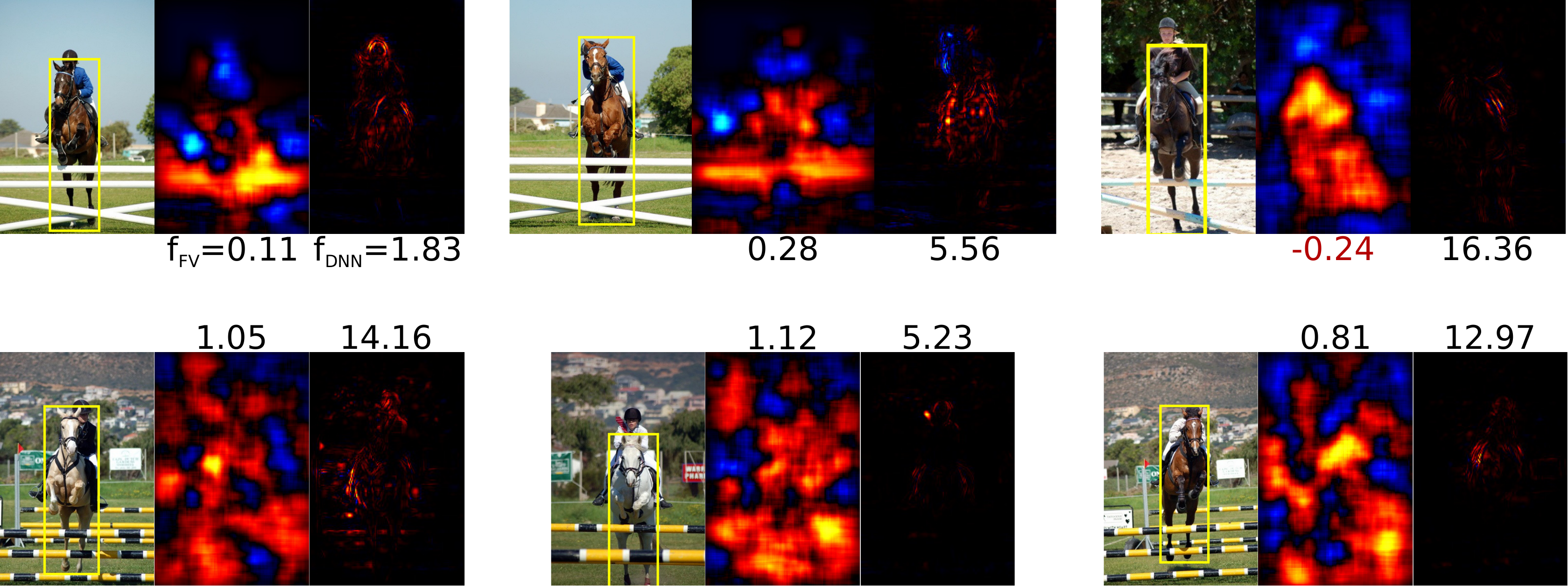}
\end{center}
\caption{Images and relevance maps for the FV and DNN model, corresponding to the group of samples of class ``horse'' with the tallest and slimmest bounding boxes.
Numbers below or above heatmaps show $f_{\text{FV}}$ and $f_{\text{DNN}}$ the predictor output of the FV and DNN model respectively.
The shown relevance maps reveal that the FV model uses the ``hurdle '' feature as an indicator for horseness, whereas the DNN is using the horses themselves.
Our results (compare Figure~\ref{fig:dicarlo-vs-spray}) show that the analysis described in~\cite{rajalingham2018large} may result in the identification of features spuriously linked to model performance.
}
\label{fig:hurdles}
\end{figure}

We further proceed with the spectral and embedding analysis for all classes of the Pascal VOC dataset. For that we compare the spectrum of the first $k=5$ eigenvalues for input images, FV relevances and DNN relevances in search for classes which are candidates for suspicious samples or learned prediction strategies. Figure~\ref{fig:eigenval-barchart} gives an overview over the measured eigenvalue spectra.

\begin{figure}[h]
\center
\includegraphics[width=0.9\textwidth]{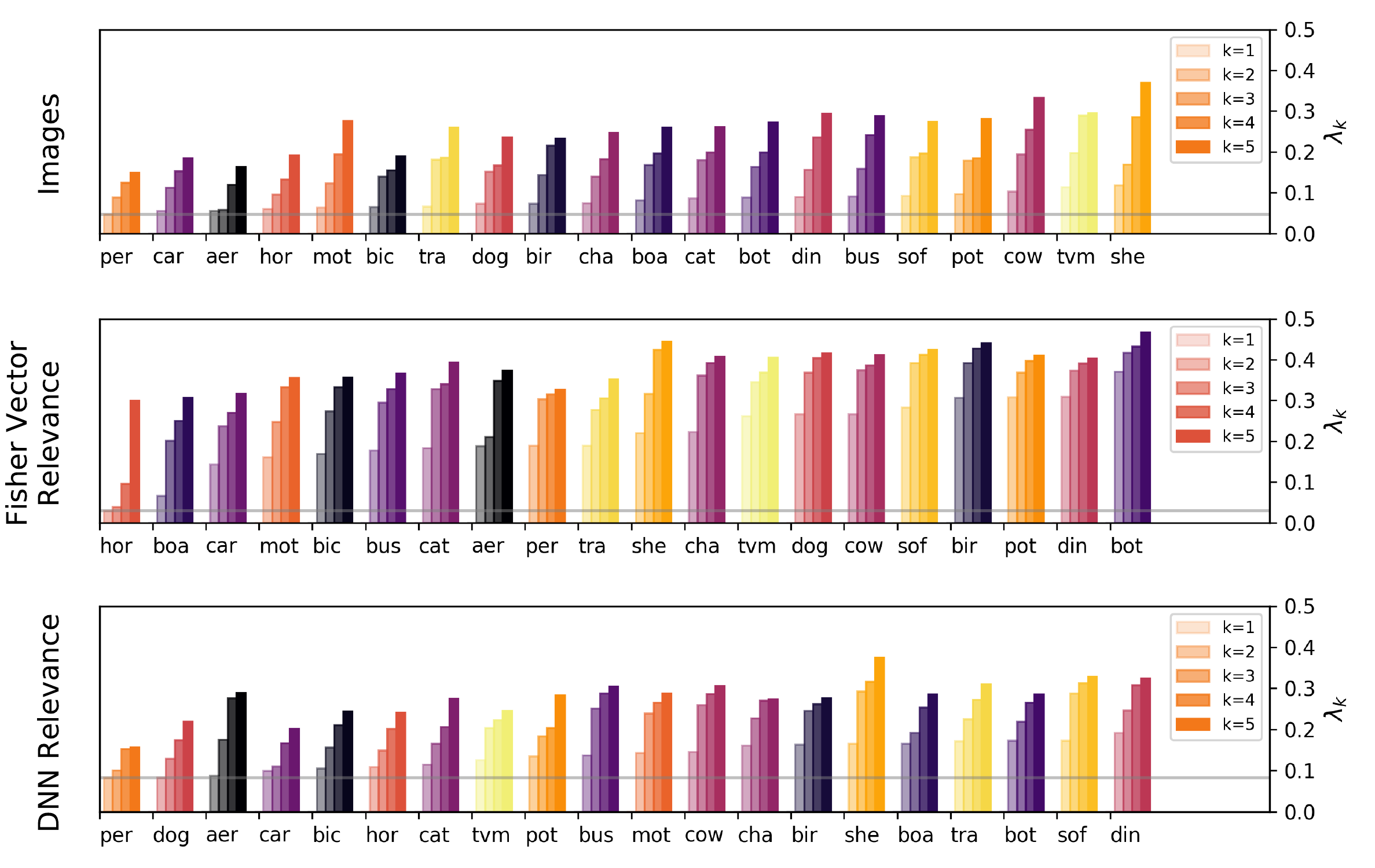}
\caption{The smallest 5 eigenvalues of $L_\text{sym}$ for different input types and all Pascal VOC object categories. The bar charts are colorized wrt.\ to the alphabetical order of the class names and ordered from left to right wrt.\ to $\lambda_2$.}
\label{fig:eigenval-barchart}
\end{figure}

\begin{figure}[h]
\centering
\includegraphics[width=\textwidth]{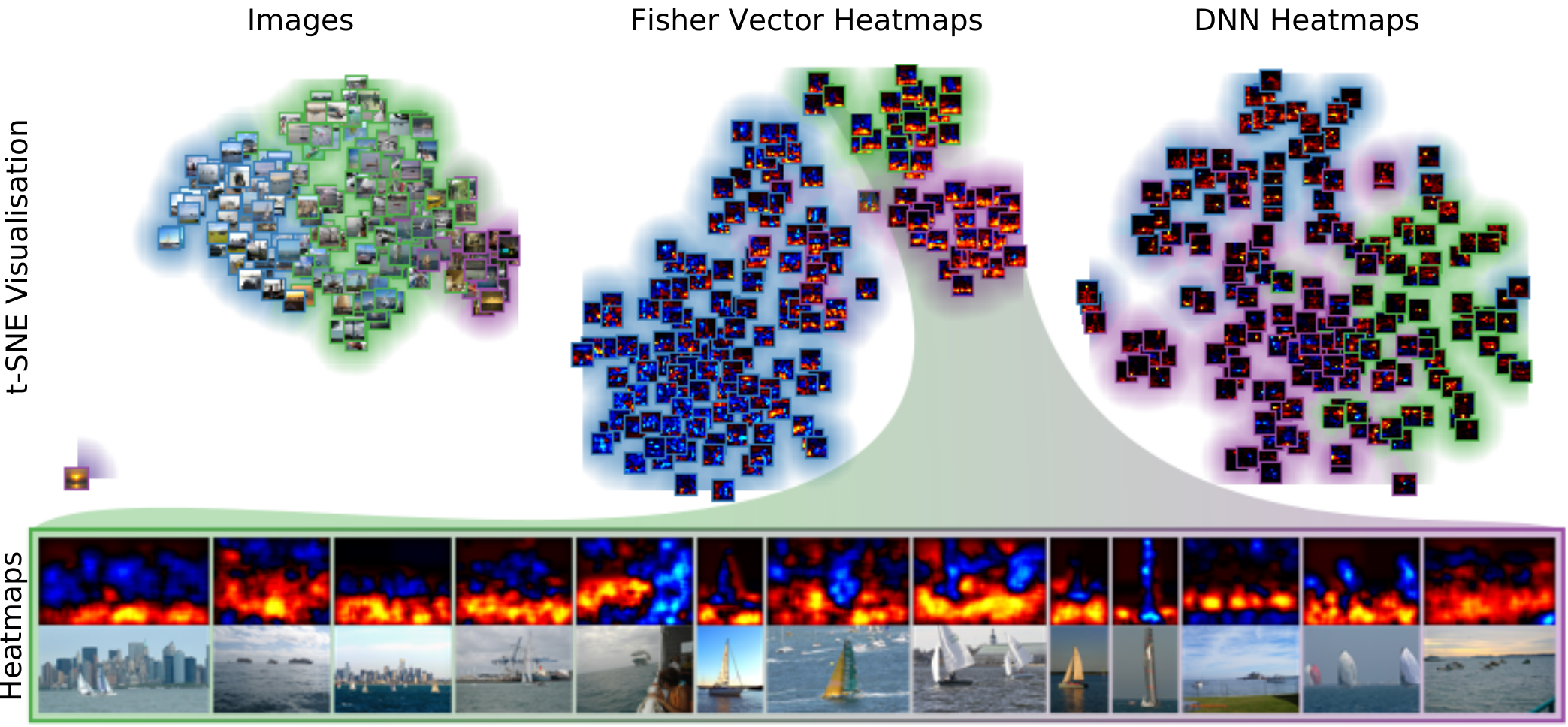}
\caption{Cluster label assigments for class ``boat'' via SC for input images, FV model relevance maps and DNN relevance maps. Embedding coordinates in $\mathbb{R}^2$ for visualization have been computed on pair-wise distances derived from the weighted affinity matrix $W$ used for SC. The samples in the bottom of the figure show FV relevance maps without preprocessing together with corresponding input images.}
\label{fig:tsne+labels-boat}
\end{figure}

We can observe, that the class ``horse'' appears first in the ranking of FV based eigenvalues, at fourth position for images and position six for DNN based eigenvalues. For the FV model, the first two to three eigenvalues are comparatively low, compared to all other classes in all other settings. The eigenvalue spectra for the majority of classes behave similarly in all settings. That is, $\lambda_2$ is significantly larger than $\lambda_1$ (which is always 0 in all cases, as expected), with the consecutive distances between the following eigenvalues being approximately equal, which speaks for one single well-connected point cloud without disjoint clusters. When comparing $\lambda_2$ in all three settings, we observe its value is gradually rising between neighboring  classes in the ranking order. Between the classes ``boat'' ranked 2nd and ``car'' ranked 3rd on the bar plot regarding the FV model, there is a noticeable gap, motivating closer inspection via Figure~\ref{fig:tsne+labels-boat}.

Image inputs for class ``boat'' are clustered due to visual similarity and do not show -- except for one outlier image showing a sundown scene -- unexpected information. The DNN embeddings, together with the relevance plots reveal the locally heterogeneous prediction structure of the DNN, which predicts based on the shape of the shown boats themselves and structures related to boats such as sails. The results for the FV model however reveal one to two weakly connected subclusters besides the main group of samples, with frequently reoccurring relevance structure especially related to true positive predictions (visualized via red aurae). The pattern the FV model has learned to predict on is the water around and below the boats, not the boats themselves. Closer inspection reveals that features from the boats themselves are rated as contradictory information by the FV classifier. We relate the strong reaction of the FV model to water patters to the spatial pyramid mapping scheme \cite{lazebnik2006beyond,bosch2007representing} used in the computation of the FV descriptor \cite{chatfield2011devil}. The use of spatial pyramid mapping allows models to incorporate a weak sense of global shapes and scene geometry into the optimization process, and is known to improve the prediction performance among Bag of Words type models. Conversely, this method of feature pooling also facilitates the development of spatial biases, which might also have affected class ``horse''.

Another class raising further interest is the class ``aeroplane''. Here $\lambda_2 \approx \lambda_3$ for the input images already and for the DNN results show an uncharacteristically fast increase in eigenvalues (i.e.\ there are large gaps between several of the smallest eigenvalues), despite the class being ranked 3rd in Figure~\ref{fig:eigenval-barchart}.

\begin{figure}[h]
\centering
\includegraphics[width=\textwidth]{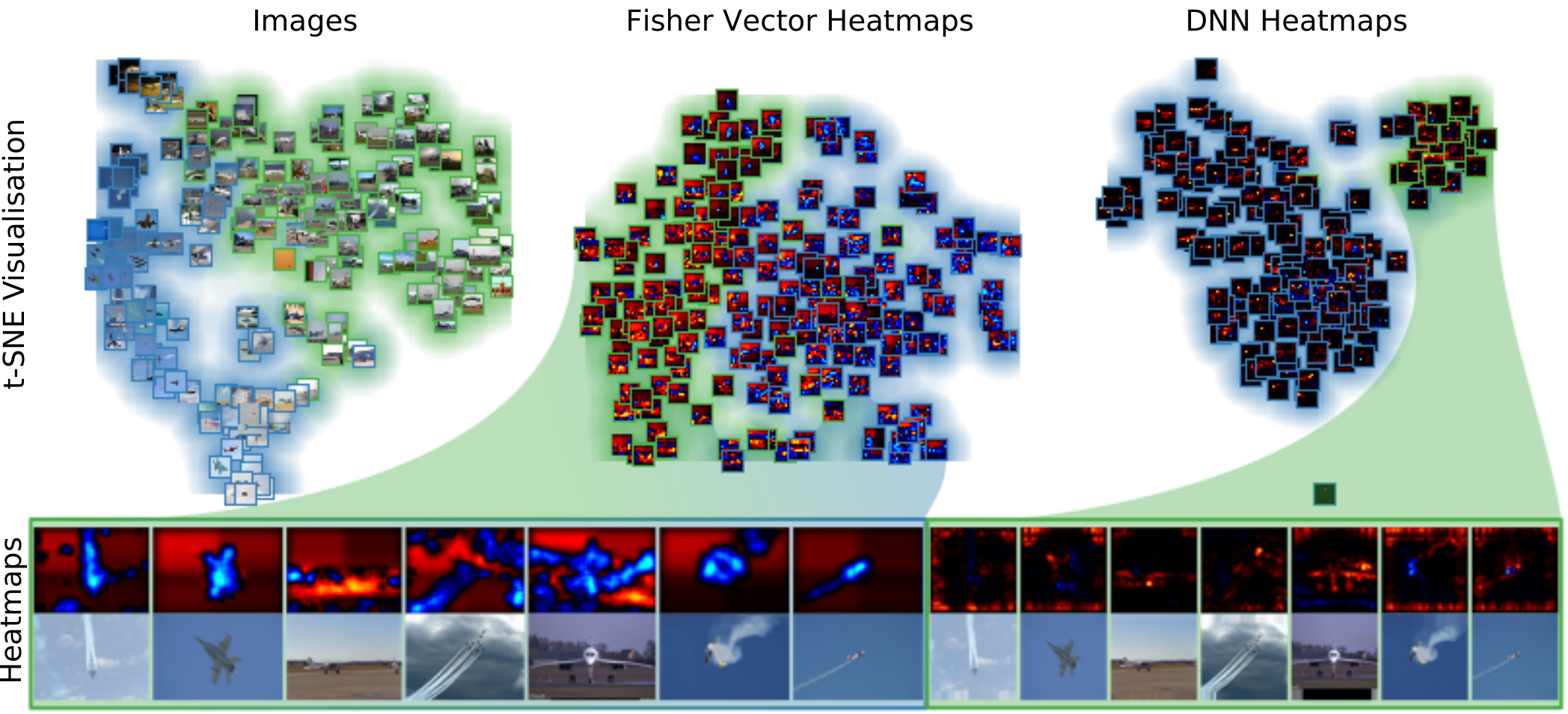}
\caption{Cluster label assignments for class ``aeroplane'' via SC for input images, FV model relevance maps and DNN relevance maps. Embedding coordinates in $\mathbb{R}^2$ for visualization have been computed on pair-wise distances derived from the weighted affinity matrix $W$ used for SC.
The samples at the bottom right (square images) show DNN relevance maps and images with strong reaction of the DNN models to the border padding.
FV relevance maps for the same images are shown to the left.
Enlarged relevance maps and images are shown without preprocessing.}
\label{fig:tsne+labels-aeroplane}
\end{figure}

Figure~\ref{fig:tsne+labels-aeroplane} shows that the low 2nd and 3rd eigenvalues for images can be explained by a cluster of samples which t-SNE embeds as the lower left arc in the visualized $\mathbb{R}^2$-coordinates. The cluster contains images of airplanes in flight in front of blue clear sky. The other half of the embedded images show planes on the ground on rollways and in front of airport structures. The DNN relevance maps cluster a subset of images in front of blue sky, but for an entirely different reason, which becomes apparent by inspecting the relevance maps of the small cluster in the top right of Figure \ref{fig:tsne+labels-aeroplane}: Relevance maps reveal that the DNN model predicts these samples based on the top and bottom border of the image.

We assume the model has learned to predict based on image information introduced as part of the input preprocessing used for training the multi-label model  for \cite{lapuschkinCVPR16}: For preparing the Pascal VOC images as inputs for our DNN model from  \cite{lapuschkinCVPR16}, we scaled down the images to 256 pixels on the longest edge and then padded the image into square shape by copying pixels from the image border, which is a common practice in computer vision. Then, the $227 \times 227$ sized image center was cropped and used as input for the DNN model.
The relevance maps show, that the DNN model picks up heavily on the border regions (especially top and bottom) introduced during image preprocessing, which reflects in the assigned cluster labels and embeddings in $\mathbb{R}^2$.
Considering the DNN model's weak dependency on context for class ``aeroplane'' -- as measured in Figure \ref{fig:context} in Section \ref{sec:importanceofcontext} -- suggests that the border artefact detected by SpRAy is occurring consistently, even though the model does not dominantly depend on it for correctly recognizing airplanes.

We can also observe another artifact for the FV model's predictions which becomes apparent from the t-SNE at closer inspection but is not as apparent in the spectrum of eigenvalues. Again, with high probability due to the use of spatial pyramid mapping, the model reacts strongly to uniform and structureless background in the images, but dominantly in the top half of the image and especially in the top left quadrant. The model's reliance on a lack of structure in the background attributes negative relevance to any object occluding the background. The model has learned to generalize the class  ``uniform image background'' instead of the intended class label ``aeroplane''. 

\subsubsection{Verifying the Detected Bias in Prediction behavior}
The results obtained with SpRAy in Figure \ref{fig:tsne+labels-aeroplane} suggest a previously unknown bias in the decision making of the DNN model, namely that the classifier for class ``aeroplane'' (as part of the trained multi-label DNN) has learned an image border devoid of structure as a characteristic describing the object class, namely class ``airplane''. To further our understanding and to verify our hypothesis of this undesired prediction bias (we wish for the DNN to predict based on the airplane itself) we compare different approaches for obtaining square images as DNN inputs from the Pascal VOC test samples, with examples shown in Figures \ref{fig:aeroplane-example} and \ref{fig:non-aeroplane-example}:\\[+3px]
{\bf copy border}: Image padding by copying pixel values from the image border, as it is performed in the current image prediction pipeline.\\[+3px]
{\bf mirror}: Mirror padding, which extends the image as needed by copying image content and may add natural structures to the image border already present in the image.\\[+3px]
{\bf crop}: Cropping of the largest possible centered image patch, which removes image content from the border and enlarges structures from the image center relatively.\\[+3px]
{\bf sky blue color}: Padding with sky-blue color sampled from a test image, uniformly applied to the padded image area.\\[+3px]
{\bf random color}: Padding with a random color, uniformly applied to the padded image area.\\[+3px]
{\bf random noise}: Padding with random pixel values, adding non-uniform random structure of high frequency to the image border.\\

\begin{figure}[!th]
\begin{center}
\includegraphics[width=0.45\textwidth]{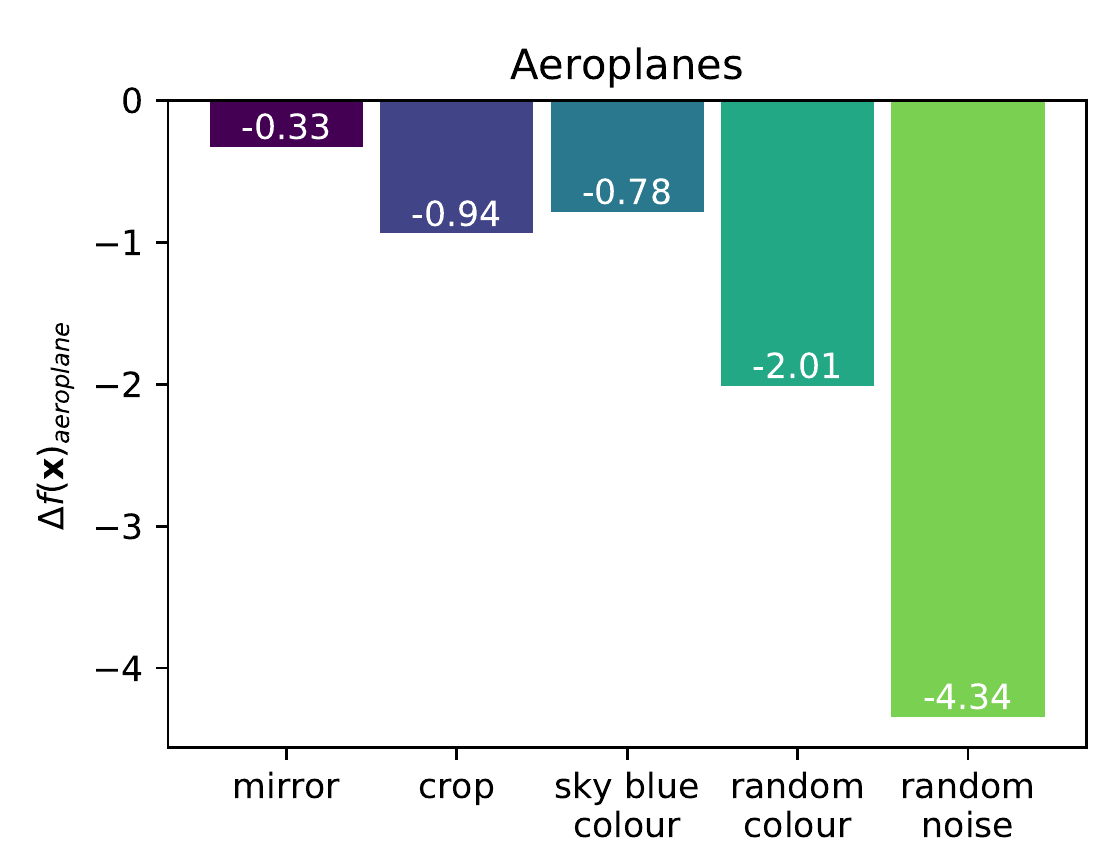}
\includegraphics[width=0.45\textwidth]{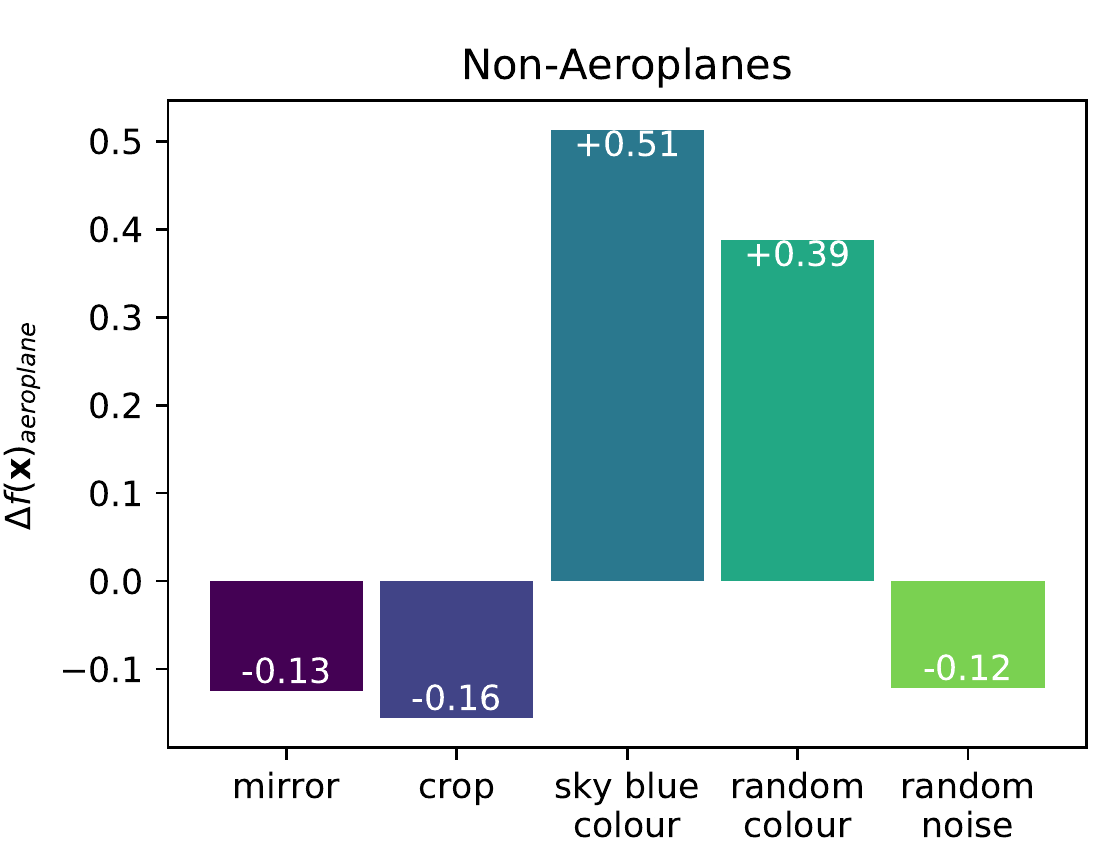}
\end{center}
\caption{Average change in predictor output with different image resizing strategies compared to border pixel copying. \emph{Left:} Averages over all ``aeroplane'' samples. \emph{Right:} Averages over all samples \emph{not} labelled as ``aeroplane''.}
\label{fig:aeroplane-vs-none-border-artefact-barplots}
\end{figure}

\begin{figure}[!th]
\begin{center}
\includegraphics[width=\textwidth]{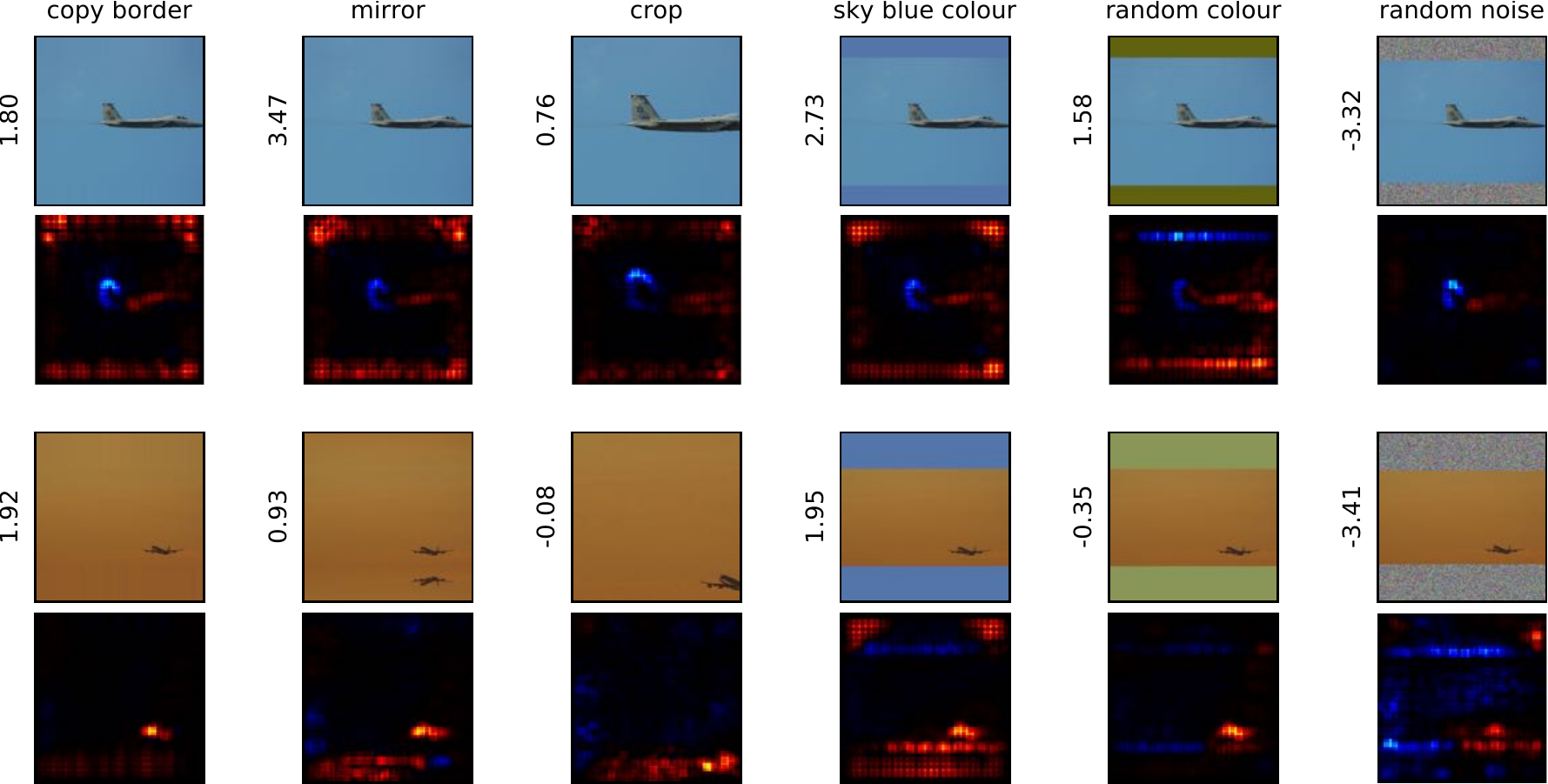}
\end{center}
\caption{Samples from class ``aeroplane'' and predicted scores for class ``aeroplane'', with corresponding relevance maps, as affected by different preprocessing strategies to obtain square images. Padding with (high frequency) random noise effectively decreases the predictor output and removes the ``border artifact''. Using low frequency areas (of the right color) for padding increases the predictor output for class ``aeroplane'' and may even introduce the ``border artifact'' in the first place.}
\label{fig:aeroplane-example}
\end{figure}

\begin{figure}[!th]
\begin{center}
\includegraphics[width=\textwidth]{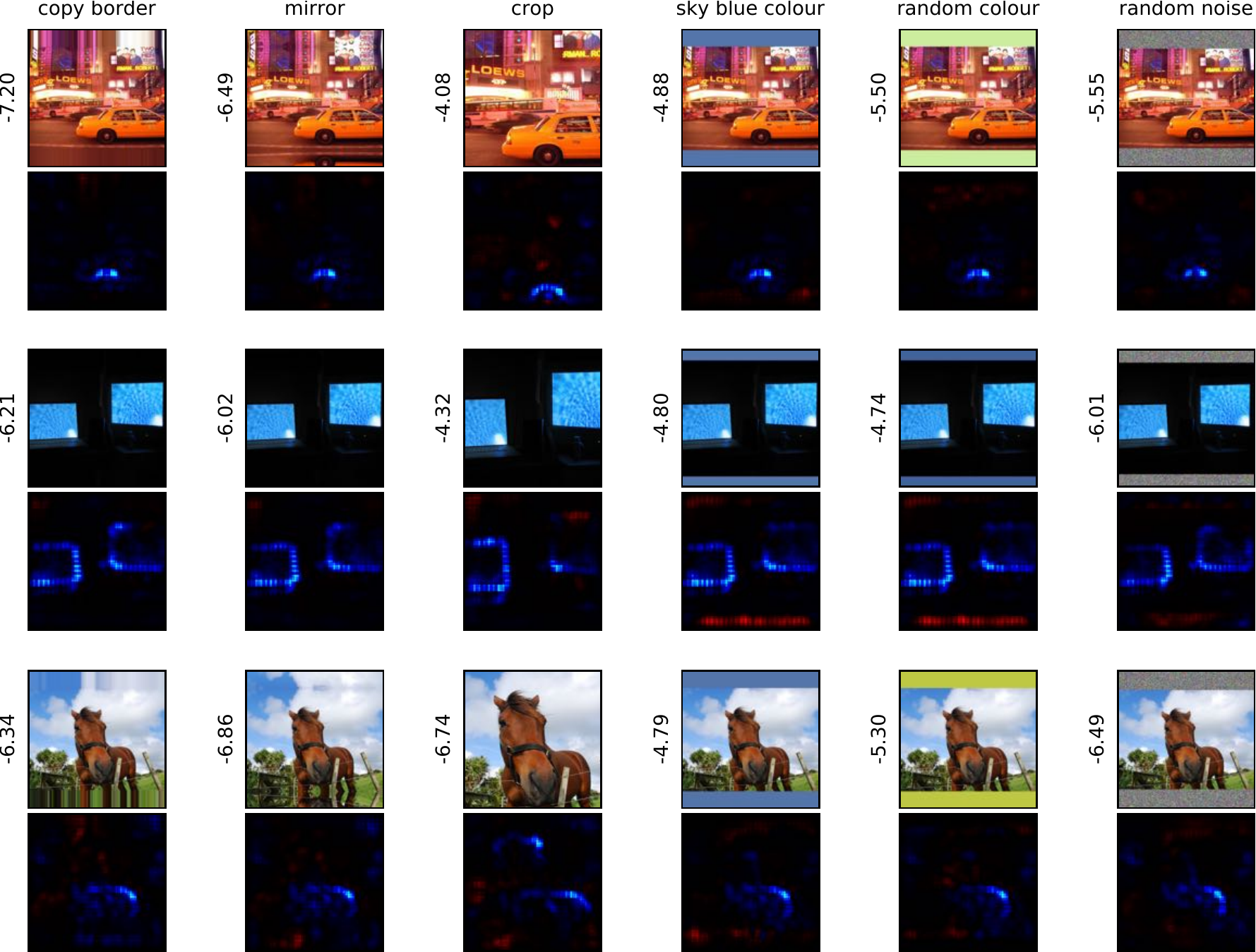}
\end{center}
\caption{Non-``aeroplane'' samples and predicted scores for class ``aeroplane'', as affected by different preprocessing strategies to obtain square images. While the DNN model clearly recognizes parts of the shown non-aeroplane objects as contradicting evidence, the addition of uniformly colored  image extensions provokes a positive response in the prediction and relevance map for class ``aeroplane''.}
\label{fig:non-aeroplane-example}
\end{figure}

We hypothesize, that the model has learned to expect uniformly colored image borders for samples belonging to class ``aeroplane''. We verify that hypothesis in Figure \ref{fig:aeroplane-vs-none-border-artefact-barplots} (left), which shows that any processing resulting in a square image besides the copying of border pixels will result in a decrease \emph{airplaneness} of the sample. Mirror padding can e.g.\ repeat the structure of clouds and cropping will altogether remove border information and move structures from the image center to the (new) image border.

Extending with random pixel values adds high frequency content to the image border, which directly contradicts what the model has learned, resulting in the highest measurable decrease in  the predictor output across all image processing approaches. This approach also reliably removes the artifact of positive prediction contribution of the image border as observable in Figure \ref{fig:aeroplane-example}. While the overall prediction scores for noise-affected samples declines drastically, the center of the relevance maps, showing the actual images, remain almost the same.

Our experiment shows that padding the images with random yet per image constant color value results in a lower decrease of the predictor output compared to random noise. This further verifies that the model expects the absence of structure-rich information. Setting the border color to a sky-blue hue (as taken from test images) reduces the decrease of $f(\x)$ to an lower extent than for cropping and a considerably lower extent than padding with randomly colors. For both examples in Figure \ref{fig:aeroplane-example}, padding with sky-blue color even (marginally) increases the predictor output compared to baseline (copy border) and enhances or even causes the strong appearance of the observed border artifact, despite not matching the \emph{natural} image background color. Further, we observe that the model reacts strongly to transitions between two uniformly colored image areas for both padding variants adding uniformly colored areas, as shown in Figure \ref{fig:aeroplane-example}. This raises further questions about the influence of padding choices of the pre-training phase on ImageNet.

We conduct a second experiment to verify our hypothesis at hand of all test  images \emph{not} being labelled as ``aeroplane'' and apply the same padding approaches (See Figure \ref{fig:non-aeroplane-example} for examples). We provide those sample images as input to the DNN and observe the effect on the model output corresponding to the prediction of airplanes. The results are shown in Figure \ref{fig:aeroplane-vs-none-border-artefact-barplots} (right) verify again that introducing structure (which is with a certain likelihood much stronger for non-``aeroplane'' images) to the border regions of the image (mirror and crop, random noise) reduce the \emph{airplaneness} of the inputs further, while padding with (a fitting) constant color results in a comparatively high increase in the prediction score.

The relevance maps in Figure \ref{fig:tsne+labels-aeroplane} show that positive relevance is strongest on the top and bottom borders of the image, while the left and right image borders only receive weak relevance attribution for aeroplane images. This classifier reaction makes sense intuitively, since photographing an object in the sky likely removes objects located on or near the ground (e.g.\ roof tops), due to the camera's tilt along the vertical axis.

\begin{figure}[!th]
\begin{center}
\includegraphics[width=0.45\textwidth]{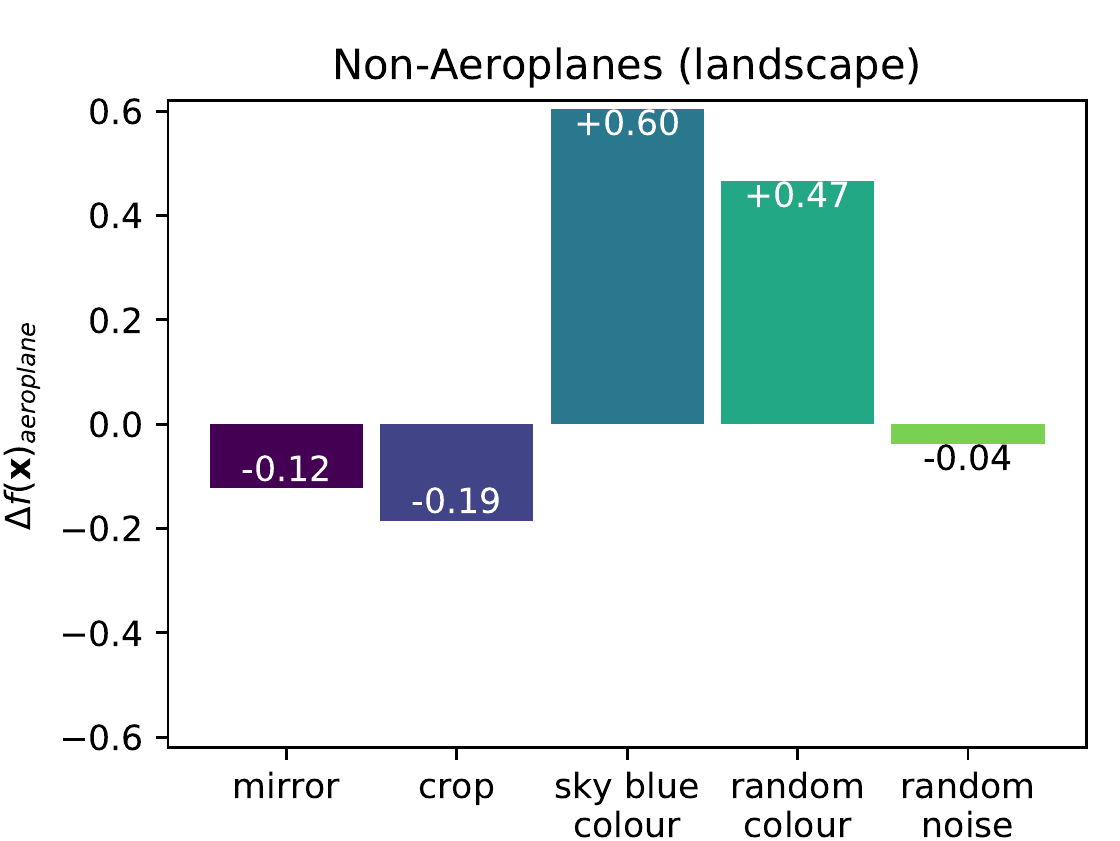}
\includegraphics[width=0.45\textwidth]{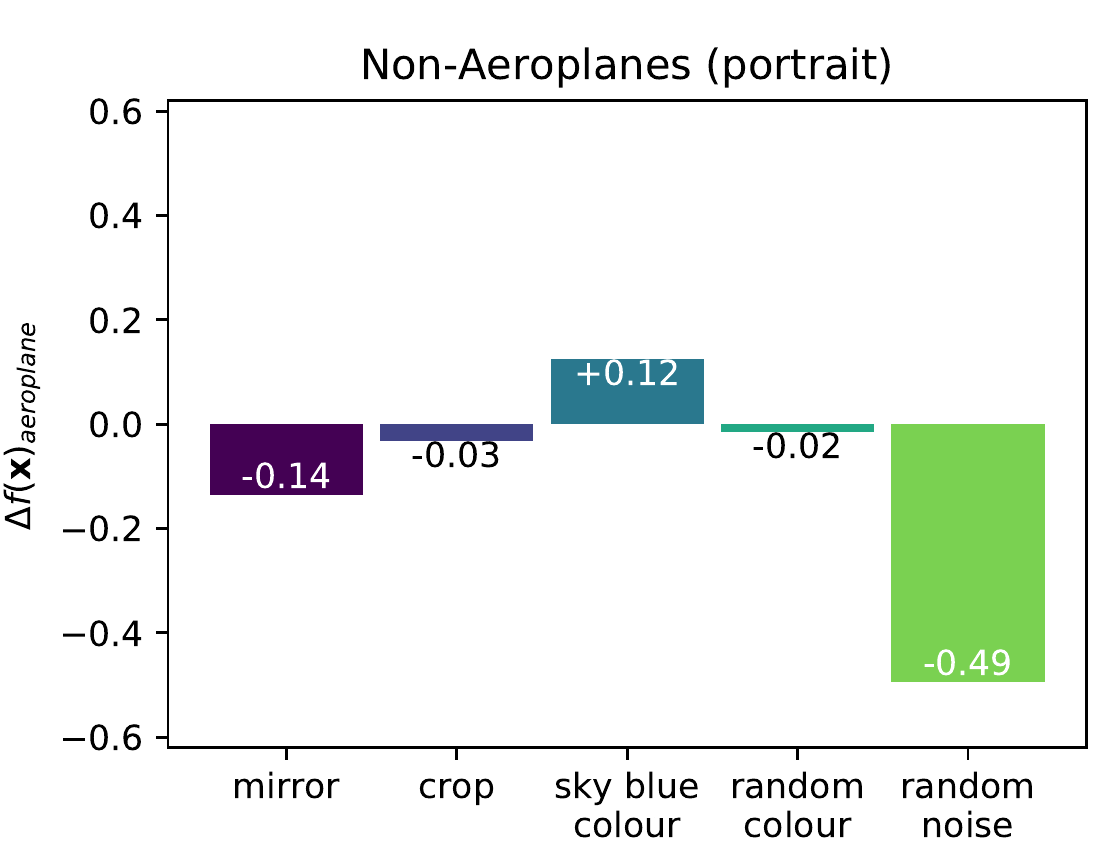}
\end{center}
\caption{Average change in predictor output with different image resizing strategies compared to border pixel copying for non-``aeroplane'' samples. \emph{Left:} Landscape format samples, padded \emph{vertically} (or cropped horizontally). \emph{Right:} Portrait format samples, padded \emph{horizontally} (or cropped vertically).}
\label{fig:vert-vs-horz-border-artefact-barplots}
\end{figure}

We further investigate and subdivide all non-``aeroplane'' images into images which require vertical padding (landscape format, $\approx 3800$ images) and images which require horizontal padding (portrait format, $\approx 900$ images). The results in Figure \ref{fig:vert-vs-horz-border-artefact-barplots} show that padding the horizontal axis (Figure \ref{fig:vert-vs-horz-border-artefact-barplots} (right)) has on average less of an effect to the increase of \emph{airplaneness} via constant border padding compared to padding of the vertical axis (Figure \ref{fig:vert-vs-horz-border-artefact-barplots} (left)).

We can conclude that the DNN has learned to detect airplanes in part based on uniform image areas of constant (sky-like) color at the top and bottom of the image. Using that model as a predictor for airplanes outside of laboratory settings might cause high rates of false positive predictions when observing the sky with a camera system and feeding the image input to the model. With SpRAy, we have pin-pointed that previously unknown artifact in the model's prediction behavior. Future iterations of the model, its training data and preprocessing choices could thus be adapted accordingly to avoid predictions of aeroplanes based on very low frequency image borders.

\subsection{Spectral Relevance Analysis on Atari Gameplay Sequences}
\begin{figure}[!th]
\centering
\includegraphics[width=0.9\textwidth]{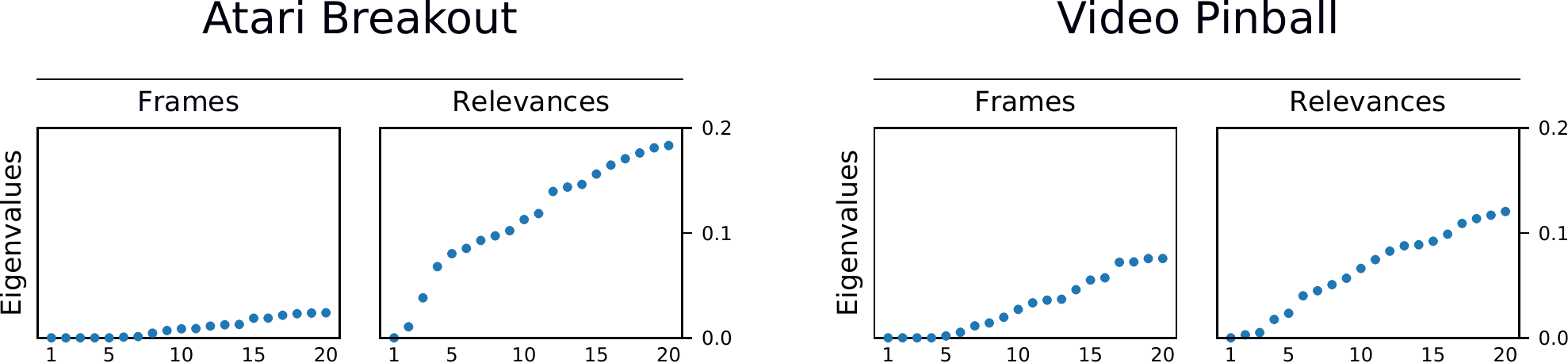}
\caption{The eigenvalue spectra for the Atari 2600 games ``Breakout'' (left) and ``Video Pinball'' (right) for frames recorded over one gameplay session each and corresponding relevance maps as computed for the playing neural network models.}
\label{fig:eigenvalue-spectra-atari}
\end{figure}

In Section \ref{sec:vis_and_comparison_gameplay} we could observe DNNs play the Atari 2600 games ``Breakout'' and ``Video Pinball''. Clearly, for each of the observed games, the respective DNN agent has transitioned through several strategic phases of gameplay. We expect that these observed phases can be detected with the SpRAy algorithm over the duration of a recorded game in the DNNs relevance responses and thus repeat the spectral relevance analysis process from Section \ref{sec:spray} with the Atari gameplay data from Section \ref{sec:vis_and_comparison_gameplay} for both the Atari 2600 games ``Breakout'' and ``Video Pinball'' as shown in Figure \ref{fig:games}. That is, we first record sessions of the trained DNNs playing both games for $\approx 95$ seconds and extract the frames rendered by the Atari emulator, as well as the relevance maps computed for the actions taken by the DNN. Then we apply SpRAy to that data. Figure \ref{fig:eigenvalue-spectra-atari} presents the eigenvalue spectra computed for relevance maps and recorded video frames for both games.

Other than for the experiments based on Pascal VOC, the eigenvalue spectra in Figure \ref{fig:eigenvalue-spectra-atari} show a series of eigenvalues close to zero for the set of frames, indicating a strong fragmentation of the input space into many smaller clusters. As expected, the eigenvalue spectra computed over the relevance values corresponding to each game frame show fewer and more distinctive groupings of samples. For the game ``Breakout'', two eigenvalues are close to zero, followed by two comparatively large eigengaps. The analysis of the sequence of ``Video Pinball'' gameplay reveals three eigenvalues very close to zero, followed by a smaller, yet still distinct eigengap. Both sets of results pronounce many smaller clusters of samples in the input space of game frames (points in time)  and fewer larger clusters among the relevance maps.

\begin{figure}[!th]
\centering
\includegraphics[width=0.9\textwidth]{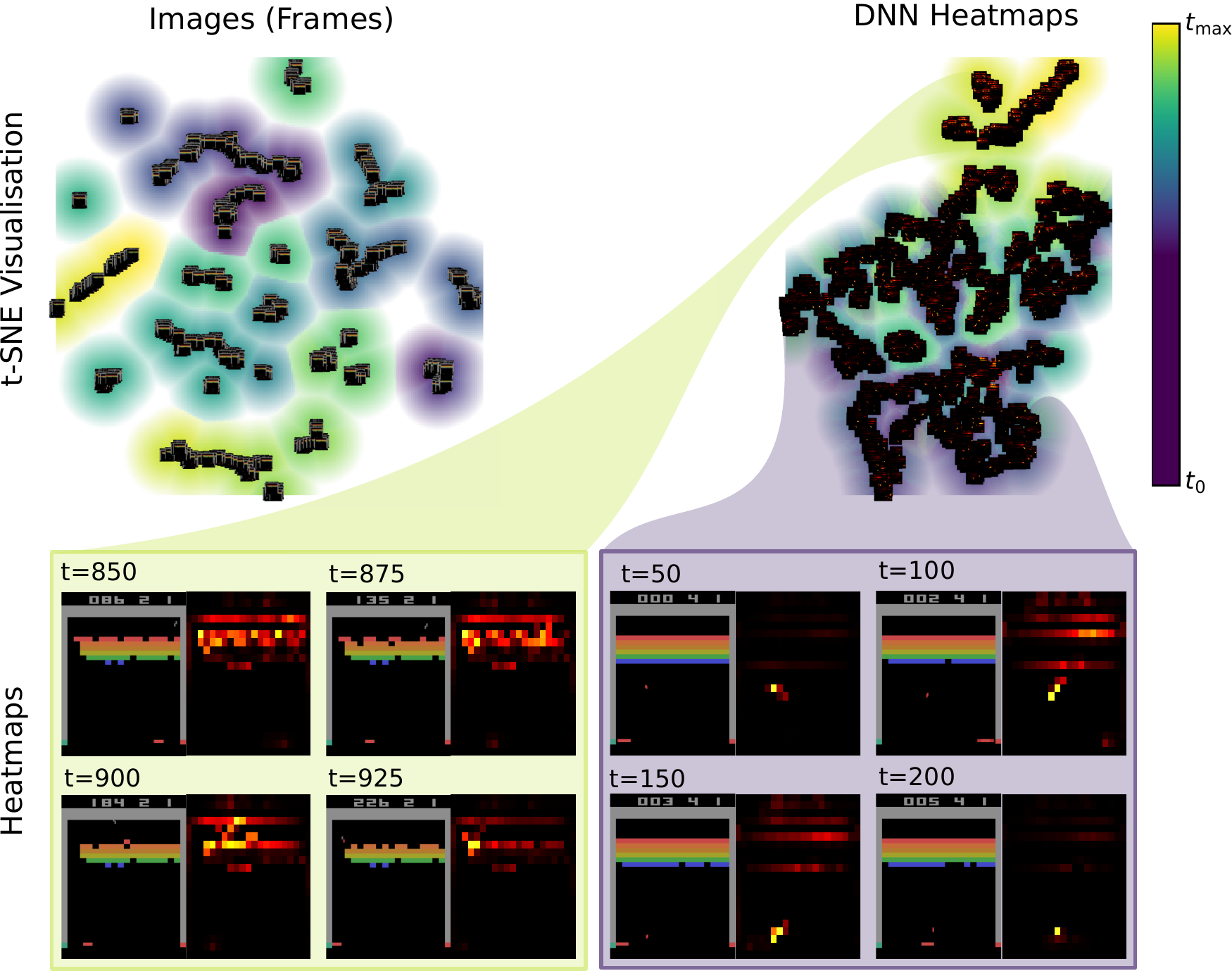}
\caption{Spectral cluster label assignments over a sequence of gameplay (931 time points) of the Atari 2600 game ``Breakout'' for input images and relevance maps. Embedding coordinates in $\mathbb{R}^2$ for visualization have been computed on pair-wise distances derived from the weighted affinity matrix $W$ used for SC. \emph{Left:} Cluster assignments and t-SNE for the frames describing all of the game play session's discrete time points. \emph{Right:} Cluster assignments and t-SNE for the relevance maps corresponding the frames recorded during gameplay. \emph{Top:} Embeddings of the inputs used for analysis. Colored aurae encode the game session's progression through time. \emph{Top Right:} Color legend linking the temporal progression to color hue.  \emph{Bottom:} Exemplary game states and relevance maps in early and late stages (at times $t$) of the game recording.}
\label{fig:tsne+labels-breakout}
\end{figure}

The clusters in the relevance maps in Figure \ref{fig:tsne+labels-breakout} together with the color-coded temporal progression of the gameplay Figure \ref{fig:tsne+labels-breakout} confirm our assumption. Relevance maps are grouped in two to three distinct clusters, covering different phases in gameplay. We can clearly identify important gameplay phases the DNN agent progresses through throughout the recorded sequence of game states. The largest homogeneously connected main group of embedded relevance maps covers an early game phase where the wall of bricks is mostly intact (Figure~\ref{fig:tsne+labels-breakout}: t$\leq$300). A second, smaller and clearly disconnected cluster contains time points from the final stages of the game (Figure \ref{fig:tsne+labels-breakout}: t$\geq$850) when the playing DNN has created the tunnel and was able to maneuver the ball above the wall of bricks, quickly increasing the score counter.

The cluster structure for the input frames show a much higher number of dense groups of samples, dominantly grouped via temporal segments (visible in cluster assignments and t-SNE) of the game. With ongoing progression of the recorded game session the top row of bricks is eroded ongoingly. Next to the fast moving ball and paddle elements this provides a slowly changing connection between points in time for the video frames.

\begin{figure}[!t]
\centering
\includegraphics[width=\textwidth]{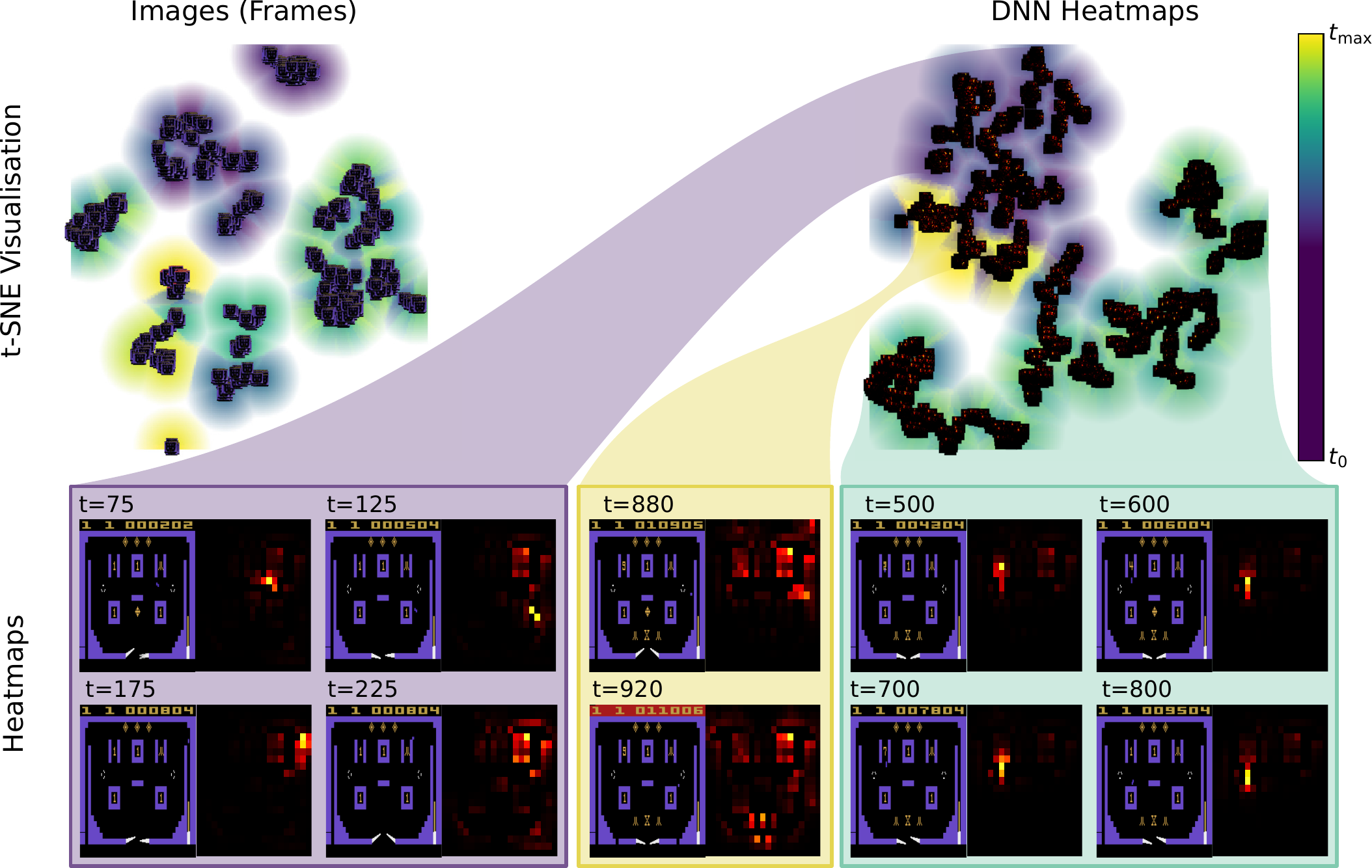}
\caption{Spectral cluster label assignments over a sequence of gameplay (941 time points) of the Atari 2600 game ``Video Pinball'' for input images and relevance maps. Embedding coordinates in $\mathbb{R}^2$ for visualization have been computed on pair-wise distances derived from the weighted affinity matrix $W$ used for SC. \emph{Left:} Cluster assignments and t-SNE for the frames describing all of the game play session's discrete time points. \emph{Right:} Cluster assignments and t-SNE for the relevance maps corresponding the frames recorded during gameplay. \emph{Top:} Embeddings of the inputs used for analysis. Colored aurae encode the game session's progression through time. \emph{Top Right:} Color legend linking the temporal progression to color hue. \emph{Bottom:} Exemplary game states and relevance maps in early and late stages (at times $t$) of the game recording.}
\label{fig:tsne+labels-pinball}
\end{figure}

For the game ``Video Pinball'' the relative size of the eigengap in Figure \ref{fig:eigenvalue-spectra-atari} is not as noticable as for ``Breakout'', but three main clusters can clearly be identified for the analyzed relevance maps. The (two to) three main clusters can also clearly be identified in the t-SNE plots in Figure \ref{fig:tsne+labels-pinball}. Again, both the assigned cluster labels and groups formed identified via t-SNE correlate with the temporal progression of the game for images as well as inputs, albeit the identification of fewer larger groups corresponding to game time intervals is more clear for the analyzed relevance maps. Again, the video frame based analysis makes use of the change (and similarity) in ball position and the state of the game screen itself. The relevance maps computed during gameplay effectively act as filters for momentarily important game areas and thus lead to a more coherent temporal groupings in the embedding space.

The first main cluster found when analyzing the relevance maps computed for the model's preferred actions (i.e.\ maximally firing network outputs) during gameplay corresponds to the early stages of the game, where the DNN spends $\approx 30\%$  of the recorded time. Here, the model makes efforts positioning the ball near the top right game element (Figure \ref{fig:tsne+labels-pinball}: $t \leq 225$), before passing through exactly four times to gain an extra ball. In this phase, most relevance is attributed to the ball, as well as the targeted top right area of the screen, where the model aims to move the ball.

Then, after having the ball pass the top right game element exactly four times, the playing DNN moves the ball to the left of the playing field, where it tries to move the ball as often as possible through the top left game element. This repeated up and down movement results in the second and third clusters (high ball position vs.\ low ball position, relevance allocated accordingly) (Figure \ref{fig:tsne+labels-pinball}: $500 \leq t \leq 800$) and explains the convoluted temporal structure within this cluster of samples. The model spends about $50\%$ of the recorded game time in that phase.

The DNN eventually makes a mistake and the ball ricochets away from the top left game element. The model re-enters the first stage of game play where it has to re-gain control over the ball to position it advantageously in the game field (Figure \ref{fig:tsne+labels-pinball}: $t \geq 880$).

Our results from Figures \ref{fig:tsne+labels-breakout} and \ref{fig:tsne+labels-pinball} demonstrate that with SpRAY, relevance maps can be used to identify and isolate important temporal phases in time series data which are not as clearly recognizable from the model's input data directly.

\subsection{Concluding Remarks on Spectral Relevance Analysis}
SpRAy enables researchers and machine learning engineers to understand the classifier's prediction strategy in detail and in only a matter of seconds, closing the gap between one-dimensional performance measures over test datasets -- such as accuracy and loss ratings -- and the manual assessment of explanation heatmaps for single predictions. With the help of relevance inputs generated with LRP, spectral analysis reveals the composition of selected sample sets in terms of different applied prediction strategies, e.g.\ whether a model's predictions are heterogeneous across all samples or show suspicious or artifactual behavior, using unintended or surprising information. The complementary embeddings computed via t-SNE -- a non-linear technique related to spectral clustering -- aid in the presentation of the spectral analysis in a human-interpretable manner. Our work presents a comparative analysis of the eigenvalue spectra of all object categories of the Pascal VOC 2007 test set -- a widely used benchmark dataset in computer vision -- which points out issues with the composition of some classes on their own and in combination with choices made regarding the model architecture and preprocessing steps for both the investigated FV and the DNN predictor. 

We found a systematic bias in the prediction strategies of the FV model for classes ``boat'', ``horse'' and ``aeroplane'' connected to the spatial pyramid mapping scheme employed to profit from global shape and geometry information. Regarding the DNN model, our pipeline has helped with the identification of a previously unknown issue caused by the preprocessing of the network inputs, which can thus be rectified in the model's next iteration in the development cycle. Our work proposes an initial application of relevance maps as features for detailed model analysis, focussing on the strategies as  applied by a trained model and as revealed by a meaningful set of explanation heatmaps. 

Applied to recorded gameplay sequences of the Atari 2600 games ``Breakout'' and ``Video Pinball'' as data with a strong temporal aspect, SpRAy was able to find and isolate important gameplay phases on relevance data observed earlier manually which was not as transparently available from the recorded game frames when used input to the analysis.

For future work, we see much potential in the diverse applicability of SpRAy. The choice of explanation method is not limited to LRP and may be used to direct the semantics of the analysis. A first preprocessing choice has been a simple sum pooling algorithm for aggregating relevance onto a smaller grid of values, which preserves the locality of the relevance responses in a robust manner. Different choices for preprocessing will guide the focus onto specific aspects within the heatmap, e.g.\ incorporating an interpolation component could act as a low pass filter and revealing different structures in the explanation. Applying Fourier transforms on the relevance maps in order to operate in the frequency domain instead of the pixel space is another promising alternative for preprocessing.

\bibliographystyle{naturemag}
\bibliography{paper}

\end{document}